\documentclass[Afour,sageh,times]{sagej}

\usepackage[table,dvipsnames,svgnames,x11names]{xcolor}

\usepackage{multirow}
\usepackage{multicol}
\usepackage{siunitx}
\DeclareSIUnit{\pixel}{px}
\DeclareSIUnit\megapixel{MP}
\DeclareSIUnit{\decibel}{dB}
\DeclareSIUnit{\fps}{fps}
\DeclareSIUnit{\USD}{\text{USD}}
\DeclareSIUnit{\ggrav}{\ensuremath{g_{\text{grav}}}}
\DeclareSIUnit{\Mbps}{Mbps}
\DeclareSIUnit{\Gbps}{Gbps}
\DeclareSIUnit{\ppm}{ppm}
\DeclareSIUnit\points{points}
\DeclareSIUnit{\herz}{Hz}

\usepackage{layouts}

\usepackage{listings}
\usepackage{url}

\definecolor{citationpurple}{HTML}{6A1B9A}
\definecolor{referenceorange}{HTML}{D95F02}
\usepackage[colorlinks,bookmarksopen,bookmarksnumbered,citecolor=citationpurple,urlcolor=DeepSkyBlue,linkcolor=referenceorange]{hyperref}

\usepackage[capitalize,nameinlink]{cleveref}
\usepackage{graphicx}
\usepackage{microtype}
\usepackage{lipsum}

\usepackage[colorinlistoftodos,prependcaption,textsize=footnotesize,  textwidth=\dimexpr\marginparwidth]{todonotes}

\reversemarginpar

\usepackage[T1]{fontenc}
\usepackage[utf8]{inputenc}
\usepackage{booktabs}
\usepackage{arydshln}
\usepackage{tabularx}
\usepackage{longtable}
\usepackage{subcaption}
\usepackage{comment}
\usepackage{makecell}
\usepackage{array}
\usepackage{amssymb}
\usepackage{amsmath}
\usepackage{import}
\usepackage[flushleft]{threeparttable}
\usepackage{adjustbox}
\setcitestyle{authoryear,aysep={,},citesep={; },sort}

\usepackage{enumitem}

\newlist{todolist}{itemize*}{2}
\setlist[todolist]{label=$\square$}
\usepackage{pifont}

\usepackage[abbreviations]{glossaries-extra}

\glssetcategoryattribute{abbreviation}{indexonlyfirst}{true}

\glssetcategoryattribute{abbreviation}{nohyperfirst}{true}

\newabbreviation{ate}{ATE}{Absolute Trajectory Error}
\newabbreviation{ape}{APE}{Absolute Position Error}

\newabbreviation{auroc}{AUROC}{Area Under the Receiver Operating Characteristic Curve}
\newabbreviation{accuracy}{Acc}{Accuracy}

\newabbreviation{cnn}{CNN}{Convolutional Neural Network}

\newabbreviation{dof}{DoF}{Degrees of Freedom}

\newabbreviation{fov}{FoV}{Field of View}
\newabbreviation{fpr}{FPR}{False Positive Ratio}
\newabbreviation{fem}{FEM}{Finite Element Method}

\newabbreviation{gmsf}{GMSF}{Graph-based Multi-Sensor Fusion}
\newabbreviation{gnn}{GNN}{Graph Neural Network}
\newabbreviation{gcn}{GCN}{Graph Convolutional Network}
\newabbreviation{gnss}{GNSS}{Global Navigation Satellite System}
\newabbreviation{imu}{IMU}{Inertial Measurement Unit}
\newabbreviation{irl}{IRL}{Inverse Reinforcement Learning}

\newabbreviation{knn}{KNN}{K-Nearest Neighbors}

\newabbreviation{lagr}{LAGR}{Learning Applied to Ground Vehicles}
\newabbreviation{lidar}{LiDAR}{Light Detection and Ranging}
\newabbreviation{lio}{LIO}{LiDAR Inertial Odometry}

\newabbreviation{mlp}{MLP}{Multi-Layer Perceptron}
\newabbreviation{mpc}{MPC}{Model Predictive Controller}
\newabbreviation{mse}{MSE}{Mean Squared Error}

\newabbreviation{ood}{OOD}{out-of-distribution}

\newabbreviation{ptp}{PTP}{Precision Time Protocol}
\newabbreviation{ppp}{PPP}{Precise Point Positioning}

\newabbreviation{rbf}{RBF}{Radial Basis Function}
\newabbreviation{rmp}{RMP}{Riemannian Motion Policies}
\newabbreviation{ros}{ROS}{Robot Operating System}
\newabbreviation{ros1}{ROS~1}{Robot Operating System}
\newabbreviation{roc}{ROC}{Receiver Operating Characteristic}
\newabbreviation{rf}{RF}{Random Forest}
\newabbreviation{rts}{RTS}{Robotic Total Station}
\newabbreviation{rte}{RTE}{Relative Trajectory Error}
\newabbreviation{rpe}{RPE}{Relative Position Error}
\newabbreviation[shortplural=TPS]{tps}{TPS}{total station}

\newabbreviation{sdf}{SDF}{Signed Distance Field}
\newabbreviation{slam}{SLAM}{Simultaneous Localization and Mapping}
\newabbreviation{svm}{SVM}{Support Vector Machine}
\newabbreviation{svc}{SVC}{Support Vector Classifier}
\newabbreviation{wvn}{WVN}{Wild Visual Navigation}

\newabbreviation{vit}{ViT}{Vision Transformer}
\newabbreviation{vio}{VIO}{Visual Inertial Odometry}

\setkeys{glslink}{hyper=false}

\newenvironment{leftitemize}
  {\begin{list}{$\bullet$}{%
      \setlength{\leftmargin}{10pt}%
      \setlength{\itemindent}{10pt}%
      \setlength{\labelwidth}{10pt}%
      \setlength{\labelsep}{5pt}%
      \setlength{\itemsep}{0.0pt}%
      \setlength{\topsep}{1.0pt}%
    }}
  {\end{list}}

\newcommand{\states}{\mathcal{X}}
\newcommand{\Measurements}{\mathcal{Z}}
\DeclareMathOperator*{\argmax}{arg\,max}

\makeatletter
\DeclareRobustCommand\onedot{\futurelet\@let@token\@onedot}
\def\@onedot{\ifx\@let@token.\else.\null\fi\xspace}
\makeatother

\newcommand{\aliastable}[1]{\textcolor{gray!70!black}{(\texttt{#1})}}
\newcommand{\alias}[1]{\textcolor{gray!70!black}{\texttt{#1}}}
\newcommand{\na}{\rule[3.0pt]{0.5cm}{0.5pt}}
\newcommand{\boxi}{\textit{Boxi} }

\definecolor{customgreen}{RGB}{13, 109, 10}
\definecolor{customred}{RGB}{165, 0, 6}
\definecolor{reviewblue}{RGB}{0, 92, 175}

\newif\ifreviewmode
\reviewmodefalse

\ifreviewmode
  \DeclareRobustCommand{\removed}[1]{\texorpdfstring{\textcolor{customred}{#1}}{#1}}
  \DeclareRobustCommand{\added}[1]{\texorpdfstring{\textcolor{customgreen}{#1}}{#1}}
  \DeclareRobustCommand{\revised}[1]{\texorpdfstring{\textcolor{reviewblue}{#1}}{#1}}
  
  \newenvironment{addedblock}{\begingroup\color{customgreen}\ignorespaces}{\endgroup\ignorespacesafterend}
  \newenvironment{revisedblock}{\begingroup\color{reviewblue}\ignorespaces}{\endgroup\ignorespacesafterend}
\else
  \DeclareRobustCommand{\removed}[1]{}
  \DeclareRobustCommand{\added}[1]{#1}
  \DeclareRobustCommand{\revised}[1]{#1}
  \excludecomment{removedblock}

\fi


\crefname{section}{Sec.}{Secs.}
\Crefname{section}{Sec.}{Secs.}

\crefname{figure}{Fig.}{Figs.}
\Crefname{figure}{Fig.}{Figs.}

\newcommand\BibTeX{{\rmfamily B\kern-.05em {i\kern-.025em b}\kern-.08em
T\kern-.1667em\lower.7ex\hbox{E}\kern-.125emX}}

\def\volumeyear{2025}

\begin{document}

\runninghead{Tuna \& Frey \textit{et al.}}

\title{GrandTour: A Legged Robotics Dataset in the Wild for Multi-Modal Perception and State Estimation}
\author{Turcan Tuna$^{\bigstar,}$\affilnum{1}, Jonas Frey$^{\bigstar,}$\affilnum{1, 2, 3}, Frank Fu\affilnum{4}, Katharine Patterson\affilnum{1}, Tianao Xu\affilnum{1}, Maurice Fallon\affilnum{4}, Cesar Cadena\affilnum{1}, Marco Hutter\affilnum{1}}

\affiliation{\affilnum{$\bigstar$}Equal Contribution \\
\affilnum{1}ETH Zurich, Switzerland\\
\affilnum{2}Stanford University, USA\\
\affilnum{3}UC Berkeley, USA\\
\affilnum{4}Oxford University, UK}

\corrauth{Turcan Tuna, Robotic Systems Lab, ETH Zurich, Zurich~8092, Switzerland.}

\email{tutuna@ethz.ch}

\begin{abstract}

Accurate state estimation and multi-modal perception are prerequisites for autonomous legged robots in complex, large-scale environments.
To date, no large-scale public legged-robot dataset captures the real-world conditions needed to develop and benchmark algorithms for legged-robot state estimation, perception, and navigation.
To address this, we introduce the GrandTour dataset, a multi-modal legged-robotics dataset collected across challenging outdoor and indoor environments, featuring an ANYbotics ANYmal-D quadruped equipped with the Boxi multi-modal sensor payload.
GrandTour spans a broad range of environments and operational scenarios across distinct test sites, from alpine scenery and forests to demolished buildings and urban areas, and covers a wide variation in scale, complexity, illumination, and weather conditions.
The dataset provides time-synchronized sensor data from spinning LiDARs, multiple RGB cameras with complementary characteristics, proprioceptive sensors, six ANYmal-mounted Intel RealSense D435i depth cameras, and the ZED2i stereo RGB-D camera.
Moreover, it includes high-precision ground-truth trajectories from satellite-based RTK-GNSS and a Leica Geosystems total station.
This dataset supports research in SLAM, high-precision state estimation, and multi-modal learning, enabling rigorous evaluation and development of new approaches to sensor fusion in legged robotic systems.
With its extensive scope, GrandTour represents the largest open-access legged-robotics dataset to date.
The dataset is available at \url{https://grand-tour.leggedrobotics.com}, on HuggingFace (ROS-independent), and in ROS formats, along with tools and demo resources.

\end{abstract}


\maketitle


\begin{figure*}
    \centering
    \includegraphics[width=1\linewidth]{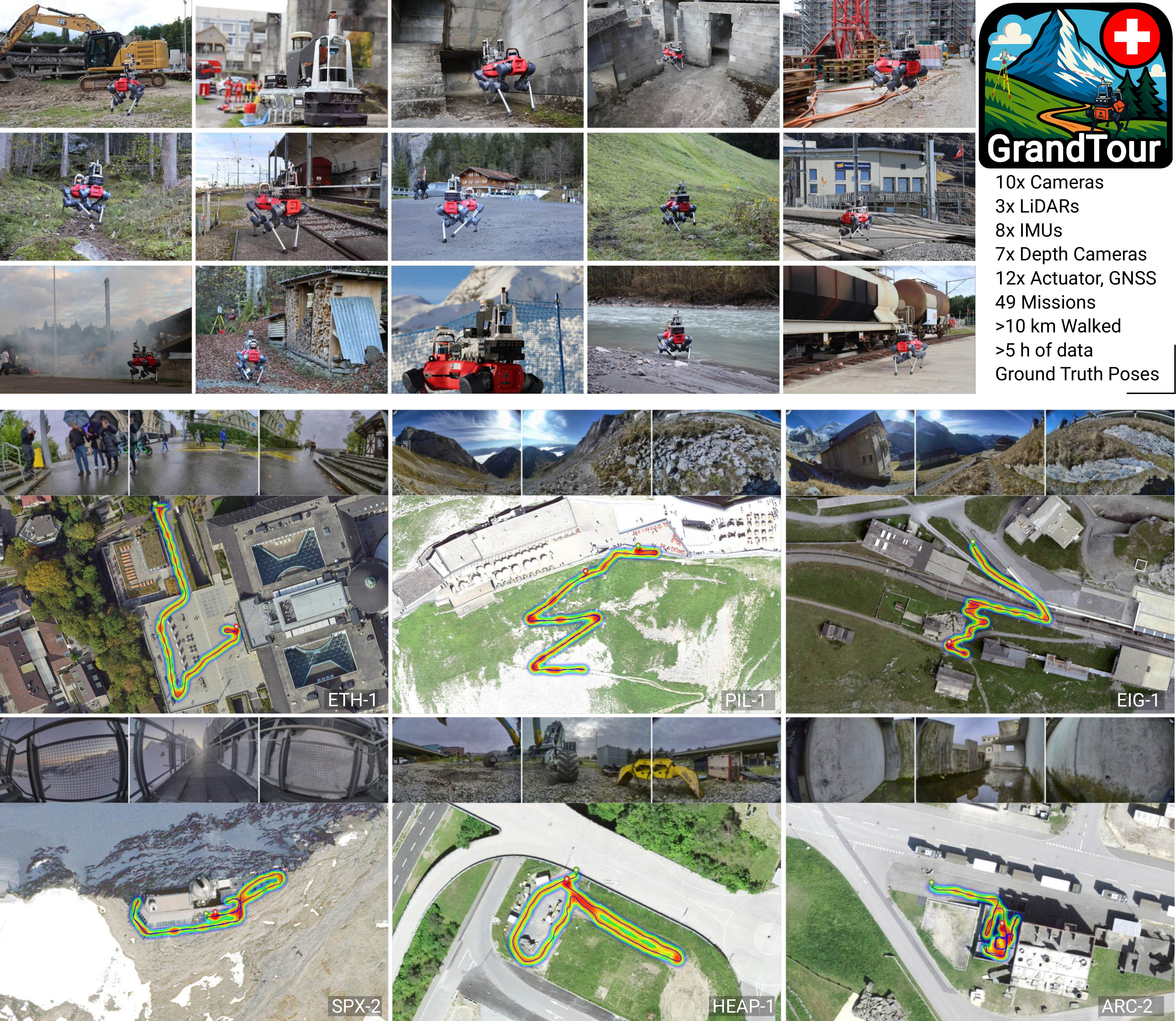}
    \caption{GrandTour dataset preview. Top: views of ANYmal traversing diverse environments during GrandTour, inset summarizes sensor suite and dataset scale (49 missions, $>$\SI{10}{\kilo\meter}, $>$\SI{5}{\hour}). Bottom: aerial imagery alongside on-board RGB views from a subset of the cameras for six missions of GrandTour (ETH-1, PIL-1, EIG-1, SPX-2, HEAP-1, ARC-2). The path color indicates the frequency with which the robot is at that location.}
    \label{fig:data_variety}
\end{figure*}

\section{Introduction}\label{sec:introduction}

\noindent Legged robots have emerged as essential platforms for traversing environments inaccessible to wheeled and aerial robots, with applications ranging from disaster response and industrial inspection to exploration of unstructured terrains.
This expanded mobility exposes autonomy systems to new challenges in robust state estimation, accurate mapping, scene understanding, and safe navigation.

Despite the well-known fact that enabling autonomy in legged robots requires advanced SLAM and multi-modal perception algorithms, targeted progress on these systems has been hindered by the lack of comprehensive public datasets.
Existing datasets, such as KITTI for autonomous driving~\citep{geiger2013vision}, EuRoC for micro-aerial vehicles~\citep{burri2016euroc}, and TUM VI for visual–inertial odometry~\citep{schubert2018tum}, have enabled breakthroughs in their respective domains.
However, analogous datasets for legged robots in real-world environments remain sparse.
Combined, these points highlight the need for legged robot datasets that jointly capture legged robot motion characteristics, the versatile environments in which legged robots are suitable, and multi-modal sensing.

For legged robots, only a handful of datasets exist, \emph{FusionPortable}~\citep{jiao2022fusionportable}, \emph{M3ED}~\citep{chaney2023m3ed}, \emph{DiTer}~\citep{jeong2024diter}, \emph{TAIL}~\citep{yao2024tail}, and \emph{TAIL+}~\citep{wang2024tailplus}, but each is limited either in scale, sensing modalities, or environmental diversity.
As a result, researchers often resort to collecting small-scale, non-representative experimental data for specific scenarios, or to simulations, to develop legged robot state estimation, mapping, and multi-modal perception-to-action methods that may not generalize to the complexity of real-world scenarios.
%
%
To address this gap, we present \textit{GrandTour}, a versatile, large-scale public dataset featuring a state-of-the-art multi-sensor payload on a quadrupedal robot deployed across a wide range of environments.
The dataset is collected using an ANYbotics ANYmal-D~\citep{hutter2016anymal} quadruped robot equipped with \boxi~\citep{frey2025boxi}, a modern and comprehensive sensor payload, yielding an unprecedented set of sensing modalities on a mobile legged robot.
GrandTour captures data from three LiDARs, 10 RGB cameras, seven IMUs (six from the \boxi and one from ANYmal-D), six ANYmal-mounted Intel RealSense D435i depth cameras plus the ZED2i stereo RGB-D camera, 12 proprioceptive sensors, and a GNSS/INS module.
The sensor configuration is described in detail later in \cref{sec:sensors} and illustrated in \cref{fig:side-by-side}.

In GrandTour's main application domains, reliable state estimation, multi-modal perception, and SLAM for legged robots remain challenging.
Firstly, the inherent dynamics of legged locomotion, such as intermittent contacts, foot slippage, and rapid body orientation changes, introduce disturbances to sensor measurements.

Secondly, legged robots are often deployed in unstructured and harsh field conditions, where poor illumination, occlusions, tight spaces, slippery terrain, airborne particles, and textureless surfaces further degrade sensing.
Finally, to address these issues, legged platforms utilize complementary exteroceptive sensors (cameras, LiDAR, GNSS) and high-rate inertial/proprioceptive sensing.
However, exploiting this suite of sensors is non-trivial due to distinct noise profiles, latencies, fields of view, and failure modes under locomotion-induced and environmental disturbances.

For classical state estimation and SLAM, GrandTour provides benchmarks for LiDAR-inertial, visual-inertial, LiDAR-only, LiDAR-inertial-visual, and \added{legged-robot specific kinematic-inertial} odometry, enabling systematic evaluation of multi-modal fusion and drift over different terrains in challenging environments.
Moreover, GrandTour has already been used in sensor fusion frameworks~\citep{nubert2025holistic} and the continuous-time state estimation problem~\citep{resple}.

For learning-based methods, the dataset has already been used to improve sim-to-real transfer~\citep{roth2025}, enable real-to-sim transfer of large-scale environments and neural scene representations~\citep{gauss}, learn physical parameters~\citep{chen2024identifying}, train foundation models for navigation~\citep{lessismore}, benchmark vision-language models for scene understanding~\citep{navitrace}, and support self-supervised representation learning~\citep{patel2025}.

Furthermore, the field deployment of legged robots relies on closed-loop autonomy, where perception and state estimation inform planning and control, while resulting motions affect sensor measurements.
Modern approaches emphasize the action-perception feedback loop~\citep{lee2020making, korthals2022multi}, and in legged systems, this loop is reinforced by the coupling between proprioception and exteroception, where gait patterns affect sensing and thus measurement quality~\citep{bloesch2012state, kim2021legged}.

Consequently, multi-modal perception has emerged as an essential tool for robust autonomy in challenging environments~\citep{cui2023deep, feng2023scale}.
Such rich multi-modal data enables both perceptive locomotion with action-perception feedback~\citep{kumar2021rma, miki2022learning,miki2024learning} and learned legged state estimation that fuses sensor measurements under locomotion-induced disturbances~\citep{buchanan2022learning, youm2025legged}.

Beyond its main applications, GrandTour also enables the extraction of a large number of cross-view image pairs under realistic legged robot motion and field conditions, which can directly support DUSt3R~\citep{wang2024dust3r} and VGGT~\citep{wang2025vggt} visual-geometry tuning for improved geometry and relocalization in legged robots.
Similarly, recent progress on foundation models for robotics~\citep{firoozi2025foundation} further increases the value of large, well-synchronized, and calibrated datasets.
GrandTour provides diverse, synchronized multi-modal legged data with accurate ground truth, making it a strong resource for fine-tuning existing foundation models to improve robustness and transfer to real-world legged deployments.

In summary, the primary contributions of this work are:

\begin{leftitemize}
    \item \textbf{GrandTour Dataset:}
    We introduce GrandTour, a large, open-access legged robotics dataset capturing 49 sequences across indoor facilities, urban outdoors, and natural terrains, representing the \revised{largest open-access legged-robotics dataset} to date.
    GrandTour offers synchronized and calibrated data from an extensive array of sensors including multi-LiDAR, multi-camera, \revised{multiple depth cameras}, IMUs, \revised{GNSS}, and full proprioception.
    Each data sequence is accompanied by high-precision ground truth, enabling rigorous benchmarking of \revised{state estimation, odometry, and sensor fusion algorithms} under realistic conditions.
    \item \textbf{A Comprehensive Localization Benchmark:} We evaluate 63 state-estimation and localization methods on six GrandTour missions to analyze their performance and methodologies in depth.
    \item \textbf{Broad Utility and Open-Source Release:} We release GrandTour to the community under an open-source license, along with tools and documentation to facilitate its use.
    Critically, the GrandTour dataset provides intermediate and processed outputs, such as legged-inertial odometry, deskewed point clouds, LiDAR-inertial odometry, terrain maps, point-cloud maps, and occupancy maps, to facilitate ease of adoption and targeted development by researchers.
\end{leftitemize}

\setcounter{footnote}{3}
\renewcommand{\thefootnote}{\arabic{footnote}}


\section{Related Work}\label{related}

\begin{table*}[t]
  \footnotesize
  \centering
  \resizebox{1\textwidth}{!}{
    \begin{threeparttable}
      \caption{Comparison of related datasets. Each row lists dataset name, \textbf{platform abbreviation(s)}, environment types, diversity of scenarios, and sensor coverage. \textbf{Platform abbreviations:} A = Aerial, W = Wheeled, L = Legged, C = Car, H = Handheld, U = UGV. The dashed horizontal divider separates datasets that include legged platforms (below) from those that do not (above).}
      \label{tab:datasets}
      \footnotesize
      \begin{tabular}{p{4.4cm}p{1.4cm}p{3.3cm}p{3.9cm}p{6.3cm}}
        \toprule
        Dataset & Platform & Environment & Diversity & Sensor coverage \\
        \midrule
        KITTI~\citep{geiger2013vision} & C & City, rural roads & 22 sequences; \SI{6}{\hour} & Stereo cams, LiDAR, GPS, IMU \\
        Oxford RobotCar~\citep{Maddern2017Robotcar} & C & Urban & \SI{1000}{\kilo\meter}; year-long; all-weather & 6 cams, 2D/3D LiDARs, GPS/INS \\
        Boreas~\citep{burnett2023boreas} & C & Repeated route & 350~km; multi-season (1~yr) & LiDAR, 360° radar, RGB cam, GNSS, IMU \\
        EuRoC~\citep{burri2016euroc} & A & Indoor machine halls & 11 sequences; varying difficulty & Stereo cam, IMU \\
        TUM~VI~\citep{schubert2018tum} & H & Indoor/outdoor & 13 sequences; HDR imagery & Stereo cams, IMU, Vicon GT \\
        Newer College~\citep{ramezani2020newer} & H & Campus buildings/quads & \SI{6.7}{\kilo\meter}; buildings \& vegetation & LiDAR, stereo RGB cams, IMU \\
        M2DGR~\citep{yin2021m2dgr} & U & Indoor/outdoor & 36 sequences; campus & Fisheye, IR\,\& event cams, LiDAR, IMUs, GNSS \\
        MulRan~\citep{kim2020mulran} & C & City streets & Multi-city; long trajectories & Radar, LiDAR \\
        MCD~\citep{nguyen2024mcd} & U, H & Multi-campus indoor/outdoor & 59~k labelled scans; 3 campuses & LiDAR, RGB, IMU, UWB \\
        Oxford Spires~\citep{tao2025oxfordspires} & H & Multi-campus, outdoor & Urban, historic landmarks & LiDAR, RGB, IMU \\
        GND~\citep{jing2025gnd} & U & Multi-campus, outdoor & 10 campuses; \SI{53}{\kilo\meter}; \SI{11}{\hour}; & LiDAR, RGB, 360° cam, GPS, IMU, Wheel enc. \\
        \\[-0.5em]
        \cdashline{1-5}\\[-0.5em]
        FusionPortable~\citep{jiao2022fusionportable} & L, H, U, C & Indoor, campus, urban & 17 sequences; multi-platform & LiDAR, RGB, event cams, IMU, GNSS \\
        FusionPortableV2~\citep{wei2025fusionportablev2} & L, H, U, C & Buildings, campus, urban & 27 sequences; \SI{2.5}{\hour}; varied motion & \revised{LiDAR, frame/event cams, IMU, GNSS, proprioception} \\
        M3ED~\citep{chaney2023m3ed} & L, W, A & Off-road trails, dense forests & Multi-robot; event-centric & Event cams, RGB, IMU, LiDAR \\
        SubT-MRS~\citep{zhao2024subt} & L, A, W & Vast and diverse & \SI{300}{\kilo\meter}; multi-year; all-weather & LiDAR, fisheye, depth, thermal cams, IMU \\
        TAIL~\citep{yao2024tail} & L, W & Granular soil & 14 sequences; & LiDAR, RGB(D) cams, IMU, GNSS, kinematics \\
        TAIL-Plus~\citep{wang2024tailplus} & L, W & Granular soil & Multi-loop; day/night & LiDAR, RGB-D, RGB, IMU, RTK-GPS \\
        DiTer~\citep{jeong2024diter} & L & Parks, slopes, sand & Diverse terrain & LiDAR, RGB-D, thermal cams, IMU, GNSS \\
        DiTer++~\citep{kim2024diter+} & L & Park, forest & Day/night & LiDAR, RGB, thermal cams, IMU, GNSS \\
        EnvoDat~\citep{nwankwo2024envodat} & L, W & Indoor, underground, urban & 26 sequences; 13 scenes; fog, rain & RGB, depth, thermal/IR, LiDAR, IMU \\
        \textbf{GrandTour (ours)}  & L & Vast and diverse  & 49 sequences, >\SI{5}{\hour}, >\SI{10}{\kilo\meter} & \revised{LiDAR, RGB/RGB-D, IMU, GNSS, joint enc.} \\
        \bottomrule
      \end{tabular}
    \end{threeparttable}
  }
\end{table*}

\subsection{Popular robotic state estimation datasets}
A variety of datasets have shaped the development of SLAM and robotic perception by pairing versatile sensor suites with high-quality ground truth. The KITTI benchmark~\citep{geiger2013vision} remains highly influential for stereo, visual odometry, and 3D perception with synchronized cameras, LiDAR, and GPS/INS ground truth. For aerial robots, EuRoC MAV~\citep{burri2016euroc} established a visual–inertial odometry evaluation using synchronized stereo and IMU data, along with pose ground truth from Vicon and Leica tracking systems. TUM VI~\citep{schubert2018tum} extended VI evaluation with a stereo camera, photometric calibration, an IMU, and motion-capture ground truth at sequence start/end. The Newer College dataset~\citep{ramezani2020newer} provided handheld LiDAR–visual–inertial recordings on the Oxford campus with centimeter-accurate trajectories obtained by registering mobile LiDAR to a millimeter-accurate TLS map. The Hilti SLAM Challenge datasets (2021–2023) benchmark high-precision SLAM in GNSS-denied industrial settings using calibrated visual/LiDAR/IMU suites with sparse ground truth~\citep{helmberger2022hilti, zhang2022hilti}. 
Together with large-scale datasets such as Oxford RobotCar~\citep{Maddern2017Robotcar} and Boreas~\citep{burnett2023boreas}, these datasets form a common reference set for robotic perception research.

However, these reference datasets primarily target wheeled, aerial, or handheld platforms and therefore do not capture the locomotion-induced disturbances, proprioception, exteroception coupling, or unstructured terrain that characterize legged autonomy.
GrandTour is intended as a legged-robot counterpart with comparable calibration and ground-truth fidelity, extending this lineage of datasets with a tightly synchronized multi-LiDAR/multi-camera/IMU payload, plus full proprioception, enabling missions with loop closures and coverage of indoor, urban, industrial, and natural environments.\looseness-1

\subsection{Multi-modal perception datasets}
While many earlier datasets already include more than one exteroceptive sensor, ongoing efforts specifically focus on collecting multimodal data.
This shift is driven by the fact that no single sensor remains reliable across all environments and failure modes. Combining complementary modalities (e.g., LiDAR geometry with visual/thermal semantics or radar robustness) enables robustness under lighting, adverse weather, sparse texture, and self-similar or dynamic scenes.

M2DGR~\citep{yin2021m2dgr} equips a ground robot with fisheye, infrared, and event cameras alongside a 32-beam LiDAR, multiple IMUs, and GNSS receivers, offering indoor and outdoor sequences with ground truth from motion-capture systems and laser trackers. FusionPortable and FusionPortableV2~\citep{jiao2022fusionportable, wei2025fusionportablev2} extend this concept by mounting \revised{unified multi-sensor payloads with LiDAR, frame/event cameras, IMU, and GNSS} across handheld, legged, and wheeled platforms to enable cross-platform benchmarking.
The MulRan dataset~\citep{kim2020mulran} highlights radar–LiDAR fusion for place recognition in urban driving, while M3ED~\citep{chaney2023m3ed} introduces synchronized event cameras, LiDAR, and inertial sensing across wheeled, legged, and aerial robots in natural environments.
The MCD dataset~\citep{nguyen2024mcd} augments cameras and LiDARs with \revised{IMU and UWB sensing} and releases sequences recorded during day, night, and adverse weather, explicitly targeting robustness under challenging conditions.

The Oxford Spires dataset~\citep{tao2025oxfordspires} was collected around historic landmarks in Oxford using a perception unit with three synchronized global-shutter color cameras, a large field of view 3D LiDAR, and an inertial sensor, with millimeter-accurate TLS scans providing ground truth for localization and 3D reconstruction.
Highlighting the limitations of existing multimodal perception datasets, the authors in~\citep{soares2025smapper} introduce the SMapper handheld data acquisition platform and subsequently the SMapper-light dataset, primarily targeting multimodal SLAM applications across diverse environments.

Collectively, these datasets highlight a growing focus on fusing LiDAR, RGB, and inertial modalities, often complemented by radar or event sensors, to develop robust perception pipelines across diverse environments and robotic platforms.\looseness-1

%
%

\subsection{Legged-robot datasets}
Compared to other platforms, public legged-robot datasets remain limited. \emph{TAIL}~\citep{yao2024tail} targets sandy, deformable terrains using a rotating 3D LiDAR, a forward-facing stereo pair, multiple downward RGB-D cameras, an IMU, and RTK, and logs proprioceptive states (gait/motor/foot status).
\emph{TAIL-Plus}~\citep{wang2024tailplus} adds more sequences with multiple loops and day–night traverses, includes an extra IMU, and was collected on beach analog sites for granular terrains. \emph{DiTer}~\citep{jeong2024diter} focuses on legged robots traversing diverse outdoor terrains and provides terrain-facing RGB-D, RGB, thermal, LiDAR, IMU, GPS, and built-in odometry. \emph{DiTer++}~\citep{kim2024diter+} extends this setting to multi-robot, multi-session legged data in lawn, park/campus, and forest environments, with heterogeneous RGB/RGB-D, thermal, LiDAR, IMU, and robot joint/contact sensing, plus day--night sessions and survey-map-based ground truth.
\revised{Multi-platform datasets such as \emph{M3ED}~\citep{chaney2023m3ed} and \emph{SubT-MRS}~\citep{zhao2024subt} include legged platforms: \emph{M3ED} emphasizes event-camera benchmarks across ground/legged/aerial platforms, and \emph{SubT-MRS} stands out for its scale and emphasis on robustness in degraded sensing conditions (e.g., low light, obscurants, and adverse environments), with synchronized LiDAR and multi-camera (including thermal) sensing and long-term collection.} However, as a multi-robot dataset spanning aerial, wheeled, and legged platforms, \emph{SubT-MRS} is not primarily optimized around legged-specific sensing and evaluation (e.g., rich proprioception, contact events, or legged-only mission structure).
\emph{EnvoDat}~\citep{nwankwo2024envodat} collects multiple sequences across indoor, outdoor, and underground scenes with ten modalities on quadruped and wheeled platforms, using RGB-D and monocular cameras rather than multi-camera rigs. \emph{FusionPortable} and \emph{FusionPortableV2}~\citep{jiao2022fusionportable, wei2025fusionportablev2} mount a unified multi-sensor payload across handheld, legged, and wheeled platforms; notably, \emph{V2} records Unitree A1 proprioception (joint encoders, foot contacts) in addition to \revised{LiDAR, frame/event cameras, IMU, and GNSS}.\looseness-1

Despite these efforts, large-scale datasets focused on legged robots remain scarce. To the best of our knowledge, \emph{GrandTour} is the \revised{largest open-access legged-robotics dataset} to date, filling a unique gap by combining multiple LiDARs, multiple RGB/depth camera bundles, high-grade IMUs, and rich proprioception with dual RTK-GNSS and survey-grade ground truth. GrandTour spans more than 49 missions across urban, industrial, and natural settings, including day/night operations and adverse weather conditions.
By combining survey-grade total-station measurements with high-end inertial sensors and dual RTK-GNSS, GrandTour achieves millimeter-level accuracy with sub-millisecond synchronization, enabling research in contact-aware perception, terrain-adaptive locomotion, and multi-modal perception that was previously limited by the available legged-robot datasets.

\section{System overview}\label{sec:system_overview}
\subsection{Sensor suite}\label{sec:sensors}

\begin{figure*}[t]
    \centering
    \includegraphics[width=.95\linewidth]{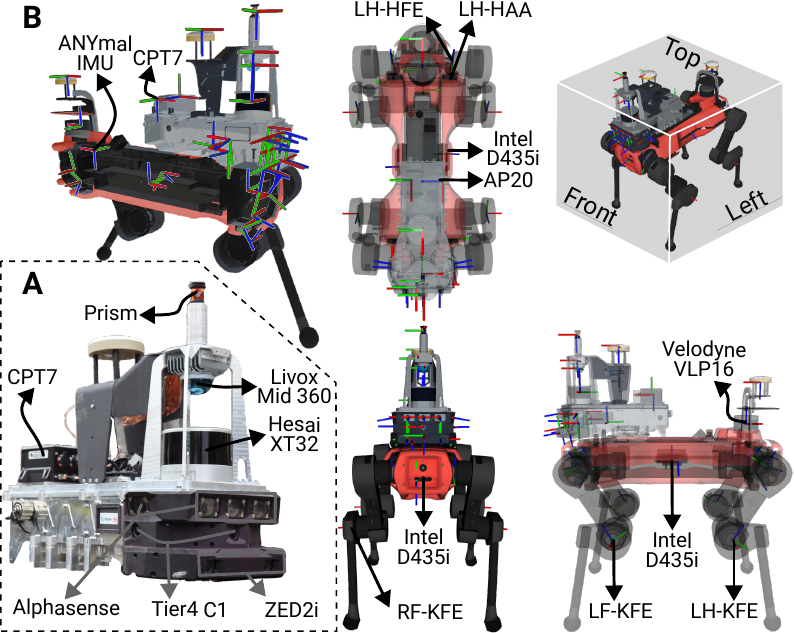}
    \caption{Sensor placement visualization of the entire GrandTour sensor suite. \textbf{A)} Shows the spatial relationships of the sensors on the \boxi payload, and \textbf{B)} shows the sensors of the ANYmal base platform components.
    Each colored axis represents a sensor or joint frame. \revised{The dataset contains seven depth-capable cameras overall: the \alias{ZED2i} stereo RGB-D camera and six ANYmal-mounted Intel RealSense D435i depth cameras.} For brevity, repeated sensors are not shown (e.g., 6$\times$ Intel RealSense D435i depth cameras).}\label{fig:side-by-side}
\end{figure*}

GrandTour uses the \boxi perception payload~\citep{frey2025boxi} in combination with the ANYmal quadruped robot.
All sensors inside \boxi and ANYmal are listed in \cref{tab:components}.
\boxi integrates exteroceptive and proprioceptive sensors into a compact, aluminum monoblock housing that is CNC machined to \SI{0.05}{\milli\metre} tolerance, providing excellent thermal stability and mechanical rigidity.
\boxi includes two high-performance rotating LiDARs (a CLICx Mid 360 (\alias{Livox}) for near field and a Hesai XT-32 (\alias{Hesai}) for long range geometric perception), together with ten RGB cameras (mixtures of high dynamic range, global shutter, and rolling shutter models), a ZED2i stereo RGB-D camera (\alias{ZED2i}), a Sevensense CoreResearch unit (\alias{CoreResearch}), and three TierIV C1 HDR cameras (\alias{HDR}), enabling wide field monocular, stereo, and multi-view imaging.
The dataset contains seven IMUs of varying quality: six from the \boxi sensor suite (\alias{HG4930}, \alias{STIM320}, \alias{ADIS}, \alias{AP20-IMU}, \alias{Bosch}, and \alias{TDK}) and one from the ANYmal-D quadruped (\alias{ANYmal-IMU}). 
The \alias{ZED2i} camera also contains an internal IMU, but this stream is not counted among the seven released dataset IMUs. 
The suite further includes \revised{proprioceptive encoders that measure joint positions, velocities, and torques}; \revised{the point foot contacts in the dataset are derived onboard-estimator outputs rather than direct measurements or ground truth}.
Finally, a NovAtel CPT7 dual-antenna RTK GNSS receiver with the Inertial Explorer software solution provides global pose estimates, whereas a Leica Geosystems custom AP20 solution enables real-time recording of the position of a Leica Mini Prism relative to the Leica Geosystems MS60 total station.
The entire sensor suite weighs about \SI{7.1}{\kilogram} and draws roughly \SI{120}{\watt}. For brevity, we will refer to each sensor by its acronym.\looseness-1

While \boxi itself already offers a versatile set of sensors, in addition to \boxi, we record sensor and actuator information from ANYmal, namely, six Intel D435i depth cameras (\alias{ANYmal-Depth}; in addition to the \alias{ZED2i} stereo RGB-D camera on \boxi), a Velodyne VLP16 LiDAR (\alias{VLP16}), an IMU (\alias{ANYmal-IMU}), and 12 joint-encoders (\alias{ANYmal-JE}), commanded velocity, and multiple intermediate processing outputs (elevation maps, state estimation, hidden states of locomotion policy).
All sensors are rigidly mounted on the robot with known extrinsic transforms and share a common timing source, utilizing different synchronization methods to ensure temporal alignment. Further details are provided in \cref{sec:calibration}, and the sensor positioning is illustrated in \cref{fig:side-by-side}.

\begin{figure*}
    \centering
    \includegraphics[width=0.9\linewidth]{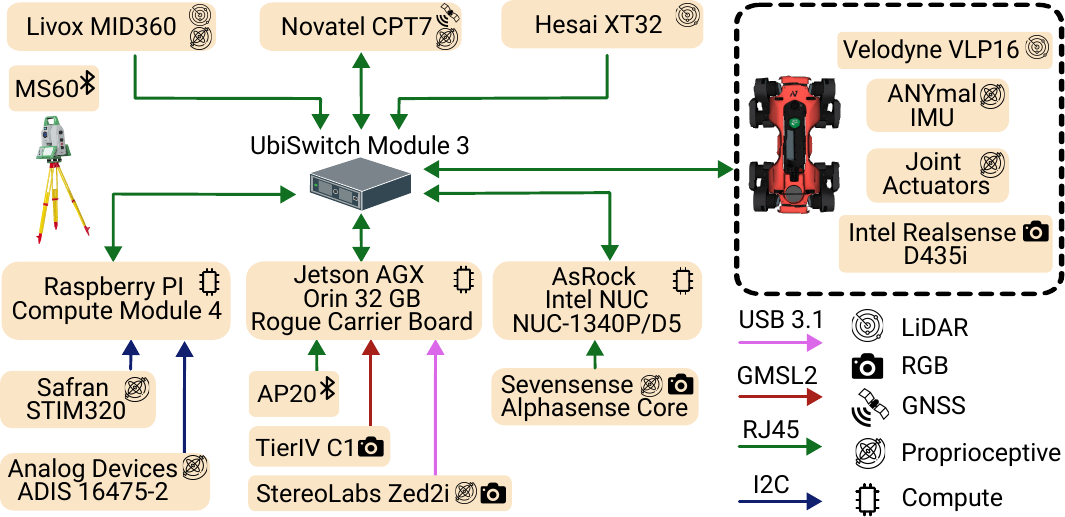}
    \caption{System architecture and sensor interfaces of the combined Boxi–ANYmal platform. All compute units (Jetson AGX Orin, Intel NUC, Raspberry Pi) are connected to the UbiSwitch Ethernet device, and the sensors are connected to their respective compute platforms via various interfaces, such as USB 3.1, GMSL2, RJ45, and I²C.}\label{fig:system_overview}
\end{figure*}

\begin{table*}[!t]
\centering
\begin{threeparttable}
\scriptsize
\setlength{\tabcolsep}{2pt}
\renewcommand{\arraystretch}{0.84}
\newcolumntype{L}[1]{>{\raggedright\arraybackslash}m{#1}}
\renewcommand{\tabularxcolumn}[1]{m{#1}}
\newcolumntype{Y}{>{\raggedright\arraybackslash}X}
\begin{tabularx}{\textwidth}{@{}L{3.0cm} L{2.35cm} Y L{2.35cm} L{1.65cm} L{2.45cm}@{}}
\toprule
\textbf{\revised{Alias}} & \textbf{Name} & \textbf{\revised{Specification}} & \textbf{\shortstack[l]{Time\\Synchronization$^\dagger$}} & \textbf{\revised{Power}} & \textbf{\revised{Description}} \\
\midrule
\multicolumn{6}{l}{\textbf{LiDARs}} \\
\aliastable{Livox} & Livox Mid-360 & \revised{vFoV: \SI{59}{\degree}, range: \SIrange{0.1}{40}{\meter}, \SI{10}{\hertz}, accuracy: $\pm$\SI{0.02}{\meter}} & \revised{IEEE 1588V2 PTP} & \revised{\SI{6.5}{\watt}} & \revised{Near-field LiDAR} \\
\rowcolor{black!5}
\aliastable{Hesai} & Hesai XT-32 & \revised{vFoV: \SI{31}{\degree}, range: \SIrange{0.05}{120}{\meter}, \SI{10}{\hertz}, accuracy: $\pm$\SI{0.01}{\meter}} & \revised{IEEE 1588V2 PTP} & \revised{\SI{10}{\watt}} & \revised{Long-range LiDAR} \\
\aliastable{VLP16} & Velodyne VLP-16 & \revised{vFoV: \SI{30}{\degree}, range: \SIrange{0.5}{100}{\meter}, \SI{10}{\hertz}, accuracy: $\pm$\SI{0.03}{\meter}} & IEEE 1588V2 PTP & \revised{\SI{8}{\watt} (typ)} & \revised{ANYmal-mounted LiDAR} \\
\midrule
\multicolumn{6}{l}{\textbf{Cameras}} \\
\rowcolor{black!5}
\aliastable{5$\times$ CoreResearch} & Sevensense CoreResearch & \revised{1440$\times$1080 @ \SI{10}{\fps}, FoV: \SI{126}{\degree}$\times$\SI{92.4}{\degree}, \SI{71.6}{\decibel}, GS, \SI{1.6}{\megapixel}} & IEEE 1588V2 PTP & \revised{\SI{12}{\watt}} & \revised{Global-shutter camera bundle} \\
\aliastable{3$\times$ HDR} & TierIV C1 & \revised{1920$\times$1280 @ \SI{30}{\fps}, FoV: \SI{120}{\degree}$\times$\SI{80.0}{\degree}, \SI{120}{\decibel}, RS, \SI{2.5}{\megapixel}} & \revised{GMSL2 interface} & \revised{\SI{1.7}{\watt}/ea.} & \revised{High-dynamic-range cameras} \\
\rowcolor{black!5}
\aliastable{ZED2i} & Stereolabs ZED2i & \revised{1920$\times$1080 @ \SI{15}{\fps}, FoV: \SI{110}{\degree}$\times$\SI{70}{\degree}, \SI{64.6}{\decibel}, RS, \SI{4}{\megapixel}} & \revised{USB buffer} & \revised{\SI{1.9}{\watt}} & \revised{Stereo RGB-D camera} \\
\aliastable{6$\times$ ANYmal-Depth} & Intel RealSense D435i & \revised{848$\times$480 @ \SI{15}{\fps}, FoV: \SI{87}{\degree}$\times$\SI{58}{\degree}, range: \SIrange{0.2}{10}{\meter}} & IEEE 1588V2 PTP & \revised{\SI{3.5}{\watt}/ea.} & \revised{ANYmal-mounted depth cameras} \\
\midrule
\multicolumn{6}{l}{\textbf{Reference}} \\
\rowcolor{black!5}
\aliastable{AP20}$^*$ & Modified Leica AP20 & \revised{MS60 total-station Bluetooth communication and synchronization} & \SI{1}{\milli\second}~(3$\sigma$) & \revised{\SI{4}{\watt} (est.)} & \revised{Prism tracking interface} \\
\aliastable{Prism} & Mini-Prism GRZ101 & \revised{Static accuracy: $\pm$\SI{2}{\milli\meter} (3$\sigma$), hFoV: \SI{360}{\degree}, vFoV: $\pm$\SI{30}{\degree}} & \revised{\na} & \revised{passive} & \revised{Total-station target} \\
\rowcolor{black!5}
\aliastable{CPT7} & \revised{NovAtel SPAN CPT7} & \revised{Dual-antenna RTK GNSS, Inertial Explorer, TerraStar-X} & IEEE 1588V2 PTP & \revised{\SI{18}{\watt} (peak)} & \revised{GNSS/INS reference} \\
\aliastable{2$\times$ GNSS Ant.} & \revised{Multiband GNSS Antenna} & \revised{NovAtel 2$\times$3GNSSA-XT-1} & \revised{\na} & \revised{\SI{1}{\watt} (each)} & \revised{CPT7 antennas} \\
\midrule
\multicolumn{6}{l}{\textbf{Proprioceptive}} \\
\rowcolor{black!5}
\aliastable{HG4930} & Honeywell HG4930 & \revised{\SI{100}{\hertz}, \SI[parse-numbers = false]{\pm400}{\degree\per\second}, \SI[parse-numbers = false]{\pm20}{\ggrav}, IMU of \aliastable{CPT7}} & \revised{HW trigger w.r.t. \aliastable{CPT7}} & \revised{\SI{2}{\watt}} & \revised{High-end MEMS IMU} \\
\aliastable{STIM320} & Safran STIM320 & \revised{\SI{500}{\hertz}, \SI[parse-numbers = false]{\pm400}{\degree\per\second}, \SI[parse-numbers = false]{\pm5}{\ggrav}} & \revised{Free-running kernel-stamped} & \revised{\SI{2.5}{\watt}} & \revised{Auxiliary IMU} \\
\rowcolor{black!5}
\aliastable{AP20-IMU} & Proprietary IMU & \SI{200}{\hertz} & \revised{HW trigger w.r.t. \aliastable{AP20}} & \revised{included} & \revised{AP20 internal IMU} \\
\aliastable{ADIS} & ADIS16475-2 & \revised{\SI{200}{\hertz}, \SI[parse-numbers = false]{\pm500}{\degree\per\second}, \SI[parse-numbers = false]{\pm8}{\ggrav}} & \revised{Free-running IIO-stamped} & \revised{\SI{0.20}{\watt}} & \revised{Auxiliary IMU} \\
\rowcolor{black!5}
\aliastable{Bosch} & Bosch BMI085 & \revised{\SI{200}{\hertz}, \SI[parse-numbers = false]{\pm2000}{\degree\per\second}, \SI[parse-numbers = false]{\pm16}{\ggrav}, IMU of \aliastable{CoreResearch}} & \revised{HW trigger w.r.t. \aliastable{CoreResearch}} & \revised{included} & \revised{Camera-suite IMU} \\
\aliastable{TDK} & TDK ICM40609 & \revised{\SI{200}{\hertz}, \SI[parse-numbers = false]{\pm2000}{\degree\per\second}, \SI[parse-numbers = false]{\pm32}{\ggrav}, IMU of \aliastable{Livox}} & \revised{HW trigger w.r.t. \aliastable{Livox}} & \revised{included} & \revised{LiDAR IMU} \\
\rowcolor{black!5}
\aliastable{ANYmal-IMU} & Proprietary IMU & \revised{\SI{400}{\hertz}, \SI[parse-numbers = false]{\pm1000}{\degree\per\second}, \SI[parse-numbers = false]{\pm8}{\ggrav}} & \revised{SW stamped w.r.t. \aliastable{VLP16}} & \revised{included} & \revised{ANYmal base IMU} \\
\aliastable{12$\times$ ANYmal-JE} & ANYmal-D Joint Encoders & \revised{12 joint positions, velocities, and torques} & \revised{SW trigger w.r.t. \aliastable{ANYmal-IMU}} & \revised{included} & \revised{Joint-encoder and actuator-state streams} \\
\bottomrule
\end{tabularx}
\begin{tablenotes}\scriptsize
\item[*] \aliastable{AP20} requires a Leica Geosystems MS60 Total Station.
\item[$\dagger$] Relative to the sensor host platform.
\end{tablenotes}
\caption{\revised{Specifications and settings of all sensors available in the GrandTour dataset.} GS=GlobalShutter, RS=RollingShutter, SW=Software, HW=Hardware.}
\label{tab:components}
\end{threeparttable}
\end{table*}

\subsection{Temporal and Spatial Calibration}\label{sec:calibration}
Precise intrinsic and extrinsic calibration is fundamental for effective multi-modal sensor fusion.
The calibration methodology for GrandTour builds upon the comprehensive procedures developed for the \boxi payload, which are thoroughly documented in \citep{frey2025boxi}.



\subsubsection{Time Synchronization}
Given the heterogeneous nature of the sensors, multi-source time synchronization methods were used, and we aimed to achieve a time synchronization formulation for the GrandTour dataset. The time-synchronization accuracy of the sensors ranges from \SI{}{\nano\second} for the IEEE1588v2 PTP protocol to \SI{}{\milli\second} for our custom software-based kernel timestamping solution, with the exception of the USB buffer-reliant \alias{ZED2i} RGB-D camera. 
\revised{Here, hardware synchronization refers to sensors being triggered at specific hardware clock times, or sensor-level timestamping, while software synchronization refers to host-side PTP and NTP alignment and kernel or driver timestamping.}

In short, the NovAtel SPAN CPT7 GNSS receiver serves as the time grandmaster, and all Precision Time Protocol (PTP)-enabled devices synchronize against this time source (Jetson, NUC, Livox, Hesai, and CoreResearch, as mentioned in \cref{tab:components}).
\removed{The Jetson within \boxi serves as a Network Time Protocol (NTP) time source for the ANYmal robot.}
\revised{With the Jetson within \boxi acting as the NTP server, NTP-based synchronization across the heterogeneous compute units is performed using \texttt{chrony}. 
This includes the ANYmal-D internal compute modules. 
We refer the reader to \boxi~\citep{frey2025boxi} and the public GrandTour repository\footnote{\url{https://github.com/leggedrobotics/grand_tour_box}} for the chrony and PTP configuration files.}
All sensors within ANYmal are guaranteed to be well time-synchronized to its internal compute module by the manufacturer. This compute module synchronizes itself with Jetson via NTP within \boxi.
All LiDARs are synchronized via the IEEE 1588v2 PTP protocol and therefore achieve significant time synchronization accuracy.
The released IMU streams are time-aligned and were verified with a custom-developed time alignment tool, ensuring time alignment sub-\SI{1}{\milli\second}, which is significantly lower than the period of every IMU stream\footnote{The \alias{STIM320} required \revised{a software-defined time alignment} based on angular velocities due to \revised{a} time-independent offset, introduced by the communication between the kernel and the driver.}.

\subsubsection{Timestamp and Exposure Time of Cameras}
The rolling shutter \alias{HDR} cameras are triggered and hardware-timestamped at the midpoint of the exposure time (fixed \SI{11}{\milli\second} exposure) of the middle row within the image (per-row exposure times can be computed).

The \alias{CoreResearch} global shutter camera is PTP synchronized, provides auto-exposure for all cameras, allows for exposure synchronization of the front-facing monocular stereo pair, and adjusts the exposure trigger signal for each camera such that the center of the exposure time aligns across all attached cameras and timestamps images at the middle of the exposure time. The \alias{ZED2i} camera is connected via a USB serial buffer; therefore, only unreliable timestamps are available and without individual line read-out timings. \revised{Users requiring precise rolling-shutter or exposure-midpoint semantics should use the per-sensor timing information and calibration notes provided with \boxi~\citep{frey2025boxi}.}

\subsubsection{Calibration of Boxi}
\paragraph{Calibration chain:}
As detailed in~\citep{frey2025boxi}, we accurately calibrate the intrinsic and extrinsic parameters of the \num{10} cameras into a single, metrically consistent bundle. This accurate camera bundle then enables all subsequent joint calibrations with the remaining sensors, including the prism, LiDARs, and IMUs.

\paragraph{Camera intrinsic and extrinsic calibration:}
For camera intrinsic and extrinsic calibration, we implemented a camera bundle calibrator inspired by Kalibr~\citep{furgale2013kalibr, rehder2016extending}, which enables simultaneous calibration and data collection with real-time feedback to guide the data collection process. This is done in two phases.
During the intrinsic calibration phase, the operator moves an AprilGrid~\citep{aprilgrid2011} throughout each camera’s field of view until sufficient observations with adequate image plane coverage are acquired to estimate camera intrinsics, as illustrated in Frey et al.~\citep{frey2025boxi}.
Given the moderate field of view of all camera lenses, we adopt the pinhole projection model. For lens distortion, we employ the Kannala–Brandt~\citep{kannalbrandt2006} equidistant model for the wide-angle fisheye \alias{CoreResearch} and \alias{HDR} cameras, and the rad-tan distortion model for the \alias{ZED2i} cameras.\looseness-1

In the subsequent extrinsic calibration phase, the calibration target is placed statically at multiple locations around \boxi, or the payload itself is reoriented to different static poses. The calibration program automatically detects these static intervals and uses only these observations to optimize the inter-camera extrinsic parameters. This restriction is necessary as the \num{10} camera shutters do not trigger simultaneously.
To guide the operator during data collection, the system periodically computes and reports the online covariance estimates of both intrinsic and extrinsic parameters, indicating when sufficient excitation has been achieved.

\paragraph{Camera to IMU calibration:}
First, we use the Allan variance tool~\citep{AllanVarianceRos} to calibrate the IMU intrinsics.
We then use Kalibr~\citep{furgale2013kalibr} to calibrate the extrinsics of all IMUs with respect to the front-facing global shutter \alias{CoreResearch} cameras.
We follow the best practices outlined in~\citep{furgale2013kalibr} for exposure time, camera frame rate, motion, and lighting. The specific parameters are detailed in~\citep{frey2025boxi}.


\paragraph{LiDAR to camera extrinsic calibration:}
To efficiently perform the LiDAR–camera extrinsic calibration, we employ a variant of the intensity-alignment method DiffCal~\citep{diffcal}.
In contrast to plane-based calibration, which requires numerous observations spanning diverse target orientations, the intensity-matching formulation is specifically data efficient and utilizes even partial observations of the calibration target within the LiDAR point cloud. 

\begin{figure}
    \centering
    \includegraphics[width=1\linewidth]{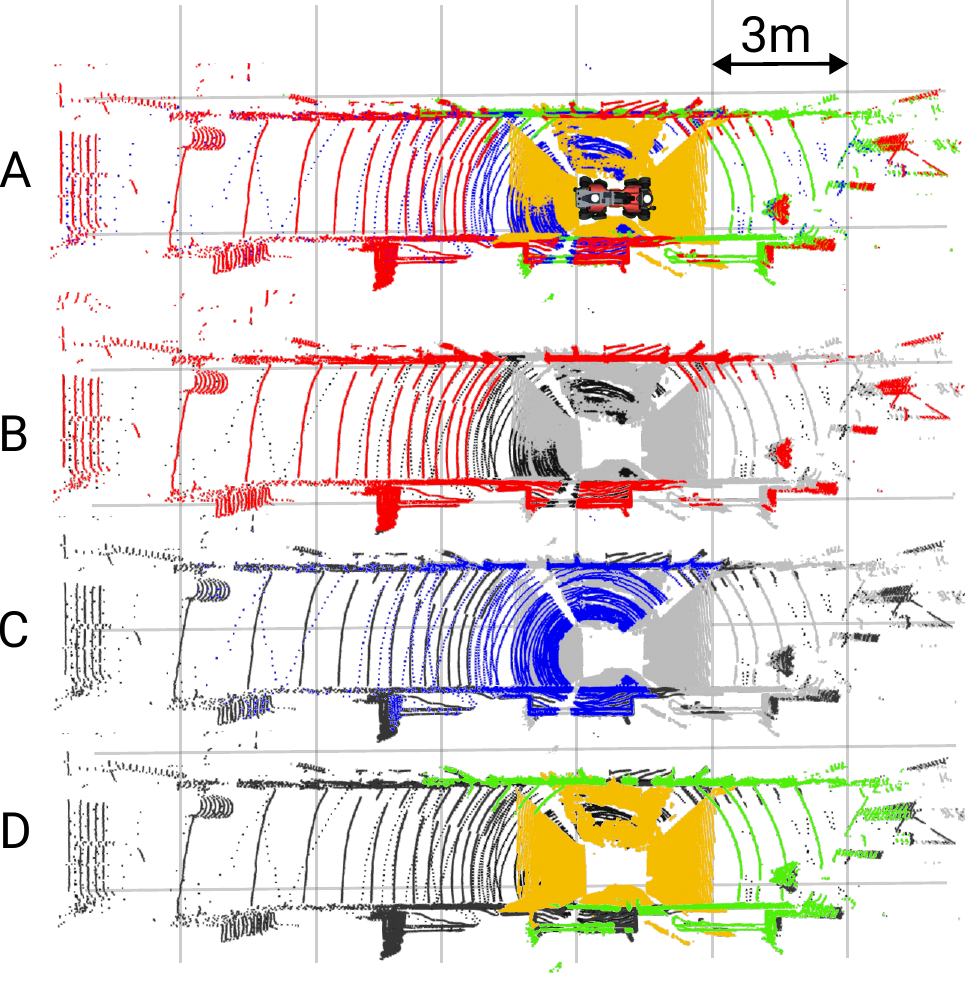}
    \caption{Top-down projection of range observations around the robot in GrandTour (scale bar: \SI{3}{\meter}), illustrating the complementary coverage of the sensor suite. \textbf{A)} Combined view with color-coding: \alias{Hesai} LiDAR (red), \alias{Livox} LiDAR (blue), \alias{VLP16} (green), and \alias{ANYmal-Depth} cameras (yellow). \textbf{B)--D)} For clarity, the sensor(s) of interest are shown in color while all other measurements are shown in gray. \textbf{B)} \alias{Hesai} LiDAR. \textbf{C)} \alias{Livox} LiDAR. \textbf{D)} ANYmal-mounted sensors: \alias{VLP16} and \alias{ANYmal-Depth}.}
    \label{fig:lidar_view}
\end{figure}

This property is particularly crucial for the \alias{Hesai} sensor, whose narrow field of view often captures only fragments of the calibration pattern. Supporting partial observations enables us to collect calibration data near the cameras, where the camera pose estimates are more accurate.
In contrast, the non-repeating scanning pattern of the \alias{Livox} provides dense coverage, enabling high-quality calibration even when the full target is not visible in each scan. On top of the LiDARs on \boxi, the \alias{VLP16} LiDAR and the six \alias{ANYmal-Depth} cameras provide further complementary observations. The complementary nature of the LiDAR and depth sensing in the GrandTour dataset is shown in \cref{fig:lidar_view}.\looseness-1

Each LiDAR on \boxi is individually calibrated against the five \alias{CoreResearch} cameras.
Using the calibrated extrinsic parameters between the cameras, the detected calibration target poses from each camera are fused and jointly calibrated with respect to each LiDAR, without refining the intra-camera extrinsic parameters.

\begin{figure*}
    \centering
    \includegraphics[width=0.99\linewidth]{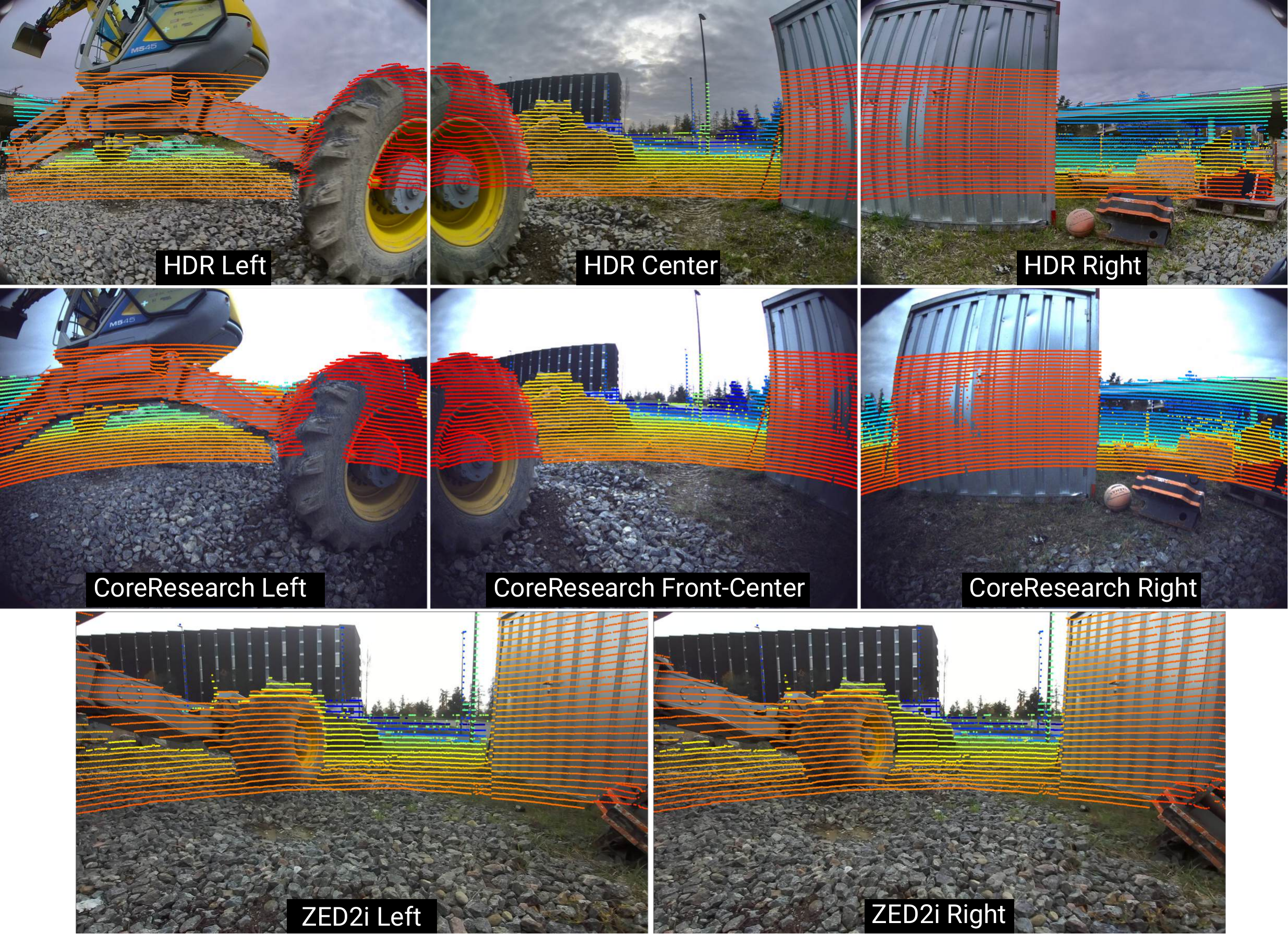}
    \caption{The calibration provided with the GrandTour dataset is validated through the overlay of point clouds from different LiDARs onto all available RGB images of GrandTour. As shown, the point clouds align with the correct visual features in the image. Points are colorized by their depth along the camera axis; red indicates closer points, and blue indicates farther points.}
    \label{fig:overlay_figure}
\end{figure*}

\paragraph{Camera-to-prism calibration}
Due to the lack of existing methods, we developed a custom calibration procedure to calibrate the front-facing \alias{CoreResearch} cameras to the Leica GRZ101 mini prism.
Using the MS60 total station position measurements and the detections of a static calibration target, we solve for a \num{9}-degree-of-freedom state: the prism position relative to the reference camera (\num{3}-degree-of-freedom), and the \num{6}-degree-of-freedom pose of the calibration target with respect to the total station.
To ensure accurate target detection, the camera is positioned within \SI{1}{\meter} of the calibration target.
\boxi is held statically at various poses around the target using a tripod while the total station tracks the prism.
Sufficient samples are collected to excite all translational and rotational degrees of freedom, ensuring the observability of the \num{9}-degree-of-freedom optimization state.
We validate the calibration by checking the consistency of prism estimates across all cameras, transforming them to the front-center camera frame using the known intra-camera extrinsic transform. The routine yields \SI{3}{\milli\meter} cross-camera consistency.
Furthermore, we validate the calibration accuracy by correlating real-world multi-modal data, as demonstrated in \cref{fig:overlay_figure}. Importantly, we have validated this overlay during a frame of motion. The exceptional alignment observed is the result of precise LiDAR motion compensation, combined with accurate camera–LiDAR extrinsics and camera intrinsics, which enable each motion-distorted LiDAR point to be correctly projected into the corresponding camera frame.

\subsubsection{Boxi-to-ANYmal calibration}
The \textit{Boxi} payload is mounted to ANYmal via a rigid, machined aluminum interface that utilizes precision locating pins and a preloaded bolt pattern, allowing for assembly until metal-to-metal contact is achieved.
The extrinsic transformation between the base of \boxi and the base of ANYmal is obtained from the CAD assembly and validated during integration by projecting the \alias{VLP16} point clouds into the camera bundle frames using the chained extrinsic transformations.
We have further verified the extrinsic calibration between the LiDARs of GrandTour by employing a map-to-map point cloud registration method under the locally no-drift assumption.

\begin{figure*}
    \centering
    \includegraphics[width=1\linewidth]{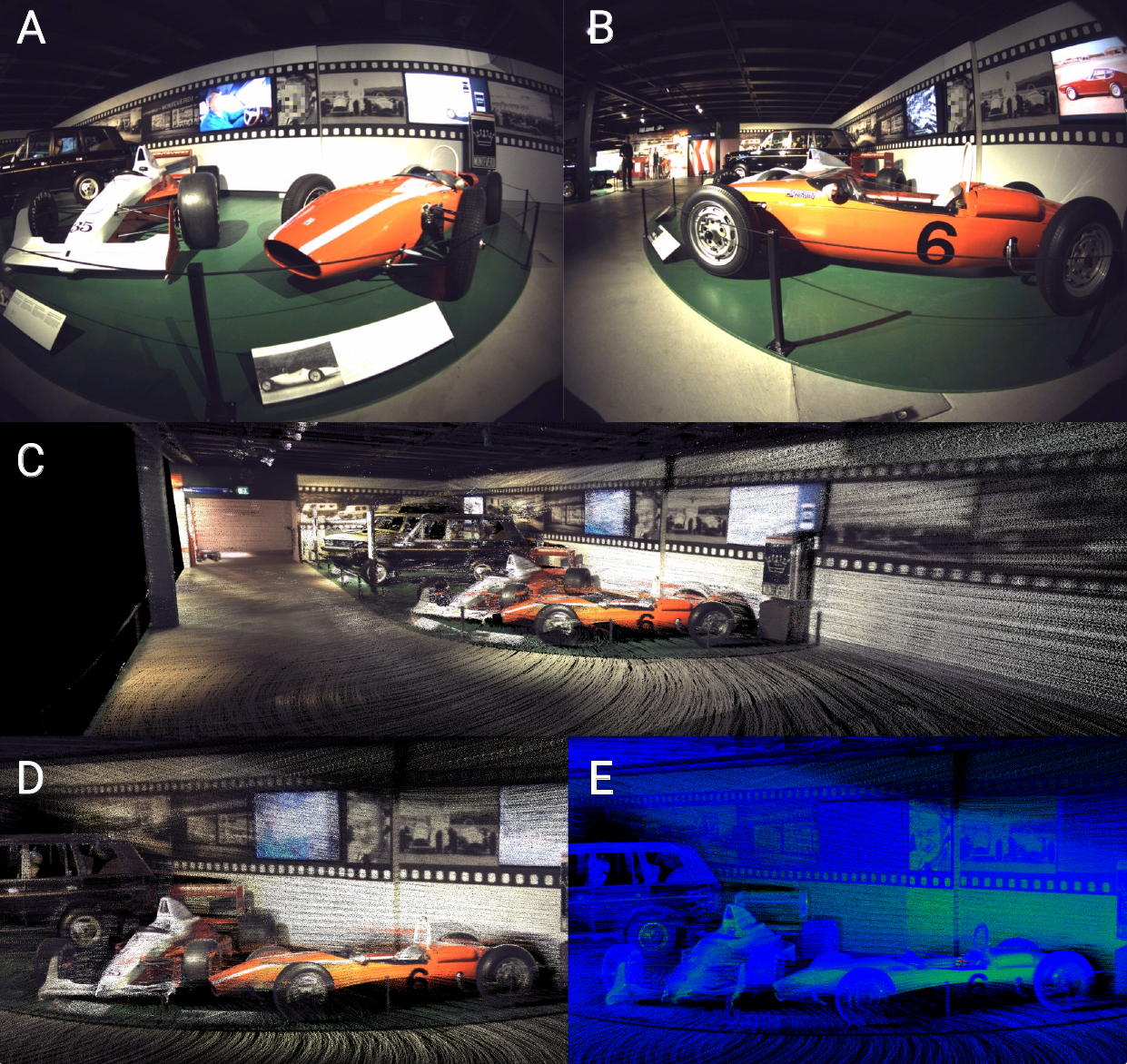}
    \caption{A colorized point cloud map of the HAUS-1 sequence is shown to highlight the accuracy of the sensor calibration. 
    {\textbf{A)} and \textbf{B)} show front-center color \alias{CoreResearch} camera images} from different moments during the sequence, whereas \textbf{C)} shows the colorized point cloud map generated by FAST-LIVO2~\citep{zheng2024fast} using \alias{Hesai} LiDAR and the front-center color \alias{CoreResearch} camera.
     \textbf{D)} shows a view rendered within the colorized map, and \textbf{E)} illustrates the same view on the intensity spectrum of \alias{Hesai} LiDAR.}
    \label{fig:colored_map}
\end{figure*}

\section{Dataset overview}\label{sec:dataset_overview}
The dataset comprises 49 distinct missions spread across a wide variety of natural and man-made environments.
Each mission spans \SI{3}{\minute} to \SI{7}{\minute} and covers a distance up to several hundred meters.
The complete list of missions is provided in \cref{tab:missions}, along with the tags for each mission and its statistics.

\begin{figure*}
    \centering
    \includegraphics[width=1.0\linewidth]{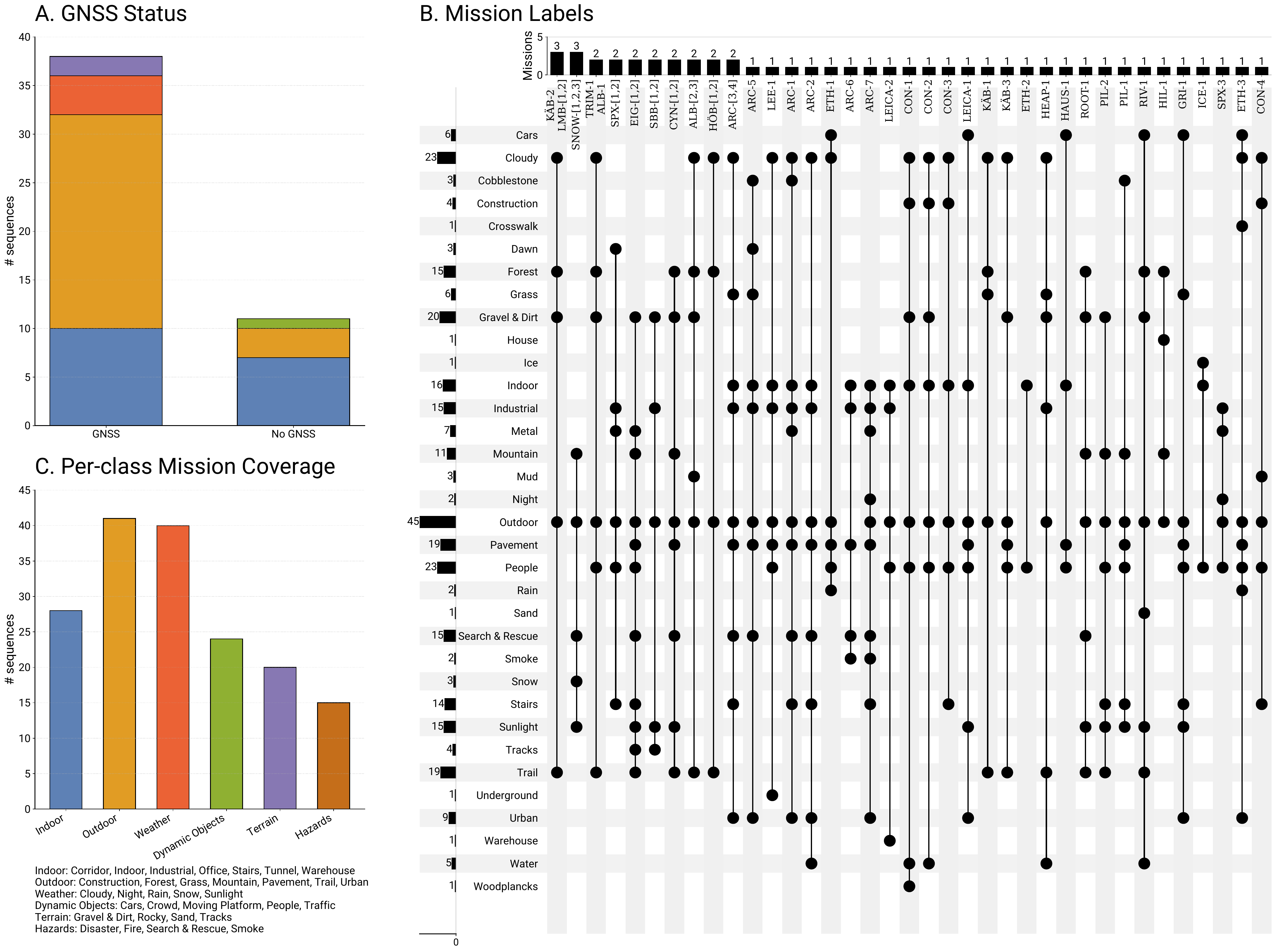}
    \caption{GNSS coverage and semantic diversity of the dataset. \textbf{A)} Number of sequences recorded with and without GNSS. \textbf{B)} The most common intersections of semantic labels across all missions, illustrating how weather, environment, terrain, and hazard conditions co-occur. \textbf{C)} Number of sequences per high-level label class, where each class aggregates the underlying labels listed below the plot. }
    \label{fig:label_statistics}
\end{figure*}

The recorded data exhibit a large variance in environmental factors, spanning day and night, sunshine, cloudy weather, and rain, as well as challenging terrains such as sand, snow, and gravel.
\cref{fig:label_statistics} summarizes the environments and other tags to facilitate dataset exploration.

Furthermore, the dataset webpage\footnote{\url{https://grand-tour.leggedrobotics.com}} offers a straightforward way to explore and access the data, featuring search functionality, preview videos for each mission, and a technical overview of the sensor setup.\looseness-1
In addition to raw sensor measurements, the dataset includes intermediate outputs such as robot pose and velocity estimates, motion-compensated point clouds, accumulated point cloud maps\footnote{\url{https://drive.google.com/drive/folders/1VV83LVWjfVmjx8WXG_QFYENjXSH2GN60?usp=sharing}}, and occupancy maps, as shown in \cref{sec:post_processed}. 

\begin{table*}[t]
  \footnotesize
  \centering
  \caption{Overview of all missions available in the GrandTour dataset.}
  \setlength{\tabcolsep}{3pt} 

  \begin{threeparttable}
    \resizebox{1.0\textwidth}{!}{%
      \rowcolors{1}{white}{gray!12}
  \begin{tabular}{p{1.3cm}p{2.6cm}p{1.4cm}p{1.6cm}p{1.9cm}p{1.0cm}p{9.7cm}}
    \toprule
    \textbf{Codename} & \textbf{Collection date} &
    \textbf{Duration [\si{\second}]} & \textbf{Walked distance [\si{\meter}]} & \textbf{MS60 coverage [\%]} & \textbf{GNSS} & \textbf{Tags} \\ \midrule
    \href{https://www.youtube.com/watch?v=aoCQftD5ePs}{ETH-1} & 2024-10-01 11:29:55 & 333 & 204.729 & 74.934 & \href{https://grand-tour.leggedrobotics.com/GNSS/2024-10-01-11-29-55_gps_map_tc.html}{Yes}   & Outdoor, Pavement, Cars, People, Rain, Cloudy \\
    \href{https://www.youtube.com/watch?v=3NZY87NuDeA}{ETH-2} & 2024-10-01 11:47:44 & 418 & 210.591 & 67.846 & No    & Indoor, People, Landmarks, Dynamic Objects \\
    \href{https://www.youtube.com/watch?v=fM6MIrsBvc8}{ETH-3} & 2024-10-01 12:00:49 & 464 & 214.495 & 6.519 & \href{https://grand-tour.leggedrobotics.com/GNSS/2024-10-01-12-00-49_gps_map_tc.html}{Yes}   & Urban, Outdoor, Pavement, Crosswalk, Cars, People, Rain, Cloudy \\
    \href{https://www.youtube.com/watch?v=fma-a3Milfc}{SPX-1} & 2024-11-02 17:10:25 & 374 & 147.114 & 75.748 & \href{https://grand-tour.leggedrobotics.com/GNSS/2024-11-02-17-10-25_gps_map_tc.html}{Yes}   & Industrial, Outdoor, Metal, Stairs, People, Dawn \\
    \href{https://www.youtube.com/watch?v=_LjIgArYKcM}{SPX-2} & 2024-11-02 17:18:32 & 383 & 184.128 & 68.168 & \href{https://grand-tour.leggedrobotics.com/GNSS/2024-11-02-17-18-32_gps_map_tc.html}{Yes}   & Industrial, Outdoor, Metal, Stairs, People, Dawn \\
    \href{https://www.youtube.com/watch?v=sOY5CQE7aiQ}{SPX-3} & 2024-11-02 17:43:10 & 213 & 101.313 & 82.302 & \href{https://grand-tour.leggedrobotics.com/GNSS/2024-11-02-17-43-10_gps_map_tc.html}{Yes}   & Industrial, Outdoor, Metal, People, Night \\
    \href{https://www.youtube.com/watch?v=_QkLKjqOE8Y}{ICE-1} & 2024-11-02 21:12:51 & 251 & 56.253  & 91.71 & \href{https://grand-tour.leggedrobotics.com/GNSS/2024-11-02-21-12-51_gps_map_tc.html}{Yes}   & Indoor, Ice, People \\
    \href{https://www.youtube.com/watch?v=jbkuw52fyWM}{SNOW-1} & 2024-11-03 07:52:45 & 172 & 36.533  & 47.39 & \href{https://grand-tour.leggedrobotics.com/GNSS/2024-11-03-07-52-45_gps_map_tc.html}{Yes}   & Search~\&~Rescue, Mountain, Outdoor, Snow, Sunlight \\
    \href{https://www.youtube.com/watch?v=CT0gSpFxBHA}{SNOW-2} & 2024-11-03 07:57:34 & 261 & 173.927 & 99.433 & \href{https://grand-tour.leggedrobotics.com/GNSS/2024-11-03-07-57-34_gps_map_tc.html}{Yes}   & Search~\&~Rescue, Mountain, Outdoor, Snow, Sunlight \\
    \href{https://www.youtube.com/watch?v=YgrBb3jAYWs}{SNOW-3} & 2024-11-03 08:17:23 & 236 & 145.253 & 99.872 & \href{https://grand-tour.leggedrobotics.com/GNSS/2024-11-03-08-17-23_gps_map_tc.html}{Yes}   & Search~\&~Rescue, Mountain, Outdoor, Snow, Sunlight \\
    \href{https://www.youtube.com/watch?v=X10J2XtXNjA}{EIG-1} & 2024-11-03 13:51:43 & 429 & 219.702 & 70.360 & \href{https://grand-tour.leggedrobotics.com/GNSS/2024-11-03-13-51-43_gps_map_tc.html}{Yes}   & Search~\&~Rescue, Mountain, Outdoor, Gravel/Dirt, Trail, Pavement, Metal, Stairs, People, Tracks, Sunlight \\
    \href{https://www.youtube.com/watch?v=psXDrqaslTw}{EIG-2} & 2024-11-03 13:59:54 & 348 & 198.286 & 72.319 & \href{https://grand-tour.leggedrobotics.com/GNSS/2024-11-03-13-59-54_gps_map_tc.html}{Yes}   & Search~\&~Rescue, Mountain, Outdoor, Gravel/Dirt, Trail, Pavement, Metal, Stairs, People, Tracks, Sunlight \\
    \href{https://www.youtube.com/watch?v=KlNfOUBtasg}{GRI-1} & 2024-11-04 10:57:34 & 455 & 266.747 & 58.072 & \href{https://grand-tour.leggedrobotics.com/GNSS/2024-11-04-10-57-34_gps_map_tc.html}{Yes}   & Urban, Outdoor, Grass, Pavement, Stairs, Cars, People, Sunlight \\
    \href{https://www.youtube.com/watch?v=i55nPlnssLA}{CYN-1} & 2024-11-04 12:55:59 & 510 & 297.702 & 23.971 & \href{https://grand-tour.leggedrobotics.com/GNSS/2024-11-04-12-55-59_gps_map_tc.html}{Yes}   & Search~\&~Rescue, Forest, Mountain, Outdoor, Gravel/Dirt, Trail, Pavement, Sunlight \\
    \href{https://www.youtube.com/watch?v=wqf2mx1NZHM}{CYN-2} & 2024-11-04 13:07:13 & 314 & 164.594 & 49.308 & \href{https://grand-tour.leggedrobotics.com/GNSS/2024-11-04-13-07-13_gps_map_tc.html}{Yes}   & Search~\&~Rescue, Forest, Mountain, Outdoor, Gravel/Dirt, Trail, Pavement, Sunlight \\
    \href{https://www.youtube.com/watch?v=pSkkWlvYHO8}{HIL-1} & 2024-11-04 16:05:00 & 430 & 326.08  & 72.191 & \href{https://grand-tour.leggedrobotics.com/GNSS/2024-11-04-16-05-00_gps_map_tc.html}{Yes}   & Forest, Outdoor, Mountain, House \\
    \href{https://www.youtube.com/watch?v=hj-QvZ-QNLA}{RIV-1} & 2024-11-04 16:52:38 & 537 & 204.016 & 64.089 & No    & Forest, Water, Outdoor, Gravel/Dirt, Sand, Trail, Cars, Sunlight \\
    \href{https://www.youtube.com/watch?v=g2ogmih2KK0}{PIL-1} & 2024-11-11 12:07:40 & 946 & 470.525 & 21.009 & \href{https://grand-tour.leggedrobotics.com/GNSS/2024-11-11-12-07-40_gps_map_tc.html}{Yes}   & Mountain, Outdoor, Pavement, Cobblestone, Stairs, People, Sunlight \\
    \href{https://www.youtube.com/watch?v=9_tcRDLLItU}{PIL-2} & 2024-11-11 12:42:47 & 409 & 273.567 & 98.661 & \href{https://grand-tour.leggedrobotics.com/GNSS/2024-11-11-12-42-47_gps_map_tc.html}{Yes}   & Mountain, Outdoor, Gravel/Dirt, Trail, Stairs, People, Sunlight \\
    \href{https://www.youtube.com/watch?v=b9Qpuo-Qhgs}{ROOT-1} & 2024-11-11 14:29:44 & 340 & 177.51  & 69.225 & \href{https://grand-tour.leggedrobotics.com/GNSS/2024-11-11-14-29-44_gps_map_tc.html}{Yes}   & Search~\&~Rescue, Forest, Mountain, Outdoor, Gravel/Dirt, Trail, Sunlight \\
    \href{https://www.youtube.com/watch?v=J46tuIPv_DY}{HAUS-1} & 2024-11-11 16:14:23 & 337 & 170.493 & 66.008 & No    & Indoor, Pavement, Cars, People \\
    \href{https://www.youtube.com/watch?v=RDo_Fv6QZ64}{H\"{O}B-1} & 2024-11-14 11:17:02 & 517 & 289.299 & 39.166 & \href{https://grand-tour.leggedrobotics.com/GNSS/2024-11-14-11-17-02_gps_map_tc.html}{Partially} & Forest, Outdoor, Trail, Cloudy \\
    \href{https://www.youtube.com/watch?v=x2DgDkivhRs}{H\"{O}B-2} & 2024-11-14 12:01:26 & 445 & 197.271 & 27.944 & \href{https://grand-tour.leggedrobotics.com/GNSS/2024-11-14-12-01-26_gps_map_tc.html}{Partially} & Forest, Outdoor, Trail, Cloudy \\
    \href{https://www.youtube.com/watch?v=A6d_wqaMPNs}{HEAP-1} & 2024-11-14 13:45:37 & 290 & 162.041 & 86.737 & \href{https://grand-tour.leggedrobotics.com/GNSS/2024-11-14-13-45-37_gps_map_tc.html}{Yes}   & Water, Industrial, Outdoor, Gravel/Dirt, Grass, Trail, Cloudy \\
    \href{https://www.youtube.com/watch?v=Xpjis5jAcu4}{K\"{A}B-1} & 2024-11-14 14:36:02 & 465 & 205.374 & 79.346 & \href{https://grand-tour.leggedrobotics.com/GNSS/2024-11-14-14-36-02_gps_map_tc.html}{Yes}   & Forest, Outdoor, Grass, Trail, Cloudy \\
    \href{https://www.youtube.com/watch?v=Ok1p75XubHQ}{K\"{A}B-2} & 2024-11-14 15:22:43 & 383 & 185.596 & 41.099 & \href{https://grand-tour.leggedrobotics.com/GNSS/2024-11-14-15-22-43_gps_map_tc.html}{Yes}   & Forest, Outdoor, Gravel/Dirt, Trail, Cloudy \\
    \href{https://www.youtube.com/watch?v=eanojOiY7Lc}{K\"{A}B-3} & 2024-11-14 16:04:09 & 475 & 364.575 & 48.3 & \href{https://grand-tour.leggedrobotics.com/GNSS/2024-11-14-16-04-09_gps_map_tc.html}{Yes}   & Outdoor, Gravel/Dirt, Trail, Pavement, People, Cloudy \\
    \href{https://www.youtube.com/watch?v=AUoIWbHRX6k}{TRIM-1} & 2024-11-15 10:16:35 & 465 & 267.811 & 72.489 & \href{https://grand-tour.leggedrobotics.com/GNSS/2024-11-15-10-16-35_gps_map_tc.html}{Yes}   & Forest, Outdoor, Gravel/Dirt, Trail, People, Cloudy \\
    \href{https://www.youtube.com/watch?v=5ah7M4KnvdM}{ALB-1} & 2024-11-15 11:18:14 & 282 & 107.976 & 41.31 & \href{https://grand-tour.leggedrobotics.com/GNSS/2024-11-15-11-18-14_gps_map_tc.html}{Yes}   & Forest, Outdoor, Gravel/Dirt, Trail, People, Cloudy \\
    \href{https://www.youtube.com/watch?v=6bYjtJ1URTY}{ALB-2} & 2024-11-15 11:37:15 & 508 & 206.572 & 65.356 & \href{https://grand-tour.leggedrobotics.com/GNSS/2024-11-15-11-37-15_gps_map_tc.html}{Yes}   & Forest, Outdoor, Gravel/Dirt, Mud, Trail, Cloudy \\
    \href{https://www.youtube.com/watch?v=dwvfxWY911w}{ALB-3} & 2024-11-15 12:06:03 & 209 & 77.971  & 88.959 & \href{https://grand-tour.leggedrobotics.com/GNSS/2024-11-15-12-06-03_gps_map_tc.html}{Yes}   & Forest, Outdoor, Gravel/Dirt, Mud, Trail, Cloudy \\
    \href{https://www.youtube.com/watch?v=5p9ps8OAUFE}{LMB-1} & 2024-11-15 14:14:12 & 528 & 204.999 & 58.058 & \href{https://grand-tour.leggedrobotics.com/GNSS/2024-11-15-14-14-12_gps_map_tc.html}{Yes}   & Forest, Outdoor, Grass, Trail, Cloudy \\
    \href{https://www.youtube.com/watch?v=L_bd1E7Z4Nk}{LMB-2} & 2024-11-15 14:43:52 & 244 & 100.881 & 18.822 & \href{https://grand-tour.leggedrobotics.com/GNSS/2024-11-15-14-43-52_gps_map_tc.html}{Yes}   & Forest, Outdoor, Gravel/Dirt, Trail, Cloudy \\
    \href{https://www.youtube.com/watch?v=l76355shpO0}{LEE-1} & 2024-11-15 16:41:14 & 529 & 331.839 & 71.635 & \href{https://grand-tour.leggedrobotics.com/GNSS/2024-11-15-16-41-14_gps_map_tc.html}{Partially} & Underground, Industrial, Indoor, Outdoor, Pavement, People, Cloudy \\
    \href{https://www.youtube.com/watch?v=43DW3_lIs24}{ARC-1} & 2024-11-18 12:05:01 & 433 & 247.878 & 71.266 & \href{https://grand-tour.leggedrobotics.com/GNSS/2024-11-18-12-05-01_gps_map_tc.html}{Yes}   & Search~\&~Rescue, Industrial, Urban, Indoor, Outdoor, Pavement, Cobblestone, Metal, Stairs, Cloudy \\
    \href{https://www.youtube.com/watch?v=qaP4F_nthMg}{ARC-2} & 2024-11-18 13:22:14 & 424 & 136.222 & 27.346 & \href{https://grand-tour.leggedrobotics.com/GNSS/2024-11-18-13-22-14_gps_map_tc.html}{Yes}   & Search~\&~Rescue, Water, Industrial, Urban, Indoor, Outdoor, Pavement, Stairs, Cloudy \\
    \href{https://www.youtube.com/watch?v=3tzX04q41hM}{ARC-3} & 2024-11-18 13:48:19 & 713 & 331.974 & 27.467 & \href{https://grand-tour.leggedrobotics.com/GNSS/2024-11-18-13-48-19_gps_map_tc.html}{Partially} & Search~\&~Rescue, Industrial, Urban, Indoor, Outdoor, Grass, Pavement, Stairs, Cloudy \\
    \href{https://www.youtube.com/watch?v=jqHJV8G53Ik}{ARC-4} & 2024-11-18 15:46:05 & 397 & 160.04  & 20.256 & \href{https://grand-tour.leggedrobotics.com/GNSS/2024-11-18-15-46-05_gps_map_tc.html}{Yes}   & Search~\&~Rescue, Industrial, Urban, Indoor, Outdoor, Grass, Pavement, Stairs, Cloudy \\
    \href{https://www.youtube.com/watch?v=dryOQ-MxZ8Y}{ARC-5} & 2024-11-18 16:59:23 & 391 & 173.303 & 42.052 & \href{https://grand-tour.leggedrobotics.com/GNSS/2024-11-18-16-59-23_gps_map_tc.html}{Yes}   & Search~\&~Rescue, Industrial, Urban, Indoor, Outdoor, Grass, Pavement, Cobblestone, Dawn \\
    \href{https://www.youtube.com/watch?v=mzRlUN4QA0Q}{ARC-6} & 2024-11-18 17:13:09 & 349 & 103.493 & 62.913 & No    & Search~\&~Rescue, Industrial, Indoor, Smoke, Pavement \\
    \href{https://www.youtube.com/watch?v=pv5KiXmnsyQ}{ARC-7} & 2024-11-18 17:31:36 & 322 & 157.244 & 35.778 & No   & Search~\&~Rescue, Industrial, Urban, Indoor, Outdoor, Smoke, Pavement, Metal, Stairs, Night \\
    \href{https://www.youtube.com/watch?v=laJDPLRxRfg}{LEICA-1} & 2024-11-25 14:57:08 & 310 & 152.656 & 80.195 & \href{https://grand-tour.leggedrobotics.com/GNSS/2024-11-25-14-57-08_gps_map_tc.html}{Yes}   & Urban, Indoor, Outdoor, Pavement, Cars, People, Sunlight \\
    \href{https://www.youtube.com/watch?v=DwQfcX0lSSY}{LEICA-2} & 2024-11-25 16:36:19 & 422 & 208.171 & 46.744 & \href{https://grand-tour.leggedrobotics.com/GNSS/2024-11-25-16-36-19_gps_map_tc.html}{Yes}   & Indoor~\&~Outdoor, Warehouse, People, Industrial \\
    \href{https://www.youtube.com/watch?v=A5clVaKHU-o}{SBB-1} & 2024-12-03 13:15:38 & 502 & 311.25  & 99.492 & \href{https://grand-tour.leggedrobotics.com/GNSS/2024-12-03-13-15-38_gps_map_tc.html}{Yes}   & Industrial, Outdoor, Gravel/Dirt, Tracks, Sunlight \\
    \href{https://www.youtube.com/watch?v=QH0qVZl0WWU}{SBB-2} & 2024-12-03 13:26:40 & 271 & 149.167 & 73.16 & \href{https://grand-tour.leggedrobotics.com/GNSS/2024-12-03-13-26-40_gps_map_tc.html}{Yes}   & Industrial, Outdoor, Gravel/Dirt, Tracks, Sunlight \\
    \href{https://www.youtube.com/watch?v=wvD5m-R7hwM}{CON-1} & 2024-12-09 09:34:43 & 388 & 263.267 & 75.999 & No & Water, Construction, Indoor, Outdoor, Gravel/Dirt, Woodplanks, People, Cloudy \\
    \href{https://www.youtube.com/watch?v=EERhlxdeOaM}{CON-2} & 2024-12-09 09:41:46 & 384 & 266.356 & 61.221 & No & Water, Construction, Indoor, Outdoor, Gravel/Dirt, People, Cloudy \\
    \href{https://www.youtube.com/watch?v=fcLAG9y0HXg}{CON-3} & 2024-12-09 11:28:28 & 796 & 188.934 & 23.632 & No & Construction, Indoor, Outdoor, Stairs, People, Cloudy \\
    \href{https://www.youtube.com/watch?v=Np2ttsGSRJA}{CON-4} & 2024-12-09 11:53:11 & 743 & 468.315 & 67.106 & No & Construction, Outdoor, Mud, Stairs, People, Cloudy \\
    \bottomrule
  \end{tabular}%
} 
    
  \end{threeparttable}
    \label{tab:missions}
\end{table*}

\subsection{Data format and storage}
The GrandTour dataset is available in two primary formats to maximize user accessibility and compatibility with existing workflows.

\begin{figure}[t]
    \centering
    \includegraphics[width=0.95\linewidth]{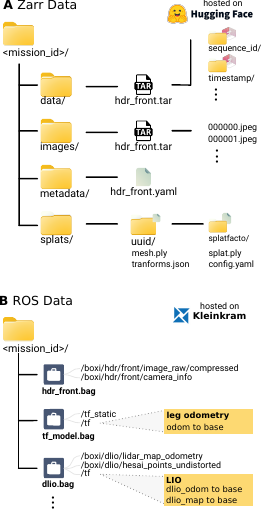}
    \caption{
    Data format overview. \textbf{A}) Zarr layout on HuggingFace. Topics are stored as multi-arrays; for instance, the \texttt{hdr\_front} contains a timestamp and sequence\_id. For images, the sequence\_id can be used to retrieve the corresponding \texttt{.jpeg} file. Metadata, including intrinsics, extrinsics, and descriptions, is provided both within the Zarr attributes and the YAML file in the metadata folder. Post-processed assets (Gaussian Splats, meshes) are stored in the \texttt{splats} folder. Only (depth)-image topics are stored as \texttt{.jpeg} or \texttt{.png} for compression. \textbf{B}) ROS data is hosted on kleinkram. Topic namespaces distinguish between \textit{Boxi} and \textit{ANYmal} platforms. Sensor intrinsics are embedded in messages, while extrinsics reside in \texttt{tf\_model.bag}. The derived outputs are separated into distinct bag files.}

    \label{fig:data_format}
\end{figure}

First, to ensure broader accessibility beyond the robotics community, we provide the dataset on HuggingFaceHub\footnote{\url{https://huggingface.co/datasets/leggedrobotics/grand_tour_dataset}} in both \textit{Zarr} and JPEG formats.
Alongside entire missions, we allow downloading individual missions and subsets of sensor information for ease of use.
For each sensor and mission, the full extrinsic and intrinsic calibration is available within the \textit{.zarr} files.
This format effectively decouples the data from the ROS ecosystem, allowing computer vision researchers, machine learning practitioners, and others without robotics expertise to easily access and process the dataset using familiar tools and libraries.
We provide a variety of minimalist Python scripts and notebooks to showcase loading, synchronization, and the use of different sensor streams.
The storage hierarchy in the HuggingFace platform is provided in \cref{fig:data_format}. The data can be easily fetched from HuggingFace through the Python API.\looseness-1
\begin{lstlisting}[language=Python]
from huggingface_hub import snapshot_download
hugging_face_data_cache_path = snapshot_download(
    "leggedrobotics/grand_tour_dataset",
    allow_patterns="*hdr_front*",
    repo_type="dataset"
)
\end{lstlisting}

Second, to facilitate rapid usage for robotics purposes and ensure backward compatibility, the dataset is available in ROS Bag (\texttt{.bag}) format.
The mission is structured into multiple bag files in a minimalist fashion, with each file corresponding to a specific stream of messages (e.g., LiDAR, IMU, image) for a particular mission.
This formatting decision allows users to download only the data they need, compared to substantially large \texttt{.bag} files that contain multiple sensor streams in an uncompressed format.

Furthermore, the bag files are compressed using the LZ4 lossless compression algorithm to reduce disk space usage.
In addition, as ROS 1 reached end-of-life, the GrandTour dataset is accompanied by type-safe conversion scripts\footnote{\url{https://github.com/leggedrobotics/grand_tour_box/blob/main/box_utils/box_auto/python/box_auto/scripts/special/generizableROS2conversion.py}} that allow users to automatically convert the \texttt{.bag} files to CRF compressed \texttt{.mcap} files, which are seamlessly interpretable by the ROS 2 ecosystem.
In this regard, we provide compatibility guarantees for our converted data with the LTS version of ROS 2 Jazzy and Ubuntu 24.04, utilizing a Docker environment.

Each mission is divided into 34 individual ROS bags, from which users can choose. Users can inspect the contents of the bag files through the Kleinkram GUI web interface.
These ROS-based formats preserve the original metadata from the deployment and can be integrated seamlessly with existing robotics software stacks.

The bag files are stored on ETH Zurich-based data servers and accessed through our in-house-built data tooling, called Kleinkram\footnote{\url{https://github.com/leggedrobotics/kleinkram}}~\citep{kleinkram}, a lightweight storage solution.
Kleinkram is developed by the Robotic Systems Lab and offers both a web and a command-line interface.
The Kleinkram web interface provides a sortable list of all missions, along with the convenience of a web GUI, while the Kleinkram CLI (available via the Python package index\footnote{\url{https://pypi.org/project/kleinkram/}}) is recommended for systematic downloads and scripting. Kleinkram CLI can be easily installed with
\begin{lstlisting}[language=sh]
pip3 install kleinkram
\end{lstlisting}
This tool allows users to retrieve files or missions by universally unique identifiers (UUIDs) and supports downloading \& uploading multiple files in a single command with multiple options for additional features. An example is provided below.
\begin{lstlisting}[language=sh]
klein download --project GrandTourDataset
               --mission 2024-10-01-12-00-49 --dest /tmp 'hdr_front'
\end{lstlisting}

\subsection{Ground Truth Pose Generation}
\paragraph{Satellite Reference}

For outdoor experiments, we obtain ground-truth poses from a NovAtel SPAN CPT7 inertial navigation system (INS), a dual-antenna GNSS receiver with the integrated \alias{HG4930} Honeywell high-end MEMS IMU.
The CPT7 provides real-time pose estimates using the TerraStar-C PRO\footnote{\url{https://hexagon.com/company/divisions/autonomy-and-positioning/rtk-from-sky}} multi-constellation correction service for Precise Point Positioning (PPP), and refined offline trajectories using Inertial Explorer\footnote{\url{https://novatel.com/products/waypoint-post-processing-software/}}, a GNSS/INS post-processing software by Hexagon\:|\:NovAtel Inc., in combination with the HxGN SmartNet correction service.\footnote{\url{https://hxgnsmartnet.com/}}

Inertial Explorer employs a Kalman filter that jointly estimates the GNSS states (position, velocity, clock offset, and carrier-phase ambiguities) and the platform states (pose, velocity), along with gyroscope and accelerometer biases.

The two filters are tightly integrated: GNSS-derived measurements bound the drift of the inertial solution, while the INS provides position and velocity estimates that stabilize and constrain the GNSS solution under challenging conditions, mitigating multipath effects and signal outages.
In addition to the platform pose, Inertial Explorer outputs linear velocity, IMU bias estimates, associated covariance information, and quality metrics.\looseness-1

We use Inertial Explorer's tightly coupled processing mode. In this mode, GNSS and inertial data are fused at the measurement level, enhancing satellite signal utilization under challenging GNSS conditions and mitigating inertial error growth even without a full GNSS position update. NovAtel recommends the tightly coupled solution due to its higher accuracy and faster convergence.

According to the CPT7 performance specifications of Inertial Explorer\footnote{\url{https://hexagondownloads.blob.core.windows.net/public/Novatel/assets/Documents/Papers/SPAN-CPT7-PS/SPAN-CPT7-PS.pdf}}, GNSS outages of up to \SI{10}{\second} result in position RMS errors in the range of \SIrange{0.01}{0.02}{\meter}, velocity RMS below \SI{0.015}{\meter\per\second}, and attitude RMS below \SI{0.01}{\degree}.

\begin{table}[t]
\centering
\caption{Performance characteristics of Inertial Explorer. All values are within $2\sigma$ variation.}\label{tab:ie}
\resizebox{1\columnwidth}{!}{
\begin{tabular}{
  c
  S[table-format=1.3]
  S[table-format=1.3]
  S[table-format=1.3]
  S[table-format=1.3]
  S[table-format=1.3]
  S[table-format=1.3]
}
\toprule
{GNSS Outage [\si{\second}]} &
\multicolumn{2}{c}{Position RMS[\si{\metre}]} &
\multicolumn{2}{c}{Velocity RMS[\si{\metre\per\second}]} &
\multicolumn{2}{c}{Attitude RMS[\si{\degree}]} \\
\cmidrule(lr){2-3} \cmidrule(lr){4-5} \cmidrule(lr){6-7}
& {Horiz.} & {Vert.} & {Horiz.} & {Vert.} & {Roll/Pitch} & {Heading} \\
\midrule
\SI{0}{\second}  & 0.01 & 0.02 & 0.015 & 0.01 & 0.003 & 0.01 \\
\SI{10}{\second} & 0.01 & 0.02 & 0.015 & 0.01 & 0.003 & 0.01 \\
\SI{60}{\second} & 0.11 & 0.05 & 0.017 & 0.01 & 0.004 & 0.014 \\
\bottomrule
\end{tabular}
}
\end{table}

\subsubsection{Robotic Total Station Reference}
To improve accuracy and expand beyond open-sky outdoor settings, we follow previous work~\citep{vaidis2024rts} and establish ground-truth trajectories using a Leica MS60 total positioning station (TPS).
The Leica MS60 total station provides 3D position measurements at \SI{20}{\hertz} with a range accuracy of \SI{1.5}{\milli\meter}. Unlike prior work, our TPS measurements are adequately time synchronized to the sensor suite using a custom Leica Geosystems AP20 autopole~\citep{frey2025boxi}.

In contrast, most other works rely on stationary laser scanners \citep{zhao2024subt, tao2025oxfordspires} to generate highly accurate point maps; however, the collection is expensive, impractical, or even impossible in dynamic environments.
Moreover, the ground truth accuracy and its frequency are limited by the measurement characteristics of the onboard LiDAR used as a reference against the map, and are highly dependent on environment characteristics (e.g., LiDAR-degenerate environments).

While our TPS measurements are accurate, TPS requires a direct line-of-sight to the reflector mounted on the robot and, therefore, is not always available throughout a deployment. This directly motivates the fusion of inertial measurements, GNSS data, and TPS measurements, described next, to achieve highly accurate, high-frequency ground-truth pose estimates.

\begin{figure}[t]
    \centering
    \includegraphics[width=1\linewidth]{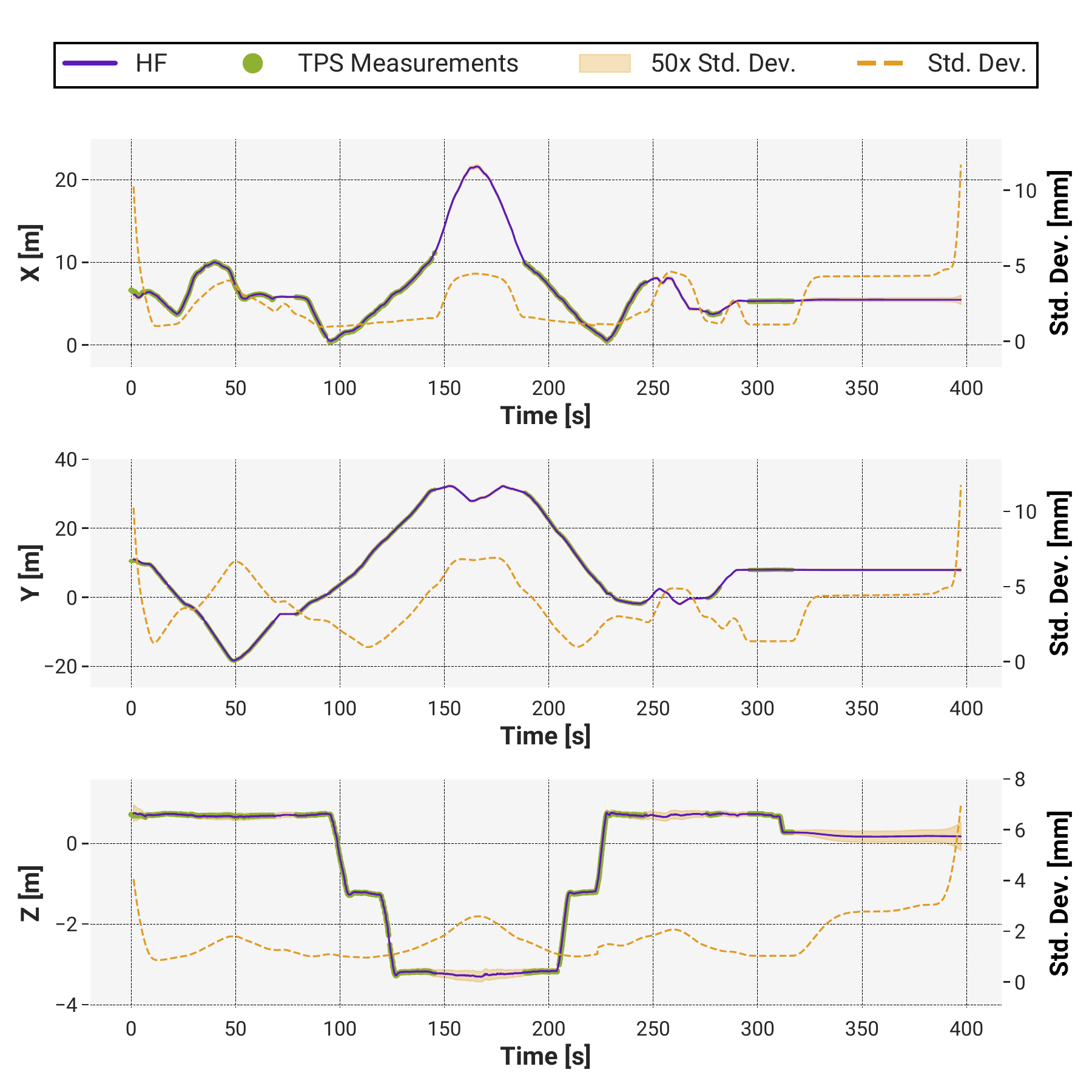}
    \caption{Holistic Fusion-based 6-DoF ground truth generation is shown for the SPX-2 sequence of GrandTour. 
    As expected, the estimate standard deviation increases when the MS60 line-of-sight is obstructed.}\label{fig:spx2_hf}
\end{figure}

\paragraph{Holistic Fusion Ground Truth}
Holistic Fusion~\citep{nubert2025holistic}, a flexible task- and setup-agnostic robot localization and state estimation library based on GTSAM~\citep{gtsam}, enables the fusion of TPS position measurements, Inertial Explorer post-processed poses, and \alias{HG4930} IMU measurements.
Holistic Fusion uses a factor graph formulation, where the state variables of the robot and the relative context variables, such as the alignment transforms, are estimated as a MAP estimate:
\begin{equation}
        \states^\star = \argmax_{\states} p(\states|\Measurements) \propto p(\Measurements|\states)p(\states).
\end{equation}
Here, $\states$ denotes the full set of robot states and dynamic state variables $\{ ^I \states_{N_I}, ^G \states_{N_G}, ^R \states_{N_R} \}$, with $^I \states_{N_I}$ denoting the robot state in the world frame, while $^G \states_{N_G}$ and $^R \states_{N_R}$ denote global (ECEF)- and reference (MS60) frame alignment-states.

The key element in this formulation is the alignment transform context variable, which actively aligns the two global fixed frames while ensuring the 6-DoF pose is retrieved by integrating IMU measurements, regardless of the availability of the Inertial Explorer post-processed pose. Despite this formulation, periods without both MS60 position measurements and Inertial Explorer post-processed unary poses require IMU-only dead-reckoning, leading to drift accumulation. Therefore, we restrict the ground-truth time span to segments with Inertial Explorer pose availability or MS60 measurements.

The combination of total station measurements, post-processed INS solution, and high-grade IMU delivers ground truth trajectories that outperform typical GNSS-only and LiDAR-only ground truth generation solutions.
This is illustrated in \cref{fig:spx2_hf}, where the Holistic Fusion output is shown to follow absolute position measurements when available and to complement them with Inertial Explorer estimates and the IMU when the line of sight is blocked.
When compared against the raw TPS position measurements, ATE for Holistic Fusion is a mean of \SI{0.0028}{\meter} with standard deviation ($\sigma$) of \SI{0.0020}{\meter} and RMSE \SI{0.0034}{\meter}, whereas the Inertial Explorer tightly coupled solution yields a mean ATE of \SI{0.132}{\meter} with $\sigma=$\SI{0.0721}{\meter} and RMSE \SI{0.1504}{\meter}. For RTE ($\Delta$ \SI{0.5}{\meter}), Holistic Fusion achieves a mean of \SI{0.0030}{\meter} with $\sigma=$\SI{0.0021}{\meter} and RMSE \SI{0.0037}{\meter}, compared to Inertial Explorer tightly coupled with mean \SI{0.0073}{\meter}, $\sigma=$\SI{0.0095}{\meter}, and RMSE \SI{0.012}{\meter}. Overall, Holistic Fusion can effectively fuse the 6-DoF Inertial Explorer estimation without violating the absolute TPS measurements when available.


\subsubsection{\revised{Ground Truth Utilization}}
\revised{For benchmark use, we recommend selecting the highest-fidelity reference trajectory available for the corresponding time span and mission segment.
The total-station reference position measurements represented in the "prism" frame should be the preferred choice if translation-only and sparse evaluation is acceptable.
The users can inspect the total-station coverage metric in \cref{tab:missions} to decide whether the coverage is sufficient for their evaluations.
For missions which have good GNSS coverage, the IE-TC post-optimized pose estimations provide dense and 6-DoF pose estimates.
These estimates are marginally less accurate than the total-station position measurements, but they provide realistic and complete ground truth, including per-pose covariance.
Advanced users may use the existing sensor-fusion solutions to get the best of both worlds by fusing the position measurements from the Total Station with the CPT7 IMU and the IE-TC pose estimations.
Users should check the released per-mission metadata for ground-truth availability before computing metrics and should crop estimator outputs to intervals where the selected reference is valid.
When developing methods that may consume auxiliary streams such as leg odometry, contact states, or derived maps, these streams should be treated as estimator outputs rather than independent ground-truth labels unless explicitly marked otherwise.}

\begin{figure*}[t]
    \centering
    \includegraphics[width=.95\linewidth]{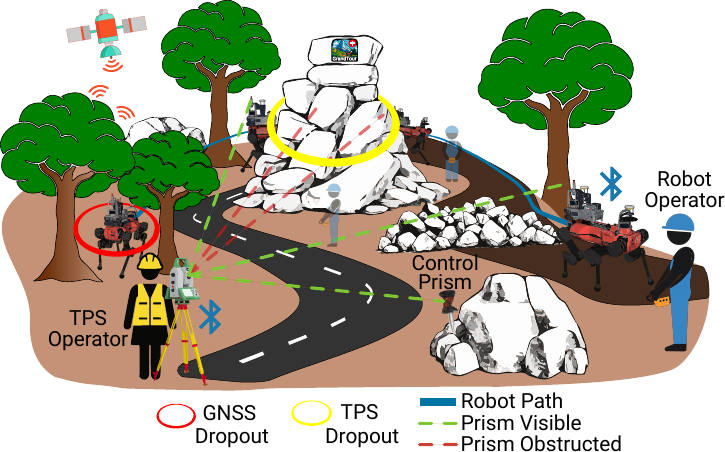}
    \caption{Data collection methodology for the GrandTour dataset.
    The ANYmal robot equipped with the \boxi sensor payload traverses environments with varying GNSS availability (red circle indicates denied regions) and total station visibility constraints (yellow circle).
    The deployment requires two operators: a Leica MS60 total positioning station (TPS) operator and a robot teleoperator. The TPS operator is responsible for maintaining uninterrupted tracking of the robot throughout the run, including re-acquisition after occlusions. The teleoperator controls the robot’s motion and is responsible for safe navigation through the environment.
    The control prism ensures the ground-truth measurement frame dictated by the TPS remains stable and static throughout the mission.}\label{fig:collection_figure}
\end{figure*}

\section{Data Collection}

In the following section, we explain the procedure for deploying ANYmal with the \boxi payload to collect a GrandTour mission.
\revised{All sequences of GrandTour were collected with the reinforcement-learning locomotion controller of \citet{miki2022learning}; commanded velocity tracking and resulting body-state characteristics should therefore be interpreted in the context of that controller when studying perception--action coupling or control-aware modeling.}
The sequence is further illustrated in \cref{fig:collection_figure}.\looseness-1

\begin{enumerate}[leftmargin=*, itemsep=0.2ex, topsep=0.6ex]
\item \textbf{Site Preparation:} Define a safe trajectory for robot teleoperation, specifically selecting diverse sites that present varying challenges.
Determine a good-coverage position for the Leica MS60 and the location of a reference prism (used to verify that the MS60 remains stationary throughout the mission).

\item \textbf{Start Recording:} Verify the validity of sensor data. Initiate recording and measure the reference prism location at the start of the sequence.
If the robot is within the line-of-sight of the total station, establish a lock on the robot.

\item \textbf{Operator Instruction:} The teleoperator follows behind the robot and aims to be not in the front-camera view.
The operator may physically intervene to ensure safety (e.g., stabilizing the robot when descending stairs or navigating near cliffs).
If the total station loses track of the robot, the operator navigates to the next location where a lock can be re-established.
In some cases, the robot must be stopped to allow for re-locking, resulting in intermittent pauses within the trajectory.

\item \textbf{Mission Completion:} Once the robot reaches the final destination, re-measure the reference prism position to validate station stability.
Stop the recording and power down the instruments.

\end{enumerate}

\begin{table*}[t]
  \footnotesize
  \centering
  \setlength{\tabcolsep}{3pt}

  \begin{threeparttable}
    \resizebox{1.0\textwidth}{!}{%
      \rowcolors{1}{white}{gray!12}

  \begin{tabular}{p{2.2cm}p{1.6cm}p{2.8cm}p{2.4cm}p{6.0cm}p{6.8cm}}
    \toprule
    \textbf{Sensor} & \textbf{Tag} & \textbf{HuggingFace} & \textbf{Rosbag} & \textbf{Topic} & \textbf{Sensor and Format} \\ \midrule

    \aliastable{CoreResearch} & <front\_center>, <left>, <right> &
    alphasense\_<tag> & alphasense &
    /boxi/alphasense/<tag>/camera\_info\newline
    /boxi/alphasense/<tag>/image\_raw/compressed &
    \SI{10}{\hertz} - 1440$\times$1080 - RGB\\

    \aliastable{CoreResearch} & <front\_left>, <front\_right> &
    alphasense\_<tag> & alphasense &
    /boxi/alphasense/<tag>/camera\_info\newline
    /boxi/alphasense/<tag>/image\_raw/compressed &
    \SI{10}{\hertz} - 1440$\times$1080 - MONO\\

    \aliastable{HDR} & <front>, <left>, <right> &
    hdr\_<tag> & hdr\_<tag> &
    /boxi/hdr/<tag>/camera\_info\newline
    /boxi/hdr/<tag>/image\_raw/compressed &
    \SI{10}{\hertz} - 1920$\times$1280 - RGB\\

    \aliastable{ZED2i} & <left>, <right> &
    zed2i\_<tag>\_image & zed2i\_images &
    /boxi/zed2i/<tag>/camera\_info\newline
    /boxi/zed2i/<tag>/image\_raw/compressed &
    \SI{15}{\hertz} - 1920$\times$1080 - RGB\\

    \aliastable{ZED2i} & -- &
    zed2i\_prop & zed2i\_prop &
    /boxi/zed2i/baro/pressure\newline
    /boxi/zed2i/magnetic\_field &
    Pressure + magnetic field\\

    \midrule

    \aliastable{Hesai} & -- &
    hesai\_packets & hesai\_packets &
    /boxi/hesai/packets &
    \SI{10}{\hertz} - Raw LiDAR packets\\
    
    \aliastable{Hesai} & -- &
    hesai & hesai &
    /boxi/hesai/points &
    \SI{10}{\hertz} - Hesai XT-32, $\sim$640{,}000 points [Single Return]\\

    \aliastable{Hesai} & -- & hesai & hesai &
    /boxi/hesai/intensity\_image\newline
    /boxi/hesai/range\_image &
    Projected range and intensity images\\

    \aliastable{Livox} & -- &
    livox\_points & livox &
    /boxi/livox/points &
    \SI{10}{\hertz} - Livox Mid360 non-repetitive LiDAR\\

    \aliastable{VLP16} & -- &
    velodyne & anymal\_velodyne &
    /anymal/velodyne/points &
    \SI{10}{\hertz} - Velodyne VLP-16 (Integrated in ANYmal)\\

    \aliastable{VLP16} & -- &
    anymal\_velodyne\_packets & anymal\_velodyne\_packets &
    /anymal/velodyne/packets &
    \SI{10}{\hertz} - Raw LiDAR packets\\

    \midrule
    \aliastable{ZED2i} & -- &
    zed2i\_depth\_image & zed2i\_depth &
    /boxi/zed2i/depth/camera\_info\newline
    /boxi/zed2i/depth/image\_raw/compressed &
    \SI{15}{\hertz} - 1920$\times$1080 - Stereolabs ZED2i Depth\\

    \aliastable{ZED2i} & -- &
    zed2i\_depth\_confidence\_image & zed2i\_depth &
    /boxi/zed2i/confidence/image\_raw/compressed &
    \SI{15}{\hertz} - 1920$\times$1080 - Stereolabs ZED2i Confidence\\

    \aliastable{ANYmal-Depth} & <front\_upper>, <front\_lower>, <rear\_upper>, <rear\_lower>, <left>, <right> &
    depth\_camera\_<tag> & anymal\_depth\_cameras &
    /anymal/depth\_camera/<tag>/depth/camera\_info\newline
    /anymal/depth\_camera/<tag>/depth/image\_rect\_raw &
    \SI{15}{\hertz} - \revised{848$\times$480} - RealSense D435i\\

    \midrule
    \aliastable{HG4930} & -- &
    cpt7\_imu & cpt7\_imu &
    /boxi/cpt7/imu &
    Honeywell HG4930, \SI{100}{\hertz}\\

    \aliastable{STIM320} & -- &
    stim320\_imu & stim320\_imu &
    /boxi/stim320/imu\newline
    /boxi/stim320/accelerometer\_temperature\newline
    /boxi/stim320/gyroscope\_temperature &
    Safran STIM320, \SI{500}{\hertz}\\

    \aliastable{ANYmal-IMU} & -- &
    anymal\_imu & anymal\_imu &
    /anymal/imu &
    \revised{ANYmal integrated IMU, \SI{400}{\hertz}}\\

    \aliastable{AP20-IMU} & -- &
    ap20\_imu & ap20\_imu &
    /boxi/ap20/imu &
    Leica AP20, \SI{200}{\hertz}\\

    \aliastable{ADIS} & -- &
    adis\_imu & adis &
    /boxi/adis/imu &
    Analog Devices ADIS16475-2, \SI{200}{\hertz}\\

    \aliastable{Bosch} & -- &
    alphasense\_imu & alphasense\_imu &
    /boxi/alphasense/imu &
    Bosch BMI085, \SI{200}{\hertz} (Integrated in CoreResearch)\\

    \aliastable{TDK} & -- &
    livox\_imu & livox\_imu &
    /boxi/livox/imu &
    TDK ICM40609, \SI{200}{\hertz} (integrated in Livox Mid360)\\

    \midrule
    \aliastable{Prism} & -- &
    prism\_position & ap20\_prism\_position &
    /boxi/ap20/prism\_position &
    Position relative to Leica MS60 Total Station\\

    \midrule
    \aliastable{ANYmal-JE} & -- &
    anymal\_state\_actuator & anymal\_state &
    /anymal/actuator\_readings &
    ANYmal joint readings, 12 actuators\\

    ANYmal & -- &
    anymal\_state\_battery & anymal\_state &
    /anymal/battery\_state &
    ANYmal battery state readings\\

    ANYmal & -- &
    anymal\_command\_twist & anymal\_command &
    /anymal/twist\_command &
    Operator-issued twist command\\

    \bottomrule
  \end{tabular}%
} 

  \end{threeparttable}
  \caption{Overview of all sensor streams recorded on a GrandTour sequence (excluding derived outputs), showing the sensor name, tag, corresponding HuggingFace dataset identifier, bag name, and ROS topics.}\label{tab:sensor_data}
\end{table*}

\begin{table*}[t]
  \footnotesize
  \centering
  \setlength{\tabcolsep}{3pt}

  \begin{threeparttable}
    \resizebox{1.0\textwidth}{!}{%
      \rowcolors{1}{white}{gray!12}

  \begin{tabular}{p{3.8cm}p{1.3cm}p{2.8cm}p{2.8cm}p{6.0cm}p{5.0cm}}
    \toprule
           \textbf{Sensor} & \textbf{Tag} & \textbf{HuggingFace} & \textbf{Rosbag} & \textbf{Topic} & \textbf{Sensor and Format} \\ \midrule

    \aliastable{Hesai} & -- & hesai\_undist & hesai\_undist &
    /boxi/hesai/points\_undistorted &
    Motion-compensated point cloud\\

    \aliastable{Livox} & -- & livox\_undist & livox\_undist &
    /boxi/livox/points\_undistorted &
    Motion-compensated point cloud\\

    \aliastable{VLP16} & -- & anymal\_velodyne\_undist & anymal\_velodyne\_undist &
    /anymal/velodyne/points\_undistorted &
    Motion-compensated point cloud\\

    \midrule
    \aliastable{CPT7} & <rt>, <tc> & cpt7\_ie\_<tag> & cpt7\_ie\_<tag> &
    /boxi/inertial\_explorer/<tag>/raw\newline
    /boxi/inertial\_explorer/<tag>/navsatfix\newline
    /boxi/inertial\_explorer/<tag>/odometry\newline
    /boxi/inertial\_explorer/<tag>/cpt7\_imu\_acc\_bias\newline
    /boxi/inertial\_explorer/<tag>/cpt7\_imu\_gyro\_drift\newline
    /tf &
    Inertial Explorer navigation solution + biases\\

    \midrule
    \aliastable{TF} & <minimal>, <model> & tf\_<tag> & tf\_<tag> &
    /tf\newline
    /tf\_static &
    Transform tree for sensor frames\\

    \midrule
    DLIO\:\citep{dlio} & -- & dlio & dlio &
    /boxi/dlio/hesai\_points\_undistorted\newline
    /boxi/dlio/lidar\_map\_odometry\newline
    /tf &
    Undistorted point cloud, \revised{LiDAR pose in map}\\

    \midrule
    \aliastable{ZED2i} & -- & zed2i\_vio & zed2i\_vio &
    /boxi/zed2i\_vio/map\newline
    /boxi/zed2i\_vio/odom\newline
    /tf &
    ZED2i VO outputs (map + odometry)\\

    \midrule
    Elevation Map \citep{miki2022elevation} & -- & anymal\_elevation & anymal\_elevation &
    /anymal/elevation\_mapping/elevation\_map\_raw &
    Elevation mapping output (raw map)\\

    \midrule
    TSIF & -- & anymal\_state\_state\_estimator & anymal\_state &
    /anymal/state\_estimator/anymal\_state\newline
    /anymal/state\_estimator/odometry\newline
    /anymal/state\_estimator/pose &
    ANYmal full state, estimator pose/odometry, contact estimation\\

    \bottomrule
  \end{tabular}%
} 

  \end{threeparttable}
  \caption{Overview of all derived outputs of the GrandTour dataset alongside a description of each output.}\label{tab:derived_data}
\end{table*}

\begin{figure}
    \centering
    \includegraphics[width=1\linewidth]{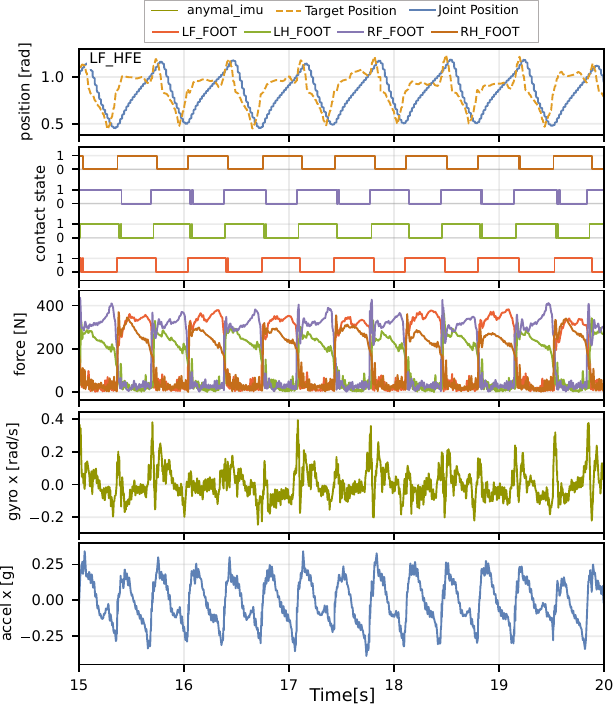}
    \caption{GrandTour dataset example from ANYmal showing synchronized proprioceptive signals over a \SI{5}{\second} window (\SIrange{15}{20}{\second} from bag start).
    From top to bottom: LF\_HFE joint position (state-estimator output, SEA-measured position, and commanded/target position), binary foot contact states \revised{from the onboard state estimator} (LF/RF/LH/RH), corresponding contact force magnitudes $\|\mathbf{f}\|$ (\si{\newton}), and IMU measurements ($\omega_x$ in \si{\radian\per\second}, $a_x$ in \textit{g} from \texttt{/anymal/imu}.}
    \label{fig:prioriData}
\end{figure}

\section{Collected Data and Derived Outputs}\label{sec:post_processed}
All available raw sensory data are detailed in \cref{tab:sensor_data}, and all derived outputs based on further processing during or after the mission are detailed in \cref{tab:derived_data}.
During the mission, we additionally record leg odometry, which is computed online using the \alias{ANYmal-IMU} data and \alias{ANYmal-JE}.
The state estimator responsible for this odometry also estimates feet contact states. 
\revised{The foot-contact states available through GrandTour, as shown in \cref{fig:prioriData}, are from the onboard state-estimator and not direct sensor measurements or ground-truth contact labels. 
Users developing contact-aware methods can derive their own contact estimates through, for example from joint-torque measurements or use these existing contact labels for ease of integration.}

In post-processing, leg odometry is used to generate motion-compensated LiDAR point clouds for each LiDAR.
Furthermore, we provide Inertial Navigation System (INS) trajectories derived from the CPT7.
These include both the real-time solution and a high-precision, tightly coupled solution generated via Inertial Explorer.
We also provide trajectory data based on Direct LiDAR-Inertial Odometry (DLIO)~\citep{dlio}.
In addition, data curation efforts ensure that the release is ready for downstream applications. 
These efforts include preserving driver-native message formats and filtering the robot body from the LiDAR point clouds.

\section{Dataset applications}\label{sec:applications}
In this section, we showcase how the GrandTour supports \textit{1)} State estimation and localization (\cref{sec:localization}), \textit{2)} Perception (\cref{sec:multi_modal}), and \textit{3)} Locomotion \& Navigation (\cref{sec:loc_nav}) applications in the robotics literature.

\subsection{State estimation and localization}\label{sec:localization}
Precise state estimation is essential for legged robots operating in real environments.
To expose failure modes across diverse motions (e.g., walking, trotting, slipping) and environments (indoor, urban, unstructured terrain), we benchmark a broad set of open-source solutions on six GrandTour sequences: SPX-2, SNOW-2, EIG-1, CON-4, ARC-2, and ARC-7 (mission metadata and preview videos are available in \cref{tab:missions}).

\begin{leftitemize}
    \item \textbf{SPX-2}: \revised{SPX-2 takes place at Jungfraujoch where the robot traverses an outdoor viewpoint deck with metal railings, grates, fences, stairs, nearby building facades, and pedestrians. 
    The robot performs a loop while repeatedly observing thin metallic structures and partially occluded walkways.
    In addition, there is only a very narrow static initialization window, challenging methods that expect a stationary initialization window.}
    
    \item \textbf{SNOW-2}: \revised{This mountain search-and-rescue mission crosses an open snow-covered area under direct sunlight, with rock faces, snow, and people in the scene. 
    Bright low-texture snow, glare, sparse vertical structure, and looped motion challenge visual feature extraction and geometric consistency.
    Snow-covered ground challenges contact estimation and feature-based LiDAR odometry.}
    
    \item \textbf{EIG-1}: \revised{The Alpine mission starts on rocky and gravel terrain and transitions through stairs, railings, pavement, and trackside station infrastructure. 
    The route combines non-planar natural structure with repetitive man-made geometry and strong sunlight, testing robustness across changing scene geometry and terrain.
    Importantly, the large elevation change challenges state-estimation methods across the board.}
    
    \item \textbf{CON-4}: \revised{CON-4 is the longest benchmark mission and traverses an outdoor construction site with mud, gravel, stairs, stacked materials, heavy machinery, cranes, workers, and surrounding buildings. 
    Its scale, cluttered foreground structure, moving people, and large loop make it particularly useful for exposing long-term drift and consistency errors.
    Changing terrain, dynamic foreground objects, and repeated occlusions further stress long-horizon consistency.}
    
    \item \textbf{ARC-2}: \revised{This search-and-rescue mission moves from an outdoor urban approach into tight concrete passages, stairwells, and bunker-like indoor--outdoor transitions. 
    Featureless concrete walls, narrow corridors, water-stained surfaces, and limited visibility challenge visual methods and can make single-LiDAR floor and ceiling observations unreliable.
    In addition, sharp turns that reduce visibility during wall-facing motion segments challenge vision-based methods.}
    
    \item \textbf{ARC-7}: \revised{ARC-7 is a night-time search-and-rescue mission where the robot traverses an industrial training environment with corrugated metal structures, stairs, smoke, and dark indoor--outdoor transitions. 
    The smoke, low-light passages, and limited re-observations significantly challenge visual methods and can lead to divergence or missing estimates.
    Most importantly, this mission challenges the dead-reckoning capabilities of vision-based methods while testing LiDAR-based methods through limited ground observability.}
\end{leftitemize}

\revised{In total, the benchmark contains 63 state-estimation methods, from different estimator families.} 
The benchmarked methods are summarized in \cref{tab:methods}.
\removed{In our evaluation, we excluded 16 additional works that we attempted to evaluate but were unable to run reliably. 
To limit speculation on performance, we do not disclose the names of these works.}
Additionally, we have omitted work that can be applied to multiple baseline solutions, such as information-aware sampling~\citep{rms, gelfand}, loop-closure~\citep{btc, kim2021scan}, offline bundle adjustment~\citep{hba}, or degeneracy-awareness~\citep{xicp, perfectlyconstrained, dcreg, kostasdegene} methods.
The evaluated methods span from (multi)-LiDAR odometry (LO), (multi)-LiDAR-inertial odometry and LiDAR-SLAM, (Stereo/Mono)-visual-inertial odometry (VIO), to multi-sensor variants (e.g., LIVO and RGB-D-inertial), \added{and legged-specific kinematic-inertial odometry methods}.

The authors spent a substantial amount of time adequately tuning and configuring each framework according to its published guidelines to conduct an objective evaluation of the framework through a third-party, i.e., the \textit{baseline} variant.
We have disabled loop-closure features for all methods, but have kept online bundle-adjustment and pose-graph-optimization features online.
While this is the case, the authors openly invite researchers to submit their \textit{author-tuned} variant to the benchmark that will be available on the dataset website.

We computed error metrics using the \texttt{evo} evaluation toolkit~\citep{grupp2017evo}.
Each estimated trajectory is transformed to be expressed in the GrandTour ground-truth reference frame (\texttt{prism} frame) using the provided frame conventions and extrinsic calibration.

For trajectory association, relying again on \texttt{evo}, this step accounts for differing output rates by matching each estimate timestamp to the closest ground-truth timestamp using threshold $t_{\max\_diff}$, set as $t_{\max\_diff}\approx 0.5\,\mathrm{median}(\Delta t_{\text{dense}})$, where $\Delta t_{\text{dense}}$ is the timestep of the higher-rate (more densely sampled) reference trajectory.
Concretely, for a \SI{20}{\hertz} reference trajectory and a \SI{10}{\hertz} estimator output, $\Delta t_{\text{dense}}=\SI{0.05}{\second}$ and thus $t_{\max\_diff}=\SI{0.025}{\second}$: if the reference timestamps are $\{0.00, 0.05, 0.10, 0.15, \dots\}$ and the estimator timestamps are $\{0.03, 0.13, 0.23, \dots\}$ (no jitter for brevity, just a fixed start offset), \texttt{evo} associates $0.03\!\rightarrow\!0.05$ and $0.13\!\rightarrow\!0.15$ with $|{\Delta}t|=\SI{0.02}{\second}\le \SI{0.025}{\second}$, so these pairs are accepted (whereas, e.g., $t_{\max\_diff}=\SI{0.01}{\second}$ would reject them).

For segment-based relative metrics, we use a path-length increment of $\Delta=\SI{0.5}{\meter}$, which is selected based on the motion characteristics of a quadruped robot.
Trajectory alignment to ground truth is performed via Umeyama least-squares~\citep{umeyama} over all correspondences.
It is important to note that the Umeyama alignment assumes that a single homogeneous 4$\times$4 transform is sufficient to align the two fixed frames (i.e., the trajectory represents the sensor's pose with respect to a single fixed frame).
In addition, we use Evo’s \texttt{point\_relation} functionality via the point-distance pose relation (\texttt{--pose\_relation point\_distance}), i.e., errors are computed as Euclidean distances between corresponding positions and are independent of orientation.

For each mission $s\in\mathcal{S}$ and each method $m\in\mathcal{M}$, we compute two ordinal ranks based on the reported mean errors: $r^{\mathrm{RTE}}_{m,s}$ by sorting $\mu^{\mathrm{RTE}}_{m,s}$ in ascending order, and $r^{\mathrm{ATE}}_{m,s}$ by sorting $\mu^{\mathrm{ATE}}_{m,s}$ in ascending order (lower is better). In case of equal means, ties are broken by the corresponding standard deviation $\sigma$ (ascending); diverged entries are assigned the worst rank $|\mathcal{M}|$ for the respective metric. The overall rank is then defined as the average of the two metric ranks,
\begin{equation}
r_{m,s} \triangleq \frac{r^{\mathrm{RTE}}_{m,s} + r^{\mathrm{ATE}}_{m,s}}{2}.
\end{equation}
Finally, the \emph{Summary} column reports the average rank across all missions,
\begin{equation}
\bar{r}_m \triangleq \frac{1}{|\mathcal{S}|}\sum_{s\in\mathcal{S}} r_{m,s},
\end{equation}
where smaller $\bar{r}_m$ indicates better overall performance.

Lastly, all methods were run on a platform with an Nvidia RTX 3090 GPU and an Intel i9-13900K CPU.\looseness-1

\begin{table*}[!ht]
\centering
\caption{Evaluated frameworks and their associated licenses.}
\label{tab:methods}

\begin{threeparttable}
  \resizebox{0.9\textwidth}{!}{%
  
    \begin{tabular}{p{6.1cm} p{2.4cm} p{5.5cm} p{1.7cm}}
    \hline
    \textbf{Method} & \textbf{Tested Category} & \textbf{Repository Link} & \textbf{License}\\
    \hline


    Adaptive-LIO~\citep{zhao2024adaptive} & LIO  &
    \href{https://github.com/chengwei0427/Adaptive-LIO}{chengwei0427/Adaptive-LIO} & BSD-3\tnote{a}\\

    AG-LOAM~\citep{agloam} & LO  &
    \href{https://github.com/UCR-Robotics/AG-LOAM}{\revised{UCR-Robotics/AG-LOAM}} & MIT\\

    
    BTSA~\citep{btsa} & LIO  &
    \href{https://github.com/thisparticle/btsa}{thisparticle/btsa} & GPL-2.0\\

    CLIC~\citep{clic} & LIVO  &
    \href{https://github.com/APRIL-ZJU/clic}{APRIL-ZJU/clic} & GPL-3.0\\


    Coco-LIC~\citep{cocolic} & LIVO  &
    \href{https://github.com/APRIL-ZJU/Coco-LIC}{APRIL-ZJU/Coco-LIC} & GPL-3.0\\

    CT-LO & LO  &
    \href{https://github.com/G3tupup/ctlo}{G3tupup/ctlo} & Unknown\\


    CTE-MLO~\citep{ctemlo} & (M)LO  &
    \href{https://github.com/shenhm516/CTE-MLO}{shenhm516/CTE-MLO} & GPL-2.0\\

    D-LI-Init~\citep{dlinit} & LIO  &
    \href{https://github.com/lian-yue0515/D-LI-Init}{lian-yue0515/D-LI-Init} & GPL-2.0\\

    DLIO~\citep{dlio} & LIO  &
    \href{https://github.com/vectr-ucla/direct_lidar_inertial_odometry}{vectr-ucla/direct\_lidar\_inertial\_odometry} & MIT\\

    DLO~\citep{dlo} & LO  &
    \href{https://github.com/vectr-ucla/direct_lidar_odometry}{vectr-ucla/direct\_lidar\_odometry} & MIT\\

    DMSA-SLAM~\citep{dmsa} & LiDAR-SLAM  &
    \href{https://github.com/davidskdds/DMSA_LiDAR_SLAM}{davidskdds/DMSA\_LiDAR\_SLAM} & MIT\\


    ENVIO~\citep{envio} & VIO  &
    \href{https://github.com/lastflowers/envio}{lastflowers/envio} & GPL-3.0\\

    FAST-LIMO & LIO  &
    \href{https://github.com/fetty31/fast_LIMO}{fetty31/fast\_LIMO} & GPL-3.0\\

    FAST-LIO2~\citep{fastlio2} & LIO  &
    \href{https://github.com/hku-mars/FAST_LIO}{hku-mars/FAST\_LIO} & GPL-2.0\\

    FAST-LIO-MULTI & (M)LIO &
    \href{https://github.com/engcang/FAST_LIO_MULTI}{engcang/FAST\_LIO\_MULTI} & GPL-2.0\\


    FAST-LIVO2~\citep{zheng2024fast} & LIVO  &
    \href{https://github.com/hku-mars/FAST-LIVO2}{hku-mars/FAST-LIVO2} & GPL-2.0\\

    Faster-LIO~\citep{fasterlio} & LIO  &
    \href{https://github.com/gaoxiang12/faster-lio}{gaoxiang12/faster-lio} & GPL-2.0\\

    GenZ-ICP~\citep{genzicp} & LO  &
    \href{https://github.com/cocel-postech/genz-icp}{cocel-postech/genz-icp} & MIT\\

    GLIM~\citep{glim} & LiDAR-SLAM &
    \href{https://github.com/koide3/glim}{koide3/glim} & MIT\\

    GLO~\citep{glo} & LO  &
    \href{https://github.com/robosu12/GLO}{robosu12/GLO} & GPL-3.0\\

    I$^{2}$EKF-LO~\citep{i2ekflo} & LO  &
    \href{https://github.com/YWL0720/I2EKF-LO}{YWL0720/I2EKF-LO} & GPL-2.0\\

    iG-LIO~\citep{iglio} & LIO  &
    \href{https://github.com/zijiechenrobotics/ig_lio}{zijiechenrobotics/ig\_lio} & GPL-2.0\\


    KISS-ICP~\citep{kissicp} & LO &
    \href{https://github.com/PRBonn/kiss-icp}{PRBonn/kiss-icp} & MIT\\

    Lightweight-VIO~\citep{lightweightvio} & VIO &
    \href{https://github.com/93won/lightweight_vio}{93won/lightweight\_vio} & MIT\\


    LIMOncello~\citep{limoncello} & LIO &
    \href{https://github.com/CPerezRuiz335/LIMOncello}{CPerezRuiz335/LIMOncello} & GPL-3.0\\

    LIO-EKF~\citep{lioekf} & LIO  &
    \href{https://github.com/YibinWu/LIO-EKF}{YibinWu/LIO-EKF} & MIT\\

    MAD-ICP~\citep{madicp} & LO  &
    \href{https://github.com/rvp-group/mad-icp}{rvp-group/mad-icp} & BSD-3\\

    MA-LIO~\citep{malio} & (M)LIO  &
    \href{https://github.com/minwoo0611/MA-LIO}{minwoo0611/MA-LIO} & GPL-2.0\\

    MINS~\citep{mins} & LIVO  &
    \href{https://github.com/rpng/MINS}{rpng/MINS} & GPL-3.0\\

    \added{ANYmal estimator~\citep{bloesch2018tsif}} & \added{KIO} &
    \added{--} & \added{--}\\

    \added{CAPO~\citep{sun2026capo}} & \added{KIO} &
    \added{\href{https://github.com/ShineMinxing/CAPO-LeggedRobotOdometry}{ShineMinxing/CAPO}} & \added{Apache-2.0}\\

    \added{Cerberus 2.0~\citep{yang2023cerberus}} & \added{KIO} &
    \added{\href{https://github.com/ShuoYangRobotics/Cerberus2.0}{ShuoYangRobotics/\allowbreak{}Cerberus2.0}} & \added{AGPL-3.0}\\

    \added{DRIFT~\citep{lin2023proprioceptive}} & \added{KIO} &
    \added{\href{https://github.com/UMich-CURLY/drift}{UMich-CURLY/drift}} & \added{BSD-3}\\


    \added{MUSE~\citep{nistico2025muse}} & \added{KIO} &
    \added{\href{https://github.com/iit-DLSLab/muse}{iit-DLSLab/muse}} & \added{BSD-3}\\

    NV-LIOM~\citep{nvliom} & LIO  &
    \href{https://github.com/dhchung/nv_liom}{dhchung/nv\_liom} & MIT\\

    OKVIS2~\citep{leutenegger2022okvis2} & VIO &
    \href{https://github.com/ethz-mrl/okvis2}{ethz-mrl/okvis2} & Custom\\


    Open-VINS~\citep{openvins} & VIO  &
    \href{https://github.com/rpng/open_vins}{rpng/open\_vins} & GPL-3.0\\


    PKO-LO~\citep{pkolo} & LO &
    \href{https://github.com/93won/lidar_odometry}{93won/lidar\_odometry} & MIT\\


    Point-LIO~\citep{pointlio} & LIO  &
    \href{https://github.com/hku-mars/Point-LIO}{hku-mars/Point-LIO} & Custom\\


    PV-LIO & LIO  &
    \href{https://github.com/HViktorTsoi/PV-LIO}{HViktorTsoi/PV-LIO} & Apache-2.0\\

    R$^2$DIO~\citep{r2dio} & \revised{RGB-D} &
    \href{https://github.com/jiejie567/R2DIO}{jiejie567/R2DIO} & GPL-3.0\\

    R-VIO2~\citep{rvio2} & Mono-VIO &
    \href{https://github.com/rpng/R-VIO2}{rpng/R-VIO2} & GPL-3.0\\

    RESPLE~\citep{resple} & (M)LO\;\&\;(M)LIO& \href{https://github.com/ASIG-X/RESPLE}{ASIG-X/RESPLE} & GPL-3.0\\

    RKO-LIO~\citep{rkolio} & LIO & \href{https://github.com/PRBonn/rko_lio}{PRBonn/rko\_lio}& MIT\\

    ROVIO~\citep{rovio} & Mono-VIO  &
    \href{https://github.com/ethz-asl/rovio}{ethz-asl/rovio} & BSD-3\\

    SchurVINS~\citep{schurvins} & VIO &
    \href{https://github.com/bytedance/SchurVINS}{bytedance/SchurVINS} & GPL-3.0\\

    Section-LIO~\citep{sectionlio} & LIO  &
    \href{https://github.com/mengkai98/Section-LIO}{mengkai98/Section-LIO} & Unknown\\


    SLICT2~\citep{slict2} & LIO & \href{https://github.com/brytsknguyen/slict}{brytsknguyen/slict} & GPL-2.0\\

    SR-LIO~\citep{srlio} & LIO  &
    \href{https://github.com/ZikangYuan/sr_lio}{ZikangYuan/sr\_lio} & GPL-2.0\\

    SuperOdom~\citep{superodom} & LIO &
    \href{https://github.com/superxslam/SuperOdom}{superxslam/SuperOdom} & GPL-3.0\\

    Traj-LO~\citep{trajlo} & LO &
    \href{https://github.com/kevin2431/Traj-LO}{kevin2431/Traj-LO} & MIT\\

    $\sqrt{\mathrm{VINS}}$~\citep{sqrtVINS} & VIO  &
    \href{https://github.com/rpng/sqrtVINS}{rpng/sqrtVINS} & LGPL-3.0\\

    VINS-Fusion~\citep{vinsfusion} & VIO &
    \href{https://github.com/HKUST-Aerial-Robotics/VINS-Fusion}{HKUST-Aerial-Robotics/VINS-Fusion} & GPL-3.0\\


    VINS-RGBD~\citep{vinsrgbd} & \revised{RGB-D} &
    \href{https://github.com/Lab-of-AI-and-Robotics/VINS-RGBD}{Lab-of-AI-and-Robotics/VINS-RGBD} & GPL-3.0\\

    Voxel-SLAM~\citep{voxelslam} & Visual-SLAM  &
    \href{https://github.com/hku-mars/Voxel-SLAM}{hku-mars/Voxel-SLAM} & GPL-2.0\\

    Voxel-SVIO~\citep{voxelsvio} & VIO  &
    \href{https://github.com/ZikangYuan/voxel_svio}{ZikangYuan/voxel\_svio} & GPL-3.0\\



    \hline
    \end{tabular}%
  } 

  \begin{tablenotes}\footnotesize
    \item[a] License not explicitly tagged by GitHub; BSD-3 inferred from the repository's ROS \texttt{package.xml}.
  \end{tablenotes}
\end{threeparttable}

\end{table*}

\subsubsection{LiDAR Odometry}\label{sec:localization_lo}
LiDAR odometry can be considered one of the simplest methods for retrieving ego-motion.
Often, the methods differ in how the trajectory is represented and how motion compensation is applied to the point cloud.
The error metrics of the evaluated open-source solutions are provided in Table~\ref{table:RTE_ATE_results_LO}.
These methods have been employed using \alias{Hesai} LiDAR.

As seen in Table~\ref{table:RTE_ATE_results_LO}, {Traj-LO} achieves the best overall average rank and is the top performer on SPX-2 and EIG-1 (best RTE/ATE), while also leading on SNOW-2 ATE and CON-4 RTE.
However, its performance drops substantially on ARC-2, where several methods outperform it. The general performance degradation on the ARC-2 sequence might be explained by the limited visibility in the tight indoor settings. 
This notion is later strengthened with the utilization of multiple LiDARs in \Cref{sec:localization_multi_lo}.

The next best overall (by average rank) are {DLO} and {I$^{2}$EKF-LO}.
In particular, {I$^{2}$EKF-LO} is best on ARC-7 (both RTE and ATE), while {DLO} is best on ARC-2 RTE (and second-best ARC-2 ATE).
Several methods exhibit mission-dependent weaknesses or missing results (e.g., failures on ARC-2 and/or CON-4), as reflected in their lower summary ranks.

Interestingly, compared to the methods discussed in \Cref{sec:localization_lio}, LiDAR-only methods are competitive. 
However, they are more prone to degradation as there is no redundancy in the system.

\begin{table*}[t]\centering
    \caption{Error metrics for the LO methods for the test sequences, per mission (best in \textbf{bold}, second best \underline{underlined}). \emph{Summary} column reports the average rank across the six missions, where for each mission the method rank is the mean of its RTE-rank and ATE-rank (lower is better); missing results are assigned the worst rank.}
    \resizebox{2\columnwidth}{!}{%
    \renewcommand{\arraystretch}{1.18}%
    \begin{tabular}{l c @{\hspace{1.2em}} cc cc cc cc cc cc}
    \toprule[1pt]
    & \multicolumn{1}{c}{Summary} & \multicolumn{2}{c}{SPX-2} & \multicolumn{2}{c}{SNOW-2} & \multicolumn{2}{c}{EIG-1} & \multicolumn{2}{c}{CON-4} & \multicolumn{2}{c}{ARC-2} & \multicolumn{2}{c}{ARC-7} \\
    \cmidrule(lr){2-2} \cmidrule(lr){3-4} \cmidrule(lr){5-6} \cmidrule(lr){7-8} \cmidrule(lr){9-10} \cmidrule(lr){11-12} \cmidrule(lr){13-14}
    & \makecell{Avg.\\rank}
    & \makecell{RTE \\ $\mu(\sigma)[\si{\centi\meter}]$ } & \makecell{ATE \\ $\mu(\sigma)[\si{\centi\meter}]$ }
    & \makecell{RTE \\ $\mu(\sigma)[\si{\centi\meter}]$ } & \makecell{ATE \\ $\mu(\sigma)[\si{\centi\meter}]$ }
    & \makecell{RTE \\ $\mu(\sigma)[\si{\centi\meter}]$ } & \makecell{ATE \\ $\mu(\sigma)[\si{\centi\meter}]$ }
    & \makecell{RTE \\ $\mu(\sigma)[\si{\centi\meter}]$ } & \makecell{ATE \\ $\mu(\sigma)[\si{\centi\meter}]$ }
    & \makecell{RTE \\ $\mu(\sigma)[\si{\centi\meter}]$ } & \makecell{ATE \\ $\mu(\sigma)[\si{\centi\meter}]$ }
    & \makecell{RTE \\ $\mu(\sigma)[\si{\centi\meter}]$ } & \makecell{ATE \\ $\mu(\sigma)[\si{\centi\meter}]$ } \\
    \midrule[1pt]

    Traj-LO         & \textbf{2.58} & \textbf{1.11 (0.77)} & \textbf{2.88 (1.26)} & \underline{1.28 (1.05)} & \textbf{1.86 (1.03)} & \textbf{1.12 (0.89)} & \textbf{3.09 (1.59)} & \textbf{1.07 (0.9)} & \underline{2.29 (1.07)} & 17.36 (30.26) & 258.43 (105.16) & \underline{1.12 (0.82)} & \underline{2.93 (2.16)} \\

    \rowcolor{gray!10} DLO             & \underline{5.08} & 1.77 (1.65) & 5.74 (2.87) & 3.07 (3.05) & 6.83 (4.65) & 2.49 (2.62) & 8 (4.63) & 4.11 (4.73) & 9.43 (6.01) & \textbf{2.91 (4.83)} & \underline{13.71 (7.09)} & 2.01 (5.26) & 5.37 (9.58) \\

    I$^{2}$EKF-LO   & \underline{5.08} & \underline{1.22 (1.16)} & \underline{3.87 (1.96)} & 4.04 (4) & 8.03 (6.75) & 4.29 (5.94) & 18.6 (13.78) & 1.32 (1.04) & 3.51 (1.67) & 3.69 (7.45) & 43.27 (16.48) & \textbf{0.85 (0.64)} & \textbf{1.64 (1.09)} \\

    \rowcolor{gray!10} CT-LO           & 5.17 & 1.55 (1.38) & 9.86 (5.33) & 1.45 (1.12) & 7.89 (2.75) & \underline{1.53 (1.33)} & 10.35 (4.38) & 1.39 (1.16) & 11.56 (3.78) & 4.23 (13.18) & 39.87 (17.31) & 1.2 (1.19) & 4.38 (1.77) \\

    CTE-MLO         & 5.17 & 2.12 (2.12) & 9 (5.21) & \textbf{1.21 (1.04)} & \underline{1.99 (1.12)} & 1.65 (1.6) & \underline{3.79 (1.61)} & \underline{1.1 (1.03)} & \textbf{2.16 (1)} & 11.13 (34.2) & 286.67 (115.25) & 2.34 (5.65) & 7.53 (16.54) \\

    \rowcolor{gray!10} AG-LOAM         & 6.00 & 1.59 (1.42) & 8.48 (4.07) & 3.92 (3.55) & 7.69 (5.35) & 3.11 (3.37) & 7.69 (5.89) & 7.47 (9.73) & 12.27 (9.33) & 3.52 (5.94) & \textbf{9.51 (11.39)} & 2.07 (4.55) & 5.3 (8.55) \\

    GenZ-ICP        & 6.50 & 1.51 (1.46) & 4.57 (1.98) & 26.4 (20.64) & 50.39 (40.15) & 3.75 (5.95) & 6.31 (4.39) & 5.57 (10.03) & 12.8 (17.82) & \underline{3.09 (3.8)} & 20.65 (8.12) & 2.55 (6.73) & 3.69 (4.32) \\

    \rowcolor{gray!10} RESPLE-LO       & 7.00 & 1.4 (1.3) & 5.11 (2.75) & 2.01 (3.06) & 2.7 (2.5) & 2.02 (2.15) & 13.27 (3.99) & 2.77 (6.34) & 4.7 (4.6) & - & - & 12.11 (17.52) & 12.87 (8.67) \\

    PKO-LO          & 7.33 & 2.64 (3.43) & 46.23 (22.59) & 2.33 (1.86) & 10.6 (3.62) & 2.42 (2.39) & 36.6 (17.62) & 2.56 (2.36) & 10.43 (5.81) & 3.71 (5.54) & 20.52 (10.64) & 1.86 (2.05) & 8.54 (3.5) \\

    \rowcolor{gray!10} GLO             & 8.67 & 1.87 (1.75) & 6.68 (2.81) & 3.84 (3.74) & 7.43 (4.06) & 3.37 (2.96) & 11.64 (5.68) & 3.16 (3.07) & 24.44 (14.42) & - & - & 6.06 (18.9) & 9.39 (18.16) \\

    MAD-ICP         & 9.00 & 2.21 (2.35) & 12.15 (4.42) & 18.17 (20.91) & 16.64 (13.25) & 6.64 (8.69) & 13.99 (6.09) & 10.43 (14.92) & 137.28 (91.17) & 6.1 (10) & 41.68 (14.32) & 1.81 (1.71) & 7.05 (2.57) \\

    \rowcolor{gray!10} KISS-ICP        & 10.58 & 3.28 (5.6) & 9.02 (4.25) & 35.78 (21.97) & 47.8 (25.96) & 7.76 (9.61) & 13.28 (6.84) & - & - & 21.66 (32.01) & 302.37 (135.03) & 3.85 (5.66) & 7.21 (4.45) \\

    \bottomrule[1pt]
    \end{tabular}%
    }
    \label{table:RTE_ATE_results_LO}
\end{table*}

\subsubsection{LiDAR-Inertial-(Visual) Odometry}\label{sec:localization_lio}
LiDAR-inertial odometry is currently one of the most widely used methods for robot ego-motion estimation and SLAM.
Compared to LiDAR odometry, the addition of high-rate inertial measurements has shown a clear advantage during highly dynamic motions.
Furthermore, the availability of observations of gravity direction enables stable state estimation.
While the advantages are vast, the addition of another modality brings challenges such as time synchronization and intrinsic-extrinsic calibration.
The development and improvement of algorithms are still of interest to the robotics community.
Under this premise, we have evaluated numerous LIO methods, including LIVO variants.
The error metrics of the evaluated open-source LIO solutions are provided in Table~\ref{table:RTE_ATE_results_LIO_LiDARSLAM}.
These methods have been employed using the \alias{Hesai} LiDAR and the \alias{HG4930} IMU.

Overall, Coco-LIC and FAST-LIVO2 obtain the best average ranks.
However, even the top-performing method achieves an average rank of around 4, indicating that no single approach dominates across all missions.
Instead, performance is clearly mission-dependent: for example, Voxel-SLAM attains the best ATE on SPX-2, DLIO attains the best ATE on EIG-1, SR-LIO attains the best ATE on CON-4, iG-LIO attains the best ATE on ARC-7, while FAST-LIVO2 attains the best RTE/ATE on ARC-2.

The confined ARC-2 mission is particularly challenging and exposes robustness differences.
Several methods exhibit drastic drift, and some methods diverge entirely.
This behavior is consistent with the reduced geometric constraint in tight environments where stable registration features are intermittently absent.

Across the other missions, many methods are tightly clustered: for several sequences, differences among the better-performing approaches are on the order of a few millimeters to about a centimeter in RTE and a few centimeters in ATE, often comparable to the reported standard deviations.
Consequently, incremental improvements across methods are difficult to claim unambiguously from aggregate ranks alone, and small rank changes can be driven by a single mission.
For this reason, in addition to summary statistics, per-mission results (and failure cases) should be emphasized when comparing LIO/LIVO systems.

Many methods claim distinct advantages, but these advantages are difficult to isolate on non-tailored real-world test cases in GrandTour. For example, Point-LIO and RESPLE can operate on a per-point basis rather than on full point-cloud scans, while D-LI Init supports initialization during motion. Moreover, solutions such as GLIM are feature-rich and under active development, but their many tuning parameters make it hard to find a configuration that performs well across missions.

It is important to note that, although two of the top three methods are LIVO methods, the advantage of vision is ambiguous. Upon closer inspection, the strongest LIVO methods weight visual data less than LiDAR data, whereas CLIC, another LIVO method, performs worse while weighting visual observations more heavily than LiDAR observations.





\begin{table*}[t]\centering
    \caption{Error metrics for LIO and LiDAR-SLAM methods (including LIVO) on test sequences, reported in \si{\centi\meter} (best in bold, second-best \underline{underlined}). \emph{Summary} column reports the average rank across the six missions, where for each mission the method rank is the mean of its RTE-rank and ATE-rank (lower is better); missing results are assigned the worst rank.
    }
    \resizebox{2\columnwidth}{!}{%
    \renewcommand{\arraystretch}{1.18}%
    \begin{tabular}{l c @{\hspace{1.2em}} cc cc cc cc cc cc}
    \toprule[1pt]
    & \multicolumn{1}{c}{Summary} & \multicolumn{2}{c}{SPX-2} & \multicolumn{2}{c}{SNOW-2} & \multicolumn{2}{c}{EIG-1} & \multicolumn{2}{c}{CON-4} & \multicolumn{2}{c}{ARC-2} & \multicolumn{2}{c}{ARC-7} \\
    \cmidrule(lr){2-2} \cmidrule(lr){3-4} \cmidrule(lr){5-6} \cmidrule(lr){7-8} \cmidrule(lr){9-10} \cmidrule(lr){11-12} \cmidrule(lr){13-14}
    & \makecell{Avg.\\rank}
    & \makecell{RTE \\ $\mu(\sigma)[\si{\centi\meter}]$ } & \makecell{ATE \\ $\mu(\sigma)[\si{\centi\meter}]$ }
    & \makecell{RTE \\ $\mu(\sigma)[\si{\centi\meter}]$ } & \makecell{ATE \\ $\mu(\sigma)[\si{\centi\meter}]$ }
    & \makecell{RTE \\ $\mu(\sigma)[\si{\centi\meter}]$ } & \makecell{ATE \\ $\mu(\sigma)[\si{\centi\meter}]$ }
    & \makecell{RTE \\ $\mu(\sigma)[\si{\centi\meter}]$ } & \makecell{ATE \\ $\mu(\sigma)[\si{\centi\meter}]$ }
    & \makecell{RTE \\ $\mu(\sigma)[\si{\centi\meter}]$ } & \makecell{ATE \\ $\mu(\sigma)[\si{\centi\meter}]$ }
    & \makecell{RTE \\ $\mu(\sigma)[\si{\centi\meter}]$ } & \makecell{ATE \\ $\mu(\sigma)[\si{\centi\meter}]$ } \\
    \midrule[1pt]

    {Coco-LIC}            & \textbf{4.17} & \textbf{0.44 (0.39)} & 3.7 (1.53) & \textbf{0.41 (0.3)} & \textbf{1.06 (0.55)} & \textbf{0.4 (0.39)} & 4.46 (1.76) & \textbf{0.43 (0.33)} & 2.4 (1.23) & \underline{1.01 (2.59)} & 8.07 (5.05) & \textbf{0.37 (0.3)} & 3.81 (3.56) \\

    \rowcolor{gray!10} {FAST-LIVO2}          & \underline{4.42} & 1.05 (0.82) & \underline{2.04 (0.97)} & 1.25 (1.01) & 1.43 (0.74) & 1.11 (0.88) & \underline{3.02 (1.38)} & 1.18 (0.94) & 2.73 (1.35) & \textbf{0.7 (0.61)} & \textbf{1.1 (0.51)} & 1.13 (0.81) & \underline{1.22 (0.56)} \\

    {PV-LIO}              & 7.50 & 1.14 (0.91) & 3.74 (1.64) & 1.32 (1) & 3.18 (1.23) & 1.24 (0.96) & 3.15 (1.16) & 1.22 (0.96) & 4.57 (1.58) & 1.05 (0.86) & 2.28 (0.93) & 0.99 (0.74) & 2.02 (0.86) \\

    \rowcolor{gray!10} {DLIO}                & 7.67 & 1.22 (1.04) & 2.74 (1.22) & 1.51 (1.19) & 1.61 (0.85) & 1.26 (0.97) & \textbf{2.65 (1.39)} & 1.43 (1.16) & 2.28 (1.06) & 1.19 (1.39) & 4.26 (1.73) & 1.15 (0.9) & 3.78 (4.6) \\

    {Fast-LIMO}           & 8.17 & 1.15 (0.99) & 5.76 (2.58) & 1.16 (0.9) & 2.41 (1.31) & \underline{1.1 (1.6)} & 8.75 (3.3) & 0.89 (0.74) & 2.65 (1.5) & 2.49 (3.35) & 10.2 (6.94) & \underline{0.61 (0.5)} & 2.31 (1.57) \\

    \rowcolor{gray!10} {Voxel-SLAM}          & 8.92 & 1.03 (0.83) & \textbf{2.03 (1.02)} & 1.37 (1.13) & 1.57 (0.81) & 1.34 (1.03) & 3.33 (1.4) & 3.24 (9.62) & 50.87 (72.99) & 1.24 (1.43) & 8.66 (5.57) & 0.91 (0.73) & 1.32 (0.65) \\

    {Faster-LIO}          & 9.08 &
    1.02 (0.89) & 2.68 (1.26) &
    1.46 (1.35) & 7.15 (2.55) &
    1.3 (1.13) & 5.38 (2.73) &
    1.51 (1.86) & 9.04 (3.93) &
    2.01 (3.49) & 5.88 (5.26) & 
    0.94 (0.73) & 1.49 (0.65) \\

    \rowcolor{gray!10} {SLICT2}              & 9.17 & 1.11 (1.64) & 3.69 (2.3) & 1.27 (1.05) & \underline{1.37 (0.65)} & 1.18 (0.88) & 4.85 (1.97) & 1.19 (0.94) & 1.97 (0.89) & - & - & 0.9 (0.68) & 1.59 (0.61) \\

    {iG-LIO}              & 9.25 & 1.47 (1.24) & 4.94 (2.28) & 1.16 (1.1) & 1.6 (1.5) & 8.09 (13.47) & 6.3 (11.19) & 1.19 (0.92) & 2.53 (1.23) & 2.66 (8.33) & 7.53 (14.53) & 0.9 (0.68) & \textbf{1.15 (0.61)} \\

    \rowcolor{gray!10} {LIMOncello}          & 10.58 & 1.92 (2.93) & 3.23 (1.67) & 3.01 (4.39) & 1.77 (2.2) & 2.39 (3.5) & 3.3 (1.95) & 2.73 (4.22) & \underline{1.88 (1.98)} & 2.1 (3.34) & \underline{1.87 (1.93)} & 1.86 (3.4) & 2.68 (5.78) \\

    {FAST-LIO2}           & 11.58 & 1.07 (0.9) & 4.15 (2.26) & 1.41 (1.13) & 6.47 (2.67) & 1.28 (0.95) & 4.97 (2.65) & 1.54 (1.86) & 8.79 (3.83) & 3.62 (10.63) & 47.63 (16.36) & 1.04 (0.72) & 2.74 (1.45) \\

    \rowcolor{gray!10} {RKO-LIO}             & 11.75 & 1.22 (1.03) & 3.45 (1.7) & 2.4 (2.41) & 3.4 (3.92) & 1.41 (0.99) & 4.31 (1.83) & 1.39 (1.12) & 2.61 (1.45) & 3.87 (4.88) & 21.54 (15.56) & 1.58 (1.14) & 6.77 (3.64) \\

    {SR-LIO}              & 12.25 & 1.78 (1.42) & 8.35 (5.46) & 1.35 (1.05) & 1.74 (0.87) & 1.14 (0.92) & 3.29 (1.52) & 1.44 (1.05) & \textbf{1.84 (0.87)} & - & - & 1.39 (1.06) & 3.05 (1.6) \\

    \rowcolor{gray!10} {RESPLE}              & 12.42 & 0.93 (0.83) & 3.83 (1.79) & 1.28 (1.63) & 2.08 (1.63) & 1.41 (1.64) & 9.7 (4.52) & 1.41 (1.99) & 4.41 (3.09) & 3.61 (4.19) & 9.71 (5.01) & 6.16 (12.8) & 33.2 (10.17) \\

    {SuperOdom}           & 14.67 & 3.9 (12.62) & 63.5 (42.63) & 1.53 (1.27) & 6.64 (6.58) & 1.47 (1.25) & 6.01 (3.03) & 1.64 (2.29) & 6.72 (4.17) & 2.82 (4.31) & 14.61 (9.93) & 1.33 (0.98) & 2.82 (1.19) \\

    \rowcolor{gray!10} {MINS}            & 14.75 & \underline{0.73 (0.61)} & 13.7 (4.81) & 10.02 (4.13) & 213.09 (130.02) & 1.1 (1.68) & 50.41 (20.14) & 0.77 (1.87) & 46.01 (20.38) & 5.69 (22.9) & 68.15 (53.05) & 0.72 (0.83) & 11.9 (3.55) \\

    {D-LI-Init}           & 16.00 & 1.26 (0.98) & 3.92 (1.94) & 1.96 (3.53) & 2.41 (3.27) & 2.46 (2.81) & 5.62 (5.04) & 1.61 (3.15) & 2.35 (2.52) & 5.06 (23.39) & 121.39 (49.16) & - & - \\

    \rowcolor{gray!10} {GLIM}                & 17.92 & 0.96 (0.79) & 4.19 (2.29) & 49.65 (28.59) & 24.73 (15.16) & 1.39 (1.07) & 8.87 (4.4) & 20.46 (24.31) & 18.8 (14.16) & 43.99 (37.63) & 103.1 (51.41) & 4.39 (5.44) & 12.21 (5.69) \\

    {BTSA}                & 18.00 & 2.41 (5.63) & 30.03 (5.99) & 1.66 (3.49) & 27.61 (5.49) & 1.67 (1.58) & 29.86 (4.85) & 1.65 (3.34) & 31.1 (1.45) & 4.21 (9.36) & 30.68 (2.12) & 2.02 (5.35) & 17.29 (8.52) \\

    \rowcolor{gray!10} {Section-LIO}         & 18.42 & 2.27 (2.29) & 10.16 (3.85) & 1.64 (1.2) & 7.27 (2.62) & 1.67 (1.2) & 7.27 (2.63) & 1.83 (1.63) & 9.4 (3.19) & - & - & 2.3 (1.8) & 11.79 (8.64) \\

    {CLIC}                & 18.67 & 4.05 (14.67) & 72.73 (69.22) & \underline{0.54 (0.42)} & 1.4 (0.66) & - & - & \underline{0.62 (0.59)} & 6.25 (4.03) & - & - & - & - \\

    \rowcolor{gray!10} {Point-LIO}           & 20.67 & 2.2 (2.74) & 20.43 (8.01) & 8.47 (11.22) & 85.59 (30.25) & 6.01 (9.65) & 60.07 (25.26) & 1.94 (3.81) & 25.83 (6) & 10.94 (21.75) & 49.84 (28.28) & 4.7 (11.14) & 26.4 (10.8) \\

    {DMSA-SLAM}           & 20.75 & 4.05 (8.01) & 48.98 (5.15) & 3.82 (4.66) & 48.01 (6.41) & 2.99 (4.19) & 47.64 (6.83) & 3.55 (6.96) & 44.93 (11.22) & 9.1 (18.22) & 49.19 (10.06) & 3.42 (2.81) & 27.98 (13.58) \\

    \rowcolor{gray!10} {NV-LIOM}             & 21.08 & 6.35 (15.86) & 33.59 (7.39) & 3.84 (5.47) & 31.59 (3.57) & 3.55 (4.94) & 31.15 (6.31) & 3.85 (9.32) & 33.26 (10.47) & 13.4 (24.95) & 29.13 (10.74) & 4.72 (6.39) & 21.21 (9.63) \\

    {Adaptive-LIO}        & 22.33 & 4.5 (11.06) & 79.01 (63.65) & 18.18 (17.36) & 27.83 (11.37) & 11.93 (17.78) & 437.33 (245.78) & 6.93 (13.45) & 323.49 (187.48) & 11.31 (24.53) & 217.2 (90.87) & 1.91 (2.49) & 4.28 (3.16) \\

    \rowcolor{gray!10} {LIO-EKF}             & 22.67 & 4.87 (4.15) & 10.84 (5.63) & 7.58 (6.78) & 11.86 (7.25) & 5.32 (6.18) & 65.67 (33.01) & 5.05 (5.57) & 9.88 (6.58) & - & - & 13.12 (10.35) & 37.79 (28.82) \\

    \bottomrule[1pt]
    \end{tabular}%
    }
    \label{table:RTE_ATE_results_LIO_LiDARSLAM}
\end{table*}

\begin{table*}[t]\centering
    \caption{Error metrics for Multi-LiDAR methods for the test sequences (best in \textbf{bold}, second best \underline{underlined}). \emph{Summary} column reports the average rank across the \revised{six} missions, where for each mission the method rank is the mean of its RTE-rank and ATE-rank (lower is better); missing results are assigned the worst rank.}
    \resizebox{2\columnwidth}{!}{%
    \renewcommand{\arraystretch}{1.18}%
    \begin{tabular}{l c @{\hspace{1.2em}} cc cc cc cc cc cc}
    \toprule[1pt]
    & \multicolumn{1}{c}{Summary} & \multicolumn{2}{c}{SPX-2} & \multicolumn{2}{c}{SNOW-2} & \multicolumn{2}{c}{EIG-1} & \multicolumn{2}{c}{CON-4} & \multicolumn{2}{c}{ARC-2} & \multicolumn{2}{c}{ARC-7} \\
    \cmidrule(lr){2-2} \cmidrule(lr){3-4} \cmidrule(lr){5-6} \cmidrule(lr){7-8} \cmidrule(lr){9-10} \cmidrule(lr){11-12} \cmidrule(lr){13-14}
    & \makecell{Avg.\\rank}
    & \makecell{RTE \\ $\mu(\sigma)[\si{\centi\meter}]$ } & \makecell{ATE \\ $\mu(\sigma)[\si{\centi\meter}]$ }
    & \makecell{RTE \\ $\mu(\sigma)[\si{\centi\meter}]$ } & \makecell{ATE \\ $\mu(\sigma)[\si{\centi\meter}]$ }
    & \makecell{RTE \\ $\mu(\sigma)[\si{\centi\meter}]$ } & \makecell{ATE \\ $\mu(\sigma)[\si{\centi\meter}]$ }
    & \makecell{RTE \\ $\mu(\sigma)[\si{\centi\meter}]$ } & \makecell{ATE \\ $\mu(\sigma)[\si{\centi\meter}]$ }
    & \makecell{RTE \\ $\mu(\sigma)[\si{\centi\meter}]$ } & \makecell{ATE \\ $\mu(\sigma)[\si{\centi\meter}]$ }
    & \makecell{RTE \\ $\mu(\sigma)[\si{\centi\meter}]$ } & \makecell{ATE \\ $\mu(\sigma)[\si{\centi\meter}]$ } \\
    \midrule[1pt]

    {CTE-MLO}             & \textbf{1.25} & \underline{1.77 (1.6)} & \textbf{3.88 (1.8)} & \textbf{1.24 (0.99)} & \textbf{2.35 (1.13)} & 1.74 (1.79) & \textbf{2.67 (1.48)} & \textbf{1.01 (0.92)} & \textbf{2.04 (0.9)} & \textbf{1.85 (2.3)} & \textbf{4.3 (3.49)} & \textbf{0.91 (0.73)} & \textbf{1.37 (0.9)} \\

    \rowcolor{gray!10} {MA-LIO}              & \underline{2.83} & 6.96 (22.75) & 44.02 (49.35) & 7.28 (10.76) & 14.17 (8.28) & \textbf{1.38 (1.09)} & \underline{4.46 (2.21)} & \underline{1.47 (1.32)} & 5.2 (2.03) & \underline{2.37 (3.06)} & \underline{8.87 (2.46)} & \underline{1.04 (0.92)} & \underline{2.82 (1.29)} \\

    {RESPLE-MLIO}         & \underline{2.83} & \textbf{1.32 (1.42)} & \underline{7.17 (3.25)} & \underline{5 (5.08)} & \underline{10.37 (5.69)} & \underline{1.48 (1.49)} & 13.63 (3.4) & 1.92 (3.25) & 11.03 (8.6) & 3.94 (4.42) & 12.81 (4.47) & 6.16 (12.8) & 33.2 (10.17) \\

    \rowcolor{gray!10} {RESPLE-MLO}          & 4.00 & 1.88 (1.8) & 8.67 (3.97) & 6.78 (7.34) & 11.64 (6.44) & 2.68 (2.51) & 14.91 (3.82) & 1.8 (1.84) & 21.69 (17.59) & - & - & 7.21 (13.32) & 33.44 (11.76) \\

    {FAST-LIO-MULTI}      & 4.25 & 4.96 (3.91) & 10.13 (5.44) & 9.47 (9.81) & 31.16 (22.79) & 5.68 (6.03) & 33.53 (12.32) & 4.57 (3.55) & \underline{5.18 (3.49)} & - & - & 4.22 (4.31) & 7.08 (5.37) \\

    \bottomrule[1pt]
    \end{tabular}%
    }
    \label{table:RTE_ATE_results_MLO_MLIO}
\end{table*}

\subsubsection{Multi-LiDAR-Inertial Odometry}\label{sec:localization_multi_lo}
This family of methods focuses on utilizing the greater coverage of multiple LiDARs on a robotic system.
As discussed in Section~\ref{sec:localization_lio}, it is common to have a limited field of view with a single LiDAR sensor. When LiDAR observations degrade in an LIO system, the only option is to perform dead-reckoning using the inertial measurements.
To prevent this, some methods incorporate multiple LiDAR measurements to increase the field of perception.
The error metrics are reported in Table~\ref{table:RTE_ATE_results_MLO_MLIO}.

To employ these methods, we used the \alias{HG4930} IMU for inertial methods and, where possible, used all three LiDARs.

For {RESPLE}, we used the \alias{Hesai} LiDAR, and the near-field \alias{Livox} LiDARs, whereas for {FAST-LIO-MULTI}, we used the \alias{Hesai} and \alias{VLP16} LiDARs due to compatibility limitations.

As seen in Table~\ref{table:RTE_ATE_results_MLO_MLIO}, {CTE-MLO} is the most consistent method across missions, achieving the best average rank and the lowest ATE on all reported missions, while also attaining the best (or second-best) RTE in most cases.
The remaining approaches are less consistent across environments: {MA-LIO} attains the best RTE on EIG-1 but exhibits substantially larger errors on other missions (e.g., SPX-2 and SNOW-2), and {RESPLE-MLIO} achieves the best RTE on SPX-2 but degrades markedly on ARC-7.
Finally, {RESPLE-MLO} and {FAST-LIO-MULTI} obtain weaker overall ranks, and both fail on ARC-2.

The analysis indicates that multi-LiDAR sensing can improve perception coverage and, consequently, the robustness of odometry estimates, but the accuracy improvement is method- and scenario-dependent rather than uniform across sequences. It should be noted that different LiDAR models may exhibit different noise and coverage characteristics, which, in turn, affect the efficacy of the employed method.

\begin{table*}[t]\centering
    \caption{Error metrics for VIO methods (including RGB-D) on test sequences reported in \si{\centi\meter} (best in bold, second-best \underline{underlined}). \textit{Summary} column reports the average rank across the six missions, where for each mission the method rank is the mean of its RTE-rank and ATE-rank (lower is better); missing results are assigned the worst rank.}
    
    \resizebox{2\columnwidth}{!}{%
    \renewcommand{\arraystretch}{1.18}%
    \begin{tabular}{l c @{\hspace{1.2em}} cc cc cc cc cc cc}
    \toprule[1pt]
    & \multicolumn{1}{c}{Summary} & \multicolumn{2}{c}{SPX-2} & \multicolumn{2}{c}{SNOW-2} & \multicolumn{2}{c}{EIG-1} & \multicolumn{2}{c}{CON-4} & \multicolumn{2}{c}{ARC-2} & \multicolumn{2}{c}{ARC-7} \\
    \cmidrule(lr){2-2} \cmidrule(lr){3-4} \cmidrule(lr){5-6} \cmidrule(lr){7-8} \cmidrule(lr){9-10} \cmidrule(lr){11-12} \cmidrule(lr){13-14}
    & \makecell{Avg.\\rank}
    & \makecell{RTE \\ $\mu(\sigma)[\si{\centi\meter}]$ } & \makecell{ATE \\ $\mu(\sigma)[\si{\centi\meter}]$ }
    & \makecell{RTE \\ $\mu(\sigma)[\si{\centi\meter}]$ } & \makecell{ATE \\ $\mu(\sigma)[\si{\centi\meter}]$ }
    & \makecell{RTE \\ $\mu(\sigma)[\si{\centi\meter}]$ } & \makecell{ATE \\ $\mu(\sigma)[\si{\centi\meter}]$ }
    & \makecell{RTE \\ $\mu(\sigma)[\si{\centi\meter}]$ } & \makecell{ATE \\ $\mu(\sigma)[\si{\centi\meter}]$ }
    & \makecell{RTE \\ $\mu(\sigma)[\si{\centi\meter}]$ } & \makecell{ATE \\ $\mu(\sigma)[\si{\centi\meter}]$ }
    & \makecell{RTE \\ $\mu(\sigma)[\si{\centi\meter}]$ } & \makecell{ATE \\ $\mu(\sigma)[\si{\centi\meter}]$ } \\
    \midrule[1pt]

    {Voxel-SVIO}      & \textbf{1.75} & 1.38 (1.13) & \underline{18.70 (11.95)} & 1.40 (1.17) & \underline{16.17 (8.97)} & \textbf{1.25 (0.98)} & \textbf{12.25 (8.09)} & \textbf{1.34 (1.46)} & \textbf{38.90 (22.06)} & \textbf{1.02 (1.11)} & \textbf{7.69 (3.53)} & \textbf{1.23 (1.23)} & \underline{14.75 (5.99)} \\

    \rowcolor{gray!10} {$\sqrt{\mathrm{VINS}}$} & \underline{3.08} & 1.90 (1.89) & 34.26 (17.32) & 1.23 (0.59) & 23.07 (12.40) & 1.33 (1.72) & 37.41 (22.99) & \underline{1.91 (3.79)} & \underline{80.18 (38.00)} & 4.05 (7.91) & 37.83 (39.06) & \underline{1.24 (2.20)} & \textbf{9.96 (3.55)} \\

    {OKVIS2}          & 4.25 &
    1.342 (1.35) & 28.85 (13.69) & 
    \underline{1.196 (0.958)} & \textbf{6.86 (3.906)} & 
    \underline{1.287 (1.58)} & \underline{30.93 (18.9)} & 
    3.06 (5.36) & 449.03 (236.35) & 
    \underline{1.50 (2.51)} & \underline{14.64 (4.22)} & 
    - & - \\ 

    \rowcolor{gray!10} {Open-VINS}       & 5.42 & \underline{1.32 (2.66)} & 42.75 (24.21) & 2.86 (0.69) & 54.83 (29.97) & 2.20 (2.39) & 85.58 (36.41) & 5.05 (10.16) & 189.11 (141.52) & 2.48 (3.05) & 17.47 (10.27) & - & - \\

    {VINS-RGBD}       & 6.00 & 6.12 (10.32) & 117.47 (44.46) & 3.07 (2.60) & 60.50 (29.28) & 3.43 (6.28) & 100.30 (42.57) & 5.56 (9.91) & 432.92 (148.89) & 19.45 (106.09) & 401.89 (118.22) & 4.41 (5.32) & 30.60 (17.47) \\

    \rowcolor{gray!10} {SchurVINS}       & 6.33 & \textbf{1.13 (1.04)} & \textbf{10.08 (5.18)} & \textbf{1.189 (1.01)} & 20.25 (9.76) & 3.17 (6.67) & 171.51 (81.31) & 3.72 (30.58) & 299.68 (188.51) & - & - & - & - \\

    {ENVIO}           & 8.00 & - & - & 3.71 (2.96) & 88.92 (42.71) & 3.09 (2.90) & 140.43 (70.03) & 33.48 (85.95) & 2041.95 (803.61) & 20.85 (41.52) & 172.18 (108.12) & 15.14 (36.18) & 95.80 (50.00) \\

    \rowcolor{gray!10} {VINS-Fusion}     & 8.50 & 22.31 (29.07) & 405.53 (147.03) & 24.65 (14.22) & 414.18 (147.93) & 27.13 (33.98) & 411.52 (183.93) & 32.56 (55.25) & 1069.63 (428.24) & 27.41 (54.28) & 180.76 (106.39) & 11.21 (14.15) & 112.21 (52.88) \\

    {ROVIO$^*$}           & 8.92 & 32.08 (36.51) & 484.61 (635.29) & 9.21 (8.28) & 168.60 (73.43) & 6.98 (8.87) & 190.66 (138.92) & 12.92 (21.97) & 474.93 (271.68) & 19.70 (59.14) & 192.08 (204.65) & - & - \\

    \rowcolor{gray!10} {R-VIO2$^*$}          & 9.33 & - & - & 3.05 (4.20) & 33.32 (13.02) & 13.90 (42.62) & 457.80 (399.06) & 5.21 (19.48) & 110.28 (162.24) & - & -  & - & - \\

    {R$^2$DIO}        & 9.75 & - & - & 8.39 (6.34) & 1005.50 (491.91) & 10.91 (27.47) & 947.15 (421.05) & 21.08 (95.47) & 691.99 (521.18) & 10.65 (12.52) & 95.29 (26.88) & - & - \\

    \rowcolor{gray!10} {Lightweight-VIO} & 11.17 & 26.81 (52.60) & 760.54 (531.81) & 24.23 (16.05) & 1091.97 (337.18) & 20.06 (17.67) & 499.50 (234.45) & - & - & - & - & - & - \\

    \bottomrule[1pt]
    \end{tabular}%
    }
    \label{table:RTE_ATE_results_VIO}
\end{table*}

\subsubsection{Visual-Inertial Odometry}\label{sec:localization_vio}
Despite the performance of LIO methods, the relatively large weight and size of LiDARs, and the higher base prices, vision-based ego-motion estimation has recently returned to the forefront.
Despite the widespread adoption, technical challenges persist in employing a visual-inertial ego-motion estimation method.
Common challenges include intrinsic calibration, sensitivity to lighting, reduced parallax due to environmental factors, and the dependence of observability in the estimation space on the platform's motion.
Despite these challenges, VIO remains relevant today, thanks to its compatibility with lightweight robotic systems and decades of well-developed understanding of the technology.
Under this premise, we have evaluated recent and well-known Stereo and Monocular VIO pipelines and reported their performance metrics in \Cref{table:RTE_ATE_results_VIO}.
To employ these methods, the \alias{CoreResearch} cameras are utilized: specifically, the front-facing RGB \alias{CoreResearch} camera for monocular methods and the front-facing gray-scale stereo pair for stereo methods.
Moreover, the two RGB-D methods are run with the \alias{ZED2i} RGB-D camera.
For all methods that require inertial data, the \SI{100}{\hertz} \alias{HG4930} IMU is utilized.

Overall, \Cref{table:RTE_ATE_results_VIO} shows that Voxel-SVIO performs consistently well across the evaluated missions, indicating strong robustness to varying scene structure and motion profiles.
At the same time, many pipelines achieve comparable relative translation errors across multiple sequences, as expected for odometry systems that primarily optimize local consistency.
In contrast, absolute trajectory error varies more widely, reflecting differences in long-horizon drift, sensitivity to degradation of visual constraints, and robustness to estimator inconsistency.

The SPX-2 mission begins with highly dynamic motion, which adversely affects initialization for pipelines that require a stationary segment.
This is reflected in \Cref{table:RTE_ATE_results_VIO} by the diverged results.
After successful initialization, several methods maintain competitive odometry estimates, but drift behavior over the remainder of the mission varies considerably.

The SNOW-2 mission covers a very bright, snowy area, which poses challenges for feature extraction and results in a poor distribution of features across the view. This is visible through the elevated ATE and RTE metrics for this sequence.

The CON-4 mission is a particularly demanding, large-scale trajectory with a large loop, and it accumulates long-term drift effects. 
Accordingly, \Cref{table:RTE_ATE_results_VIO} highlights the variety of absolute accuracy on CON-4 across methods, where relative errors remain comparable.

Finally, the ARC missions expose the sensitivity of VIO pipelines to illumination and featureless views.
As seen in \Cref{table:RTE_ATE_results_VIO}, changing lighting conditions adversely affect all methods on ARC-2 and ARC-7, where ARC-7 includes an extended very dark interval during which multiple pipelines diverge or fail to produce valid estimates.

These failures are particularly consequential in our evaluation because the benchmarking setup and many methods lack an external reinitialization mechanism or an automated restart strategy; consequently, once a method diverges, it typically remains failed for the remainder of the mission rather than recovering after the adverse segment.
These findings emphasize the practical importance of robust exposure control, photometric normalization (or relighting), explicit failure detection, and reinitialization for deployment.
\subsubsection{\added{Kinematic-Inertial Odometry}}\label{sec:localization_legged}
\begin{figure*}[t]
    \centering
    \includegraphics[width=1.0\linewidth]{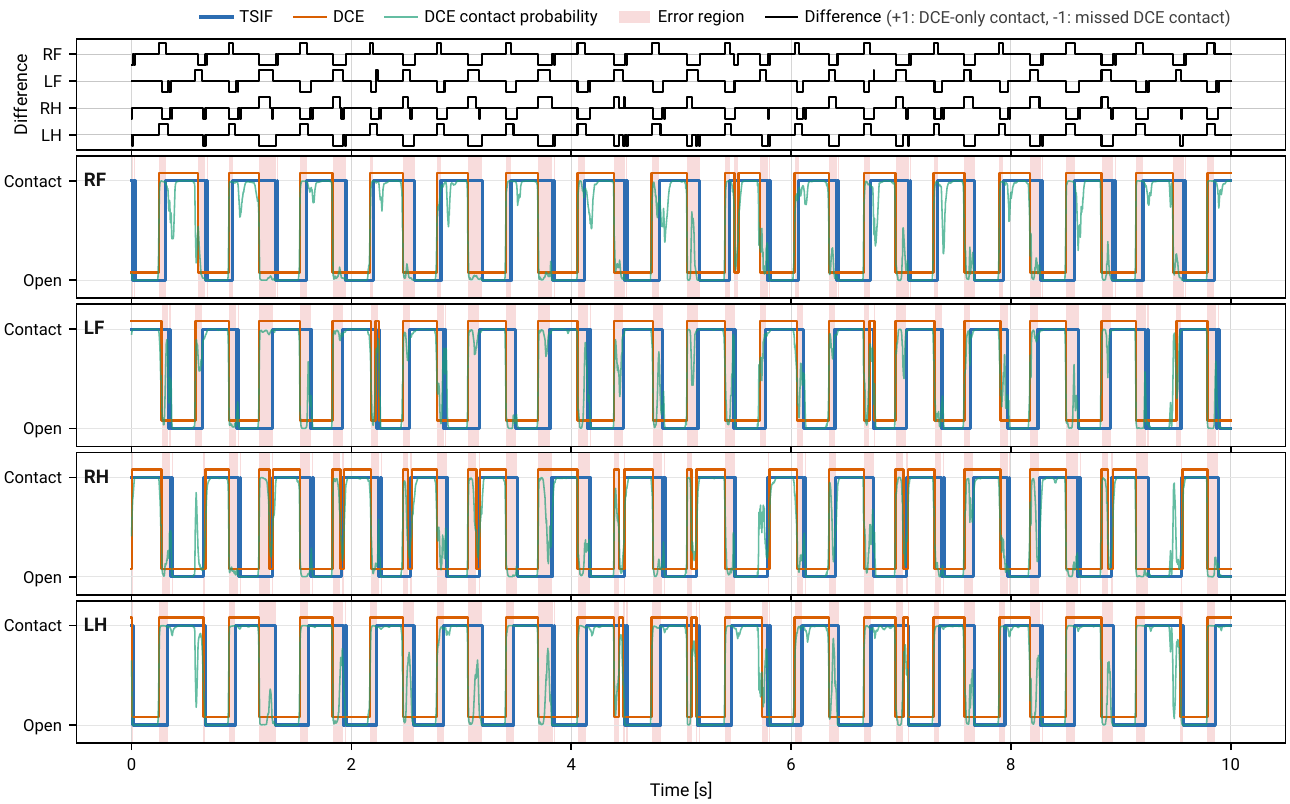}
    \caption{\added{The figure compares binary contact outputs from the onboard ANYmal-D state estimator with learned contacts and contact probabilities from DCE over a \SI{10}{\second} window of the SNOW-2 mission. 
Shaded regions indicate contact-state mismatches; the top row summarizes the signed difference per leg ($+1$: DCE-only contact, $-1$: missed DCE contact).}}
    \label{fig:snow2_contact_dce_vs_anymal}
\end{figure*}

\added{To analyze the maturity of the pure-proprioceptive methods GrandTour benchmarks quadruped legged-specific kinematic-inertial estimators: the ANYmal-D onboard estimator based on TSIF~\citep{bloesch2018tsif}, MUSE~\citep{nistico2025muse}, Cerberus 2.0~\citep{yang2023cerberus}, CAPO~\citep{sun2026capo}, and DRIFT~\citep{lin2023proprioceptive}. 
We additionally test DRIFT with learned contacts from Deep Contact Estimator (DCE)~\citep{lin2021legged}.
These methods use combinations of inertial measurements, leg kinematics, contacts, and robot dynamics, and therefore target failure modes that are not directly represented by exteroceptive-only or generic LIO/VIO pipelines.}
\added{
GrandTour provides joint positions, velocities, and torques, and includes onboard-estimator contact outputs, but it does not provide direct measured foot-contact labels. 
For the KIO benchmark, contact-aware methods were therefore supplied with proxy contact measurements derived from the available proprioceptive streams, such as torque/GRF-based contact estimates.
The DRIFT variant with DCE contacts instead uses DCE contact estimates; DCE is therefore evaluated as a contact-estimation component rather than as a standalone trajectory estimator. 
These proxies and learned contacts are part of the benchmark integration and inclusion efforts hence, should not be interpreted as ground-truth contacts.}
\added{\Cref{table:RTE_ATE_results_KIO} reports the resulting point-distance benchmark against the AP20 prism reference position, and \cref{fig:snow2_contact_dce_vs_anymal} visualizes the SNOW-2 contact-estimation inputs used to compare onboard TSIF contacts with DCE learned contacts. As seen in the figure, the DCE contact estimation consistently starts earlier than the onbard state-estimation contact estimation and it ends earlier. We ensured that this is not result of a systematic delay in evaluation but rather how the contact modelling is done.
The onboard ANYmal estimator, MUSE (Smoother), and MUSE (InEKF) obtain the strongest average ranks, while CAPO is competitive on several absolute-error entries, including CON-4.
It should be mentioned that the MUSE (Smoother) implementation is currently an offline-only implementation. 
The larger mission-dependent drift of Cerberus 2.0, DRIFT, and DRIFT + DCE despite locally reasonable RTE values on some sequences illustrates both the value of the proprioceptive benchmark and the sensitivity of contact-aware KIO methods to robot-specific integration and contact estimation.
Importantly, there is a drastic difference between the accuracy achieved during on-plane motion and the off-plane (z-axis) motion.
This observation alone implies the importance of contact estimation and environment agnostic contact dynamic formulation.
}

\begin{table*}[t]\centering
    \caption{\added{Error metrics for legged-specific kinematic-inertial odometry methods on the six GrandTour benchmark sequences, reported in \si{\centi\meter} (best in \textbf{bold}, second-best \underline{underlined}). \emph{Summary} column reports the average rank across the six missions, where for each mission the method rank is the mean of its RTE-rank and ATE-rank (lower is better). Metrics use the AP20 prism reference, the point-distance RTE/ATE protocol, $\Delta=\SI{0.5}{\meter}$, and no scale correction. For methods requiring explicit contact or stance inputs, proxy contact/stance measurements derived from the available proprioceptive streams were integrated because GrandTour does not provide direct measured foot-contact labels. The DRIFT + DCE row evaluates DRIFT with learned contacts from Deep Contact Estimator.}}
    \resizebox{2\columnwidth}{!}{%
    \renewcommand{\arraystretch}{1.18}%
    \begin{tabular}{l c @{\hspace{1.2em}} cc cc cc cc cc cc}
    \toprule[1pt]
    & \multicolumn{1}{c}{\added{Summary}} & \multicolumn{2}{c}{\added{SPX-2}} & \multicolumn{2}{c}{\added{SNOW-2}} & \multicolumn{2}{c}{\added{EIG-1}} & \multicolumn{2}{c}{\added{CON-4}} & \multicolumn{2}{c}{\added{ARC-2}} & \multicolumn{2}{c}{\added{ARC-7}} \\
    \cmidrule(lr){2-2} \cmidrule(lr){3-4} \cmidrule(lr){5-6} \cmidrule(lr){7-8} \cmidrule(lr){9-10} \cmidrule(lr){11-12} \cmidrule(lr){13-14}
    & \makecell{\added{Avg.}\\\added{rank}}
    & \makecell{\added{RTE} \\ \added{$\mu(\sigma)[\si{\centi\meter}]$} } & \makecell{\added{ATE} \\ \added{$\mu(\sigma)[\si{\centi\meter}]$} }
    & \makecell{\added{RTE} \\ \added{$\mu(\sigma)[\si{\centi\meter}]$} } & \makecell{\added{ATE} \\ \added{$\mu(\sigma)[\si{\centi\meter}]$} }
    & \makecell{\added{RTE} \\ \added{$\mu(\sigma)[\si{\centi\meter}]$} } & \makecell{\added{ATE} \\ \added{$\mu(\sigma)[\si{\centi\meter}]$} }
    & \makecell{\added{RTE} \\ \added{$\mu(\sigma)[\si{\centi\meter}]$} } & \makecell{\added{ATE} \\ \added{$\mu(\sigma)[\si{\centi\meter}]$} }
    & \makecell{\added{RTE} \\ \added{$\mu(\sigma)[\si{\centi\meter}]$} } & \makecell{\added{ATE} \\ \added{$\mu(\sigma)[\si{\centi\meter}]$} }
    & \makecell{\added{RTE} \\ \added{$\mu(\sigma)[\si{\centi\meter}]$} } & \makecell{\added{ATE} \\ \added{$\mu(\sigma)[\si{\centi\meter}]$} } \\
    \midrule[1pt]

    \added{ANYmal estimator} & \added{\textbf{2.17}} & \added{1.37 (1.18)} & \added{\textbf{16.25 (10.16)}} & \added{1.97 (2.23)} & \added{\textbf{86.31 (47.61)}} & \added{\textbf{1.69 (2.94)}} & \added{\textbf{60.83 (39.42)}} & \added{1.41 (1.79)} & \added{\underline{95.27 (51.93)}} & \added{\textbf{1.24 (1.48)}} & \added{16.31 (12.60)} & \added{\textbf{1.05 (0.83)}} & \added{15.00 (7.36)} \\
    \rowcolor{gray!10} \added{MUSE (Smoother)} & \added{\underline{2.25}} & \added{\textbf{1.01 (1.42)}} & \added{\underline{28.48 (14.69)}} & \added{1.88 (2.29)} & \added{138.71 (67.70)} & \added{\underline{1.83 (5.37)}} & \added{\underline{78.55 (38.13)}} & \added{\textbf{0.86 (1.12)}} & \added{308.70 (244.59)} & \added{\underline{1.41 (2.04)}} & \added{\textbf{11.64 (5.90)}} & \added{\underline{1.39 (3.02)}} & \added{\textbf{12.46 (6.80)}} \\
    \added{MUSE (InEKF)} & \added{2.75} & \added{\underline{1.02 (1.51)}} & \added{29.15 (15.42)} & \added{\underline{1.88 (2.29)}} & \added{138.65 (67.71)} & \added{1.84 (5.37)} & \added{78.76 (38.21)} & \added{\underline{0.87 (1.11)}} & \added{307.99 (243.28)} & \added{1.41 (2.03)} & \added{\underline{11.90 (5.95)}} & \added{1.40 (3.03)} & \added{\underline{12.80 (7.13)}} \\
    \rowcolor{gray!10} \added{DRIFT} & \added{4.67} & \added{8.03 (90.48)} & \added{1090.31 (643.32)} & \added{\textbf{1.60 (2.47)}} & \added{138.48 (70.60)} & \added{2.11 (10.51)} & \added{940.15 (419.74)} & \added{1.27 (3.66)} & \added{503.16 (399.94)} & \added{1.92 (5.01)} & \added{19.34 (12.09)} & \added{1.75 (3.02)} & \added{176.83 (101.46)} \\
    \added{CAPO} & \added{5.00} & \added{3.84 (6.48)} & \added{134.88 (80.73)} & \added{4.58 (3.24)} & \added{\underline{134.23 (67.69)}} & \added{4.83 (20.37)} & \added{334.36 (109.32)} & \added{4.54 (5.66)} & \added{\textbf{74.83 (52.09)}} & \added{4.86 (9.33)} & \added{56.11 (54.35)} & \added{3.04 (3.21)} & \added{13.29 (7.46)} \\
    \rowcolor{gray!10} \added{MUSE} & \added{5.50} & \added{2.95 (2.76)} & \added{75.71 (30.51)} & \added{3.66 (2.57)} & \added{166.94 (69.03)} & \added{5.12 (8.22)} & \added{341.79 (201.32)} & \added{4.13 (10.39)} & \added{203.18 (88.60)} & \added{3.42 (4.82)} & \added{49.34 (18.49)} & \added{4.59 (10.35)} & \added{41.64 (19.30)} \\
    \added{Cerberus 2.0} & \added{6.00} & \added{3.70 (15.74)} & \added{389.13 (227.86)} & \added{2.41 (2.63)} & \added{441.86 (163.30)} & \added{2.78 (9.18)} & \added{727.67 (378.84)} & \added{2.87 (33.93)} & \added{2376.13 (1847.59)} & \added{10.17 (28.84)} & \added{201.47 (63.57)} & \added{2.16 (8.19)} & \added{266.50 (158.61)} \\
    \rowcolor{gray!10} \added{DRIFT + DCE} & \added{7.67} & \added{9.02 (71.01)} & \added{1382.62 (721.30)} & \added{3.47 (3.48)} & \added{700.72 (329.45)} & \added{4.89 (12.13)} & \added{1562.53 (662.90)} & \added{5.11 (38.69)} & \added{4082.33 (3021.16)} & \added{24.29 (91.86)} & \added{735.75 (465.84)} & \added{3.05 (6.34)} & \added{535.37 (312.57)} \\

    \bottomrule[1pt]
    \end{tabular}%
    }
    \label{table:RTE_ATE_results_KIO}
\end{table*}

\begin{table*}[t]\centering
    \caption{Error metrics for Fast-LIMO, Fast-LIO, and Faster-LIO both with and without their online calibration feature, reported in \si{\centi\meter} (with the better performing version indicated in \textbf{bold}).}
    \resizebox{2\columnwidth}{!}{
    \begin{tabular}{l c cc cc cc cc cc cc}
    \toprule[1pt]
    & & \multicolumn{2}{c}{SPX-2} & \multicolumn{2}{c}{SNOW-2} & \multicolumn{2}{c}{EIG-1} & \multicolumn{2}{c}{CON-4} & \multicolumn{2}{c}{ARC-2} & \multicolumn{2}{c}{ARC-7} \\
    \cmidrule(lr){3-4} \cmidrule(lr){5-6} \cmidrule(lr){7-8} \cmidrule(lr){9-10} \cmidrule(lr){11-12} \cmidrule(lr){13-14}
    & \makecell{Online \\ Calib}
    & \makecell{RTE \\ $\mu(\sigma)[\si{\centi\meter}]$ } & \makecell{ATE \\ $\mu(\sigma)[\si{\centi\meter}]$ }
    & \makecell{RTE \\ $\mu(\sigma)[\si{\centi\meter}]$ } & \makecell{ATE \\ $\mu(\sigma)[\si{\centi\meter}]$ }
    & \makecell{RTE \\ $\mu(\sigma)[\si{\centi\meter}]$ } & \makecell{ATE \\ $\mu(\sigma)[\si{\centi\meter}]$ }
    & \makecell{RTE \\ $\mu(\sigma)[\si{\centi\meter}]$ } & \makecell{ATE \\ $\mu(\sigma)[\si{\centi\meter}]$ }
    & \makecell{RTE \\ $\mu(\sigma)[\si{\centi\meter}]$ } & \makecell{ATE \\ $\mu(\sigma)[\si{\centi\meter}]$ }
    & \makecell{RTE \\ $\mu(\sigma)[\si{\centi\meter}]$ } & \makecell{ATE \\ $\mu(\sigma)[\si{\centi\meter}]$ } \\
    \toprule[1pt]

    \makecell{Fast-LIMO}  & Yes & \textbf{1.15 (0.99)} & \textbf{5.76 (2.58)} & 1.16 (0.90) & 2.41 (1.31) & 1.10 (1.60) & 8.75 (3.30) & 0.89 (0.74) & \textbf{2.65 (1.50)} & \textbf{2.49 (3.35)} & 10.20 (6.94) & 0.61 (0.50) & \textbf{2.31 (1.57)} \\
    \makecell{Fast-LIMO}  & No  & 1.32 (1.05) & 6.99 (3.62) & \textbf{1.01 (0.86)} & \textbf{2.11 (1.08)} & \textbf{0.89 (0.77)} & \textbf{5.79 (1.92)} & \textbf{0.89 (0.68)} & 2.73 (1.55) & 2.71 (3.73) & \textbf{9.64 (6.17)} & \textbf{0.59 (0.47)} & 2.45 (1.54) \\[0.3em]
    \cdashline{1-14}\\[-1.8ex]

    \makecell{Faster-LIO} & Yes & 
    1.02 (0.89) & \textbf{2.68 (1.26)} & 
    1.46 (1.35) & \textbf{7.15 (2.55)} & 
    \textbf{1.30 (1.13)} & 5.38 (2.73) & %
    1.51 (1.86) & \textbf{9.04 (3.93)} & 
    2.01 (3.49) & \textbf{5.88 (5.26)} & 
    0.94 (0.73) & \textbf{1.49 (0.65)} \\ 
    
    \makecell{Faster-LIO} & No  & \textbf{0.99 (0.82)} & 2.85 (1.32) & \textbf{1.46 (1.15)} & 7.18 (2.48) & 1.32 (1.11) & \textbf{5.13 (2.66)} & \textbf{1.50 (1.88)} & 9.25 (3.99) & \textbf{2.00 (3.75)} & 11.75 (5.08) & \textbf{0.89 (0.69)} & 1.52 (0.68) \\[0.3em]
    \cdashline{1-14}\\[-1.8ex]
    
    \makecell{Fast-LIO}   & Yes & \textbf{1.07 (0.90)} & 4.15 (2.26) & 1.41 (1.13) & 6.47 (2.67) & 1.28 (0.95) & 4.97 (2.65) & 1.54 (1.86) & 8.79 (3.83) & 3.62 (10.63) & 47.63 (16.36) & \textbf{1.04 (0.72)} & 2.74 (1.45) \\
    \makecell{Fast-LIO}   & No  & 1.11 (0.83) & \textbf{3.61 (2.13)} & \textbf{1.24 (1.03)} & \textbf{1.65 (0.88)} & \textbf{1.19 (0.87)} & \textbf{3.71 (1.55)} & \textbf{1.24 (0.96)} & \textbf{1.86 (0.80)} & \textbf{3.00 (7.47)} & \textbf{8.66 (9.73)} & 1.17 (0.82) & \textbf{2.64 (1.43)} \\

    \bottomrule[1pt]
    \end{tabular}
    \label{table:online_calib_lio}
    }
\end{table*}

\subsubsection{Ablation: Online-Calibration}\label{sec:ablation_online_calibration}
It is well known in the robotics community that the quality of the extrinsic calibration between tightly coupled sensors is crucial for achieving top-quality results.
Conversely, obtaining high-quality extrinsic calibration is a tedious process that is often neglected in real-world research but is deeply considered in industry.
Hence, some solutions include an online-calibration module that estimates sensor-to-sensor calibration online.
The error metrics are reported in \Cref{table:online_calib_lio} for both online calibration enabled and disabled cases.

The results are mixed and do not indicate a consistent advantage of enabling online calibration.
For {Fast-LIMO}, online calibration improves some missions (e.g., SPX-2 and ATE on CON-4/ARC-7), but is worse on others (notably SNOW-2 and EIG-1, and ARC-7 RTE).
For {Faster-LIO}, online calibration tends to help ATE more often (including a large gain on ARC-2 ATE), while RTE is frequently similar or slightly worse.
For {Fast-LIO}, enabling online calibration is worse on most missions and metrics, in several cases by a large margin in ATE. 
This may indicate poor observability or implementation/configuration issues.

A plausible explanation is that online-calibration parameters are not uniformly observable: their observability depends on the motion excitation and the geometric constraints imposed by the observations.
When the data weakly constrain some calibration modes, estimating them online can have limited benefit and may correlate with other states.
Accordingly, online calibration of LIO systems should be treated as a scenario-dependent feature rather than a guaranteed improvement. Lastly, in real-world robotic systems, sensors are mounted rigidly, and variations in mounting pose due to environmental conditions are often negligible compared to sensor noise~\citep{frey2025boxi}.
\subsubsection{Summary: State estimation and localization}\label{sec:application:discussion:localization}
Across the evaluated LO, LIO/LIVO, multi-LiDAR, VIO, and \added{legged-specific KIO} pipelines (\Cref{table:RTE_ATE_results_LO,table:RTE_ATE_results_LIO_LiDARSLAM,table:RTE_ATE_results_MLO_MLIO,table:RTE_ATE_results_VIO,table:RTE_ATE_results_KIO}), relative translation errors are often comparable on the easier missions, while absolute trajectory errors separate methods much more strongly.
In general, LiDAR-inertial methods are more accurate and drift-stable than pure LO and especially than VIO, with Coco-LIC and FAST-LIVO2 consistently among the strongest, while Traj-LO represents the most competitive pure-LO baseline when conditions are favorable.
The large-scale CON-4 mission presents a clear challenge for long-term consistency: several methods remain locally reasonable but accumulate substantial drift, with particularly severe degradation visible in multiple VIO pipelines (e.g., ENVIO, VINS-Fusion, Open-VINS) compared to the better LiDAR-inertial solutions.
Multi-LiDAR methods can be highly competitive, such as the CTE-MLO, which performs consistently, but the benefit is method-dependent (e.g., MA-LIO and the RESPLE multi-LiDAR variants vary more), reflecting the added sensitivity to calibration/synchronization and configuration.
Failure cases are present in every category: VIO shows the most frequent failures on the hardest sequences, while LiDAR-based methods fail less often but exhibit higher error metrics due to visibility and feature-extraction reliability.
Within missions, poorer local accuracy generally correlates with higher drift (e.g., weaker LO baselines such as KISS-ICP tend to drift more), but good local performance does not guarantee low accumulated error over long trajectories.
Overall, cross-mission robustness, initialization resilience, and explicit failure handling/recovery matter at least as much as incremental metric gains on any single sequence.

\paragraph{Sensitivity to tuning:}
Across nearly all evaluated methods, performance is highly sensitive to parameter choices, and many methods are susceptible to overtuning to specific motion profiles, sensor characteristics, or environments.
In practice, this results in significant performance variability when transferring a configuration between missions, even within the same platform and sensor suite.
Notably, this sensitivity is not limited to complex pipelines with many hyperparameters: methods marketed as simple, robust, or adaptive, and often described as approaches that ``just work,'' also exhibit substantial performance swings under modest changes in conditions (e.g., dynamics, lighting, scene structure) or when default settings are used.
Overall, our analysis suggests that there is still room for methods that are truly adaptive to the environments in which they are deployed.

\paragraph{Technical depth and extendability:} There is a corpus of open-source methods that build on top of cornerstone works such as Fast-LIO2~\citep{fastlio2}, DLIO~\citep{dlio}, Vins-Mono~\citep{vinsmono}, and LOAM~\citep{loam}. 
In several cases, inherited implementation complexity and legacy components make it difficult to assess which algorithmic elements are active in a given configuration. 
This complicates third-party evaluation and motivates careful reporting of configuration choices alongside numerical results.

\paragraph{Benchmarking practice}: State estimation and localization papers are commonly evaluated against publicly available baselines, but evaluation protocols remain fragmented. 
One recurring challenge is the configuration and tuning of third-party solutions: reproducing reported performance often requires method-specific expertise, and a configuration that is strong on one dataset may transfer poorly to another. 
In addition, centimeter-level trajectory metrics are sensitive to synchronization, alignment, output rate, and reinitialization assumptions, especially for continuous-time methods or systems that recover after failures in a new fixed frame. 
In this work, we apply a common protocol across all evaluated pipelines and report per-mission results to make these sensitivities visible.

\begin{figure*}[t]
    \centering
    \includegraphics[width=1.0\linewidth]{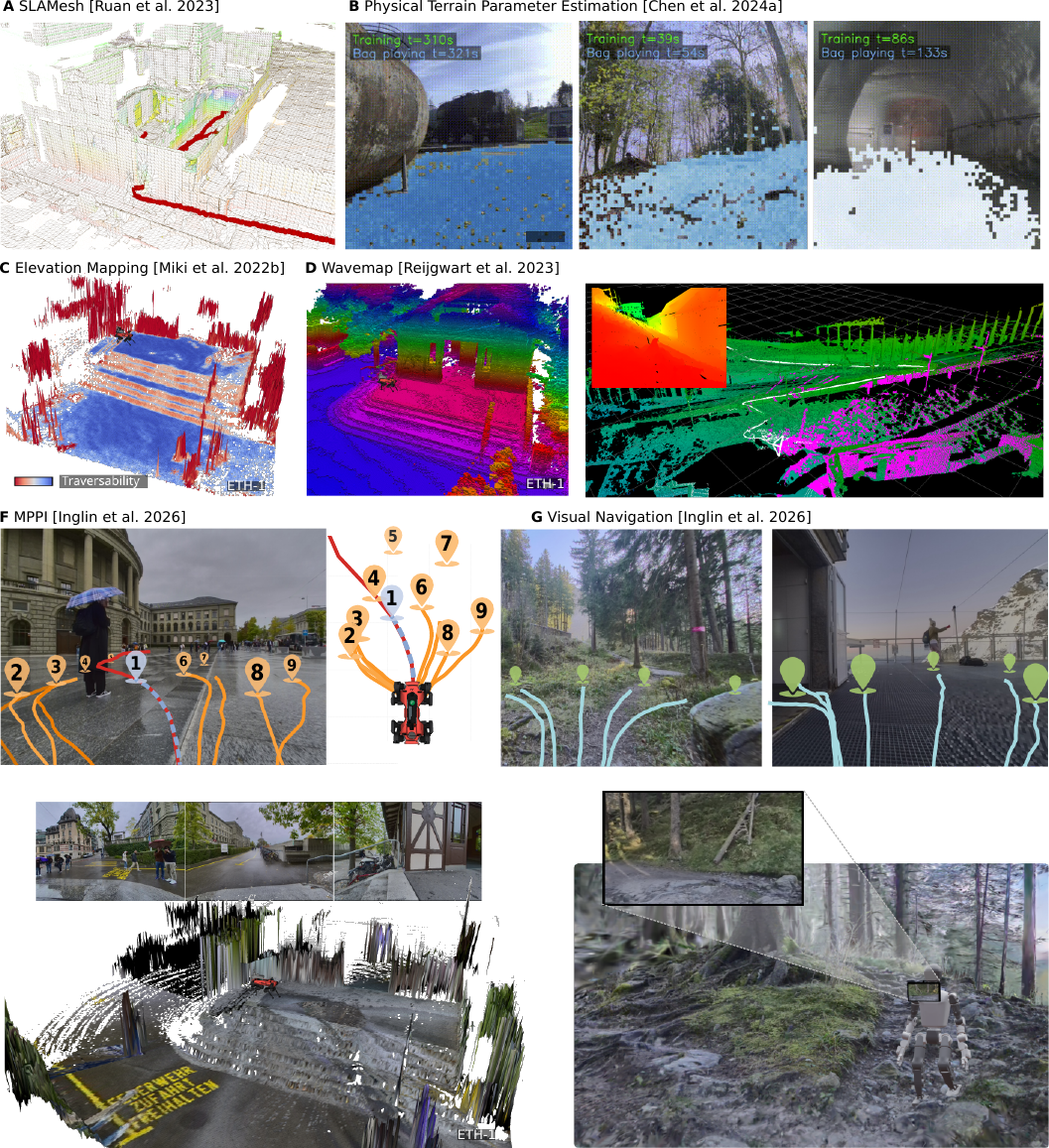}
\caption{Overview of GrandTour applications for perception, locomotion, and navigation.
\textbf{(A)} Online mesh generation with SLAMesh~\citep{slammesh}.
\textbf{(B)} Prediction of friction parameters from RGB images using training data collected via proprioceptive terrain interaction~\citep{chen2024identifying}.
\textbf{(C)} Integration with Elevation Mapping Cupy~\citep{miki2022elevation}, enabling the assessment of foothold placement scores.
\textbf{(D)} WaveMap~\citep{reijgwart2023wavemap} 3D probabilistic occupancy mapping integration.
\textbf{(E)} LiDAR-based online mesh generation and depth rendering with the ImMesh~\citep{immesh} meshing pipeline.
\textbf{(F)} Illustration of MPPI-based path planning~\citep{lessismore} and the teleoperated trajectory.
\textbf{(G)} Less is More~\citep{lessismore} visual navigation predictions.
\textbf{(H)} Projection of camera images onto the map using Multi-modal Elevation Map (MEM)~\citep{erni2023mem}.
\textbf{(I)} GaussGym~\citep{gauss} enables training RGB-based locomotion policies in simulation, providing photorealistic rendering via 3D Gaussian Splatting and mesh reconstructions.}

    \label{fig:multi_modal_perception}
\end{figure*}

\subsection{Perception}\label{sec:multi_modal}
Today, GrandTour supports multiple works aimed at improving the perception capabilities of legged robots. While a comprehensive discussion of these contributions is beyond the scope of this section, we highlight several studies and integrated tools. These examples are illustrated in \cref{fig:multi_modal_perception}.

\textbf{LiDAR-based 3D mesh generation:} As shown in \cref{fig:multi_modal_perception}-(A,E), we have integrated open-source 3D online mesh generation methods ImMesh~\citep{immesh} and SLAMesh~\citep{slammesh} to demonstrate the performance of state-of-the-art online mesh generation. These LiDAR-based online mesh generation methods can accelerate the convergence of radiance field methods, enable better foothold placement for locomotion, or improve mission planning through increased environmental awareness.

\textbf{Multi-Modal Elevation Mapping:} We integrated MEM~\citep{miki2022elevation, erni2023mem}, shown in \cref{fig:multi_modal_perception}-(H), a GPU-accelerated elevation-mapping framework that allows the fusion of depth measurements into a local or global elevation map representation. Furthermore, visual data such as RGB images, semantic classes, and latent features can be easily integrated as separate layers into the same representation.

\textbf{WaveMap:} As shown in \cref{fig:multi_modal_perception}-(D), we integrated WaveMap~\citep{reijgwart2023wavemap}, a 3D multi-resolution volumetric mapping framework that fuses depth measurements into a wavelet-based 3D representation. This framework supports query of occupancy, depth rendering, and enables high-rate planning applications.

\textbf{Geometric Understanding:} Accurate time synchronization between LiDAR and cameras enables the generation of ground-truth metric geometry.
While various datasets exist, only a limited number of datasets provide collections in natural environments. 
Using the GrandTour dataset, we trained and evaluated monocular depth estimation networks with a specific focus on natural environments. 
This can be further extended to stereo depth estimation.

\textbf{Physical Properties Detection:} The method of \citet{chen2024identifying}, shown in \cref{fig:multi_modal_perception}-(B), enables the estimation of simulation parameters, corresponding to physical terrain properties, from proprioceptive joint information.
 These parameters, such as terrain stiffness and friction, can then be used to train a vision-based decoder that predicts the same terrain properties directly from visual input.

\textbf{Dynamic Point Segmentation:} We integrated functionality to filter out dynamic points based on pre-trained image segmentation models. 
This facilitates more accurate volumetric mapping of environments by directly removing depth measurements associated with dynamic semantic classes. The generated masks are available on HuggingFace.

\textbf{NerfStudio:} We provide converters that enable the direct transformation of the GrandTour dataset into NerfStudio-compatible formats for training neural scene representations~\citep{nerfstudio}. 
GrandTour enables the study of the effects of integrating sparse and dense depth measurements into multimodal neural rendering research.

\subsection{Locomotion and Navigation}\label{sec:loc_nav}
The GrandTour dataset supports the development of locomotion and navigation capabilities for legged robots. 
All deployments are collected using the \revised{reinforcement-learning locomotion controller of \citet{miki2022learning}}, and the robot is deliberately and safely teleoperated throughout the environment with specific navigation goals or inspection targets in mind. 
In the following, we highlight projects that directly benefited from the GrandTour dataset or extended it:

\textbf{NaviTrace:} This project \citep{navitrace} uses a subset of the HDR camera images within GrandTour to establish a visual question answering (VQA) benchmark. This benchmark evaluates the capabilities of vision-language models to generate navigation behavior. Specifically, human-annotators perform a key frame selection and annotate the selected key frames with navigation task instructions and an image trace (set of image coordinates) defining the navigation behavior, for different robot embodiments. This dataset enables evaluating and fine-tuning Vision-Language models for navigation.

\textbf{MPPI Path Planning:}
The GrandTour dataset facilitates the development of classical motion planners. 
For example, in previous work, we implemented an MPPI-based path planner that takes as input the robot’s local traversability analysis, computed using the MEM framework—along with a goal position, and outputs a set of safe actions that lead the robot toward that goal. 
This workflow is shown in \cref{fig:multi_modal_perception}-(F).
However, specifying the desired robot behavior requires careful tuning of numerous hyperparameters. Using a recorded reference behavior, we demonstrated that black-box optimization methods can automatically adjust these parameters to minimize the error between the executed and MPPI-predicted trajectories. 
This results in a highly effective, robust path-planning system.

\textbf{Visual Navigation:} The GrandTour dataset can be seamlessly integrated into existing projects, for example, to support learning visual navigation costs, which in prior work required collecting specialized datasets for testing and benchmarking~\citep{frey2023}. In more recent work, the dataset has been used for imitation learning. A key strength of GrandTour is the availability of dense geometric information around the robot, which enables the generation of paths to arbitrary goals using methods such as MPPI-based planning, as illustrated in \cref{fig:multi_modal_perception}-(G). This greatly increases the diversity of available navigation trajectories, enabling state-of-the-art visual navigation performance even when training on a comparatively small dataset. The synthetic paths are available on HuggingFace.

\textbf{GaussGym:} While recent work has shown that even limited data can suffice to learn world models that support training locomotion policies~\citep{li2025robotic}, the dominant paradigm in robotics remains sim-to-real transfer. 
GaussGym~\citep{gauss} enables real-to-sim transfer to enhance simulation capabilities by directly integrating deployment environments within the GrandTour into simulation as shown in \cref{fig:multi_modal_perception}-(I). 
This is achieved by neural radiance fields and scene reconstruction techniques.
Furthermore, GaussGym~\citep{gauss} demonstrates how RGB-based visual locomotion policies can be trained fully in simulation.

\section{Discussion and limitations}\label{sec:discussion}
The GrandTour dataset represents an extensive multi-modal data resource for legged robotics research, yet several limitations must be acknowledged to guide future developments.
While the dataset covers a wide range of environmental conditions, its current geographic scope is limited to Switzerland. 
This regional focus may introduce biases related to specific architectural styles, vegetation, and ambient lighting conditions.
The dataset today does not contain revisits of the same environments under varying conditions (different seasons, weather patterns, or times of day).
Despite the comprehensive sensor suite, the dataset currently lacks certain emerging modalities, such as radar, FMCW LiDAR, and event cameras.
\revised{While, GrandTour does not currently include annotated semantic labels or dynamic-object segmentation masks, we view this as a natural direction for community-driven extension and welcome contributed annotations on top of the released sequences.}

\section{Conclusion and future work}\label{sec:conclusion}
In this work, we have presented the GrandTour dataset, a comprehensive multi-modal legged robotics dataset providing diverse sensor modalities, environmental coverage, motion characteristics, and high-precision ground truth.
The dataset contains 49 sequences spanning a wide range of environments, illumination conditions, and weather types, exceeding the scope and diversity of existing legged robot datasets.\looseness-1

A distinguishing feature of GrandTour is the availability of centimeter-level ground truth poses through NovAtel's Inertial Explorer solution when GNSS is available, as well as millimeter-level and time-synchronized ground truth positions obtained through Leica MS60 total station measurements and accurate extrinsic calibration.

Using time-synchronized RTK-GNSS/total-station reference trajectories, we establish a benchmark for real-time \revised{state estimation and localization} on GrandTour by evaluating \revised{63 state-estimation methods} across six representative missions and reporting per-mission ATE/RTE, ranks, and failure cases for LO, LIO/LIVO, multi-LiDAR(-inertial), VIO/RGB-D, and \added{KIO} methods. 
The benchmark exposes mission-dependent robustness limits, initialization sensitivity, and tuning effects under real-world legged-robot motion and sensing conditions.
\added{The KIO results further indicate that current kinematic-inertial estimators are still not mature, robot-agnostic modules: many depend on hardware-specific contact assumptions, force-torque sensing, or estimator-provided contact states, which limits reliable out-of-the-box transfer to arbitrary legged robots when direct force-torque or contact sensors are unavailable.}
For each pipeline, we implemented the required dataset interfaces, if necessary, and ran it in an online, real-time configuration. 
We then spent substantial effort configuring each method according to its paper and repository guidelines, verifying correct synchronization and calibration usage, and making only the minimal changes required for stable execution rather than per-sequence overfitting. 
To keep comparisons consistent, we disabled loop closure for all methods while retaining their online optimization components, and evaluated every output with the same protocol \revised{while taking each method's output frequency into account.} 
\removed{Methods that could not be run reliably were excluded (16 methods have been excluded) rather than speculated about.}

Furthermore, the dataset is accompanied by post-processed outputs and open-source software tools to facilitate straightforward data access and analysis.
\revised{The synchronized multi-modal dataset supports research on sensor fusion, depth estimation, real-to-sim transfer, and many perception applications, from training foundation models to self-supervised cross-modal learning.}
Furthermore, the GrandTour can support the development and benchmarking of traversability assessment and navigation.

Looking ahead, we plan to expand GrandTour in several aspects.
First, we aim to incorporate additional missions featuring extreme perception challenges, such as those in volcanic scenery and extended-duration deployments, a wider range of environments, and a broader set of robots, featuring bipeds and wheeled-legged robots. 
\added{Future missions and extensions are planned to include ground truth scans through a Leica RTC-360 stationary scanner, which will facilitate mapping quality evaluations.}
Second, we will upgrade the sensor suite to include emerging technologies such as higher-resolution solid-state LiDARs, FMCW LiDARs, radars, and event cameras.
\added{Third, we will continue expanding GrandTour dataset with different embodiments such as wheeled-legged, wheeled, tracked systems to facilitate embodiment-agnostic software and autonomy intelligence.}

Through these systematic expansions, GrandTour will continue to serve as a valuable benchmark and resource for the robotics and computer vision research communities, bridging the gap between controlled laboratory experiments and real-world deployment scenarios for legged robots.

\begin{acks}
We thank William Talbot for integrating the ADIS 16475 IMU and serving as an alpha-tester; Julian Nubert for his early support in development and for running Holistic Fusion on the GrandTour data; Yves Inglin and Changang Chen for visual navigation; Alejandro Escontrela and Justin Kerr for the integration of GaussGym; Carlos Pérez for assistance in running Limoncello; Victor Reijgwart for help integrating Wavemap; Cyrill P\"{u}ntener for expertise in network configurations; Dr.~Ylenia Nistic\`o for supporting evaluation of MUSE; and Benjamin Krummenacher for his support.

This work was supported in part by Hexagon\:|\:Leica Geosystems. We are grateful to Benjamin Müller, Benjamin Sch\"oll, Bernhard Sprenger, Stefan Eisenreich, and Jonas Nussdorfer for their expertise in networking, calibration, time-synchronization, and for their assistance in guiding us with the collection of calibration data. 
We further thank Leica Geosystems for providing the MS60 total station and a custom AP20 unit, which were essential for obtaining high-precision ground truth measurements. 
Additionally, we appreciate the advice of Dr. Matej Varga and Dr. Jemil Avers Butt (Department of Civil, Environmental, and Geomatic Engineering, ETH Zurich) on robotic total station operation.
Moreover, we acknowledge the support of Hexagon\:|\:NovAtel, with special thanks to Ryan Dixon for guidance on the utilization of the NovAtel CPT7 and Inertial Explorer solutions.\looseness-1

We acknowledge that the International Foundation High Altitude Research Stations Jungfraujoch and Gornergrat (HFSJG), 3012 Bern, Switzerland, made it possible for us to carry out our experiment(s) at the High Altitude Research Station at Jungfraujoch. We also thank the custodians, Mrs.~Daniela Bissig and Mr.~Erich Bissig, guide Ms.~Doris Graf Jud (Jungfraubahnen), and Annette Fuhrer (Jungfraubahnen) for their support during the Jungfraujoch trip.

This work would not have been possible without the combined efforts of these partners and collaborators.

Finally, this work is partially funded by the National Centre of Competence in Research Robotics (NCCR Robotics), the ETH RobotX research grant, which is funded through the ETH Zurich Foundation, and the European Union's Horizon 2020 research and innovation program under grant agreement No. 101070405 and No.~101070596. This work was also supported by the Open Research Data Grant at ETH Zurich and by the Swiss National Science Foundation (SNSF) as part of project No. 227617. This work was supported by the Swiss National Science Foundation (SNSF) [20CH-1\_229464/1] under CHIST-ERA grant CHIST-ERA-23-MultiGIS-07.
The research leading to these results has received funding from armasuisse Science and Technology.
\end{acks}

\bibliographystyle{SageH}
\bibliography{references}

\end{document}